\documentclass[10pt,twocolumn,letterpaper]{article}

\makeatletter
\def\@fnsymbol#1{\ensuremath{\ifcase#1\or \dagger\or \ddagger\or
   \mathsection\or \mathparagraph\or \|\or **\or \dagger\dagger
   \or \ddagger\ddagger \else\@ctrerr\fi}}
    \makeatother
\usepackage[pagenumbers]{cvpr} 

\usepackage{amsmath}
\usepackage{amssymb}
\usepackage{booktabs}

\usepackage{amsmath,amsfonts,bm}

\def\eqref#1{equation~\ref{#1}}

\def\1{\bm{1}}

\def\vx{{\bm{x}}}

\def\vdelta{{\bm{\delta}}}

\def\mA{{\bm{A}}}

\def\mP{{\bm{P}}}

\def\mU{{\bm{U}}}

\DeclareMathAlphabet{\mathsfit}{\encodingdefault}{\sfdefault}{m}{sl}
\SetMathAlphabet{\mathsfit}{bold}{\encodingdefault}{\sfdefault}{bx}{n}

\def\sP{{\mathbb{P}}}

\def\sR{{\mathbb{R}}}

\DeclareMathOperator*{\argmax}{arg\,max}

\usepackage[ruled,linesnumbered]{algorithm2e}
\usepackage{url}
\usepackage{color}
\usepackage{multirow}
\usepackage{makecell}
\usepackage{subfigure} 
\usepackage{graphicx}
\usepackage{caption}
\usepackage{tablefootnote}
\usepackage{enumitem}
\usepackage{cite}
\newcommand*\samethanks[1][\value{footnote}]{\footnotemark[#1]}

\usepackage[pagebackref,breaklinks,colorlinks]{hyperref}

\usepackage[capitalize]{cleveref}
\crefname{section}{Sec.}{Secs.}
\Crefname{section}{Section}{Sections}
\Crefname{table}{Table}{Tables}
\crefname{table}{Tab.}{Tabs.}

\def\cvprPaperID{4357} 
\def\confName{CVPR}
\def\confYear{2022}

\newcommand{\BK}[1]{\textcolor{red}{BK: #1}}
\newcommand{\AM}[1]{\textcolor{cyan}{AM: #1}}

\definecolor{neonpurple}{rgb}{0.3,0,1}

\begin{document}

\title{Certified Adversarial Defenses Meet Out-of-Distribution Corruptions: Benchmarking Robustness and Simple Baselines}

\author{Jiachen Sun\protect\thanks{Work partially done during internship at Lawrence Livermore National Laboratory (LLNL).}\\
\small{University of Michigan}\\
{\tt\small jiachens@umich.edu}
\and
Akshay Mehra\samethanks\\
\small{Tulane University}\\
{\tt\small amehra@tulane.edu}
\and
Bhavya Kailkhura\\
\small{Lawrence Livermore National Laboratory}\\
{\tt\small kailkhura1@llnl.gov}
\and
Pin-Yu Chen\\
\small{IBM Research}\\
{\tt\small pin-yu.chen@ibm.com}
\and
Dan Hendrycks\\
\small{UC Berkeley}\\
{\tt\small hendrycks@berkeley.edu}
\and
Jihun Hamm\\
\small{Tulane University}\\
{\tt\small jhamm3@tulane.edu}
\and
Z. Morley Mao\\
\small{University of Michigan}\\
{\tt\small zmao@umich.edu}
}

\maketitle
\vspace{-0.25cm}

\begin{abstract}
Certified robustness guarantee gauges a model's robustness to test-time attacks and can assess the model's readiness for deployment in the real world.
In this work, we critically examine how the adversarial robustness guarantees from randomized smoothing-based certification methods change when state-of-the-art certifiably robust models encounter out-of-distribution (OOD) data.
Our analysis demonstrates a previously unknown vulnerability of these models to low-frequency OOD data such as weather-related corruptions, rendering these models unfit for deployment in the wild.
To alleviate this issue, we propose a novel data augmentation scheme, \emph{FourierMix}, that produces augmentations to improve the spectral coverage of the training data.
Furthermore, we propose a new regularizer that encourages consistent predictions on noise perturbations of the augmented data to improve the quality of the smoothed models.
We find that \emph{FourierMix} augmentations help eliminate the spectral bias of certifiably robust models enabling them to achieve significantly better robustness guarantees on a range of OOD benchmarks.
Our evaluation also uncovers the inability of current OOD benchmarks at highlighting the spectral biases of the models.
To this end, we propose a comprehensive benchmarking suite that contains corruptions from different regions in the spectral domain. 
Evaluation of models trained with popular augmentation methods on the proposed suite highlights their spectral biases and establishes the superiority of \emph{FourierMix} trained models at achieving better-certified robustness guarantees under OOD shifts over the entire frequency spectrum.\footnote{Pre-print under review. Codes and benchmark datasets will be open-sourced upon paper acceptance.}

\end{abstract}

\vspace{-0.3cm}
\section{Introduction}
\label{sec:intro}

Developing machine learning (ML) systems that are robust to adversarial variations in the test data is critical for applied domains that require ML safety \cite{mlsafety}, such as autonomous driving and cyber-security. 
Unfortunately, a large body of work in this direction has fallen into the cycle where new empirical defense techniques are proposed, followed by new adaptive attacks breaking these defenses~\cite{athalye2018obfuscated,tramer2020adaptive}. 
Therefore, significant efforts have been dedicated to developing methods that are certifiably robust \cite{wong2018provable,gowal2018effectiveness,raghunathan2018certified} which provide provable robustness guarantees. 
Most promising among these certified defenses are \emph{randomized smoothing (RS)} based certified defenses \cite{li2018second,lecuyer2019certified,cohen2019certified} which are scalable to deep neural networks (DNNs) and high-dimensional datasets.
Specifically, the RS-based certification procedure relies on a smoothed version of the original classifier, which outputs the class most likely returned by the original classifier under random noise perturbations of the input. 
Prediction from the RS procedure at the test time is accompanied by a \emph{radius} in which the predictions of the smoothed classifier are guaranteed to remain constant, thereby making them resilient to adversarial attacks within the neighborhood.
Training methods such as \cite{cohen2019certified,Zhai2020MACER,salman2019provably} have been proposed to maximize the \emph{average certified radius (ACR)}, and models trained using these procedures achieve state-of-the-art (SOTA) adversarial robustness guarantees, all while assuming that the test data is identically distributed to the training data.
In this work, we take a critical look at the current status of certifiably robust ML and consider whether these certifiably robust models are ready for deployment in the real world.

\begin{figure}

\includegraphics[width=\linewidth]{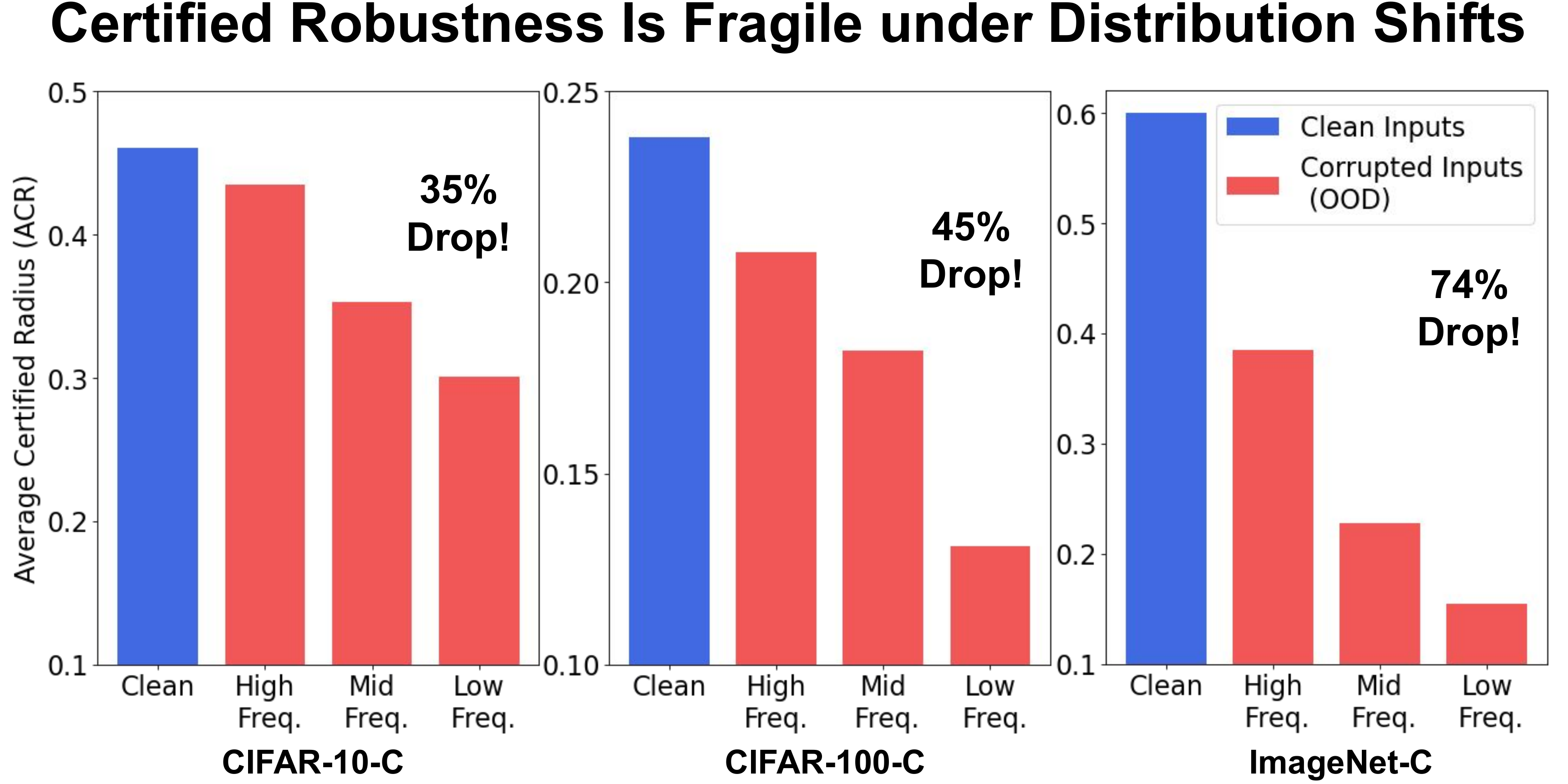}
\vspace{-0.5cm}
\caption{Robustness guarantees of certified models~\cite{cohen2019certified} degrade significantly on out-of-distribution data, suggesting that these models are not yet ready for deployment in the real world.}\label{fig:teaser}
\vspace{-0.5cm}
\end{figure}

Our work takes the first steps towards answering this question by evaluating RS-based provably robust ML models on out-of-distribution (OOD) shifts, as mismatches between the training and deployment distributions are ubiquitous in the real world. 
Our analysis shows that
\noindent \textbf{OOD data pose a serious threat to certifiably robust models.}
We therefore highlight a previously unrecognized threat to certifiably robust models and thereby show that these models are not yet ready for deployment in the real world.
Specifically, we found state-of-the-art certifiably robust models to be surprisingly brittle to low-frequency perturbations, such as weather-related corruptions (\eg fog and frost). 
Vulnerability to such corruptions could lead to a detrimental performance of ML models on safety-critical applications.  
For example, $35\%$--$75\%$ performance drop is observed on low-frequency corruptions rendering RS-based robustness guarantees useless (Figure~\ref{fig:teaser}).

Motivated from our analysis, which shows RS-based smoothed classifiers to be brittle to low-frequency corruptions, we propose a novel data augmentation method that uses \textbf{spectrally diverse yet semantically consistent augmentations} of the training data. 
Specifically, our proposed data augmentation method \emph{FourierMix} generates augmented data samples by using Fourier-based transformations on the input data to increase the spectral coverage of the training data. 
\emph{FourierMix} randomly perturbs the amplitude and phase of the images in the training data and then combines them with the affine transformation of the data, producing spectrally diverse augmentations.
To encourage the model to produce consistent predictions on different augmentations of the data, we propose a \emph{hierarchical consistency regularizer (HCR)}.
The use of HCR as the regularizer leads to semantic consistency of representations across random noise perturbations as well as FourierMix generated augmentations of the same input image. 
The proposed scheme consistently achieves significantly better certified robustness guarantees as compared to existing state-of-the-art data augmentation schemes extended to build a smoothed classifier, across a range of OOD benchmarks. 
We further analyze these smooth models using 
Fourier sensitivity analysis in the spectral domain. In comparison to other methods, models trained on \emph{FourierMix} augmentations coupled with hierarchical consistency regularization are significantly more resilient to perturbations across the entire frequency spectrum. 

Our empirical evaluation of certifiably robust models on various OOD benchmark datasets uncovers another peculiar phenomenon--\textbf{popular benchmark datasets may be biased towards certain frequency regions}.
Due to the complexity of real-world data, it is extremely challenging and tedious to uncover the spectral biases of the models and to identify their failure modes.
Because of this, improvements in the performance of the models on these benchmark datasets may not generalize to other OOD scenarios. 
Thus, we should be cautious and avoid over-reliance on a specific leaderboard, especially to judge the robustness of models on OOD data. 
To enable the designers to understand the spectral biases of their models and obtain a more comprehensive view of the model robustness to OOD data, we propose a complementary new benchmark that includes a collection of OOD test sets, each focusing on specific frequency ranges while collectively covering the entire frequency spectrum.
Evaluation of the certified robustness of different models on the proposed dataset shows that the smooth models obtained after training with existing data augmentation schemes are indeed biased towards certain frequencies regions.
This justifies the observed performance (and ranking) variations across different benchmarks. 
On the other hand, models trained with our \emph{FourierMix} based data augmentations perform significantly better than the competitors across the entire frequency spectrum.
This further demonstrates that our data augmentation can produce spectrally diverse data which help alleviate the biases of the models to different frequencies. Our main contributions are as follows: 
\vspace{-0.15cm}
\begin{itemize}[leftmargin=*] 
\setlength{\itemsep}{1pt}
\setlength{\parskip}{2pt}
\item We highlight a previously unknown vulnerability of certified adversarial defenses to OOD data and show their failure on mid/low-frequency perturbations of the data.
\item We present a novel data augmentation scheme \emph{FourierMix} that produces spectrally diverse augmentations of the training data combined with a new consistency regularization term. Our evaluation shows that they significantly enhance the robustness of certifiably robust models to OOD data (\eg $18.3\%$/$26.3\%$ average improvements on CIFAR-10/100-based OOD benchmarks compared to other baselines).
\item We propose a novel OOD benchmark suite providing broad spectral coverage of the corruptions that may be encountered at test time. The dataset can help probe the spectral biases of the models, which may not be unveiled by evaluation on existing OOD benchmark datasets.
\end{itemize}

\section{Related Work}
\label{sec:related}
Deep neural networks (DNNs) trained using standard gradient descent optimizers~\cite{ruder2016overview} have been shown vulnerable to adversarial examples~\cite{szegedy2013intriguing}. 
A number of white- and black-box attacks have been proposed~\cite{7958570,chen2017ead,xiao2018generating,CPY17zoo,ilyas2018black} to construct adversarial examples with small $\ell_{p}$ distances to the original data that mislead these DNN models.
Besides adversarial attacks, recent studies have devoted efforts to characterizing model performance under out-of-distribution (OOD) shifts~\cite{hendrycks2019benchmarking, bulusu2020anomalous}, where natural corruptions lead to a significant impact on the accuracy of SOTA ML models. 
Thus, it has become imperative to study how ML models can be made robust to test data coming from different distributions when the models are deployed in the real world.

\noindent\textbf{Certified Robustness and Defenses.}\quad The authors in~\cite{szegedy2013intriguing} have discovered the adversarial examples in DNN models, after which many defenses have been presented to mitigate such vulnerability~\cite{athalye2018obfuscated}. However, many of the proposed countermeasures have been shown to rely on gradient obfuscation, limiting malicious agents from accessing the accurate gradients. Such defenses are vulnerable to adaptive attacks, which give a false sense of security~\cite{athalye2018obfuscated} of the models. Certified defenses are thus highly desirable. Along with a prediction on the test point, these defenses output a certified radius $r$ such that for any $||\vdelta||_2<r$, the model continues to have the same prediction. Such techniques include convex polytope\cite{wong2018provable}, recursive propagation\cite{gowal2018effectiveness}, and linear relaxation\cite{raghunathan2018certified,zhang2018efficient}. These methods provide a lower bound on the perturbation required to change the model's prediction on a target point. However, such methods can merely be applied to shallow models, which limits their practicality. Recently,~\cite{li2018second,lecuyer2019certified,cohen2019certified,mohapatra2020higher} have proposed randomized smoothing (RS)-based certified defenses that produce better lower bounds and are scalable to large networks. In this paper, we study the OOD robustness of such certified defenses. Unlike a recent work~\cite{mehra2021robust}, which uses data poisoning attacks to hurt the robustness guarantees of the RS-based models, our work demonstrates the failure of these models on test-time corruptions, which might be encountered by the model deployed in the real world.


\noindent\textbf{Robustness against Common Corruptions -- Benchmarks and Defenses.}\quad
Pioneering studies have identified vulnerabilities of deep learning models to common corruptions. 
Dodge~\etal find that standard trained DNNs are vulnerable to blur and Gaussian noise~\cite{dodge2016understanding}. 
Hendrycks \etal~\cite{hendrycks2019benchmarking} present CIFAR-10/100-C and ImageNet-C, consisting of fifteen different common corruptions with five severity levels to facilitate robustness evaluations of CIFAR~\cite{krizhevsky2009learning} and ImageNet~\cite{deng2009imagenet} models. 
Recently, Mintum~\etal further propose CIFAR-10/100-$\bar{\text{C}}$ and ImageNet-$\bar{\text{C}}$ to provide  new corruptions~\cite{mintun2021interaction}. 
There are two popular lines of work on improving the robustness against common corruptions: \textit{test-time adaptation}~\cite{schneider2020improving} and \textit{data augmentation}~\cite{cubuk2019autoaugment,hendrycks2019augmix}.
The authors in \cite{saenko2010adapting} propose a method to update the batch normalization (BN) statistics for improving domain adaptation. Another recent method, TENT\cite{wang2021tent} updates both the affine transformation and statistics of BN by using self-entropy minimization. On the other hand, methods such as AutoAugment~\cite{cubuk2019autoaugment} leverages reinforcement learning to learn an augmentation policy that produces a diverse set of augmentations to help make the models robust to OOD data. Another popular method, AugMix~\cite{hendrycks2019augmix} achieves impressive performance improvement on corrupted data using augmentations generated by mixing up images obtained from applying randomly sampled operations along with using a Jenson-Shannon based consistency loss during training. Unlike existing data augmentation schemes which intend to improve the empirical robust accuracy of the models, the data augmentation schemes of interest to this paper aim to improve the adversarial robustness guarantees on OOD data.

\noindent\textbf{Certified Semantic Robustness.}\quad Recent work \cite{mohapatra2020towards,fischer2020certified,li2021tss} have also focused on developing techniques to provide performance guarantees to seen (or known) common corruption types (such as rotation or brightness changes). However, in this work, we are interested in more realistic scenarios with unseen (or unknown) test-time corruptions. It is worth noting that the susceptibility analysis and defense techniques developed in this work can be extended to SOTA semantic robustness techniques.

\section{Are Certifiably Robust Models Ready for Deployment in the Wild?}
\label{sec:rs_on_c}

\begin{figure*}[t]
\begin{minipage}[]{0.485\linewidth}
\begin{center}
\includegraphics[width=\linewidth]{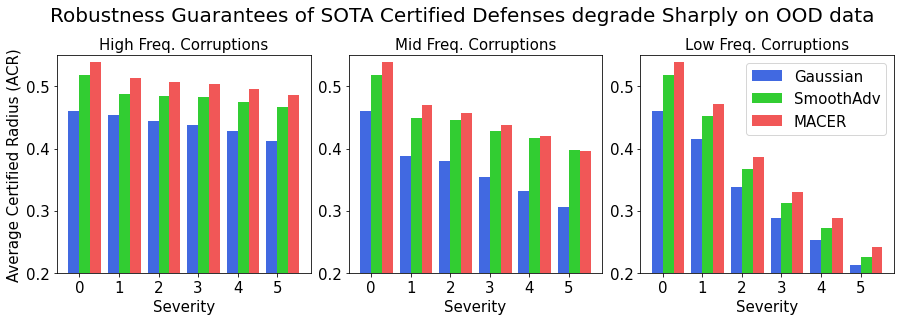} 
\end{center}
\vspace{-0.3cm}
\caption{Randomized smoothing based models~\cite{cohen2019certified,salman2019provably,Zhai2020MACER} suffer up to 56\% decreases in their certified robustness on \textbf{mid-to-low} frequency corruptions from CIFAR-10-C. Severity 0 is in-distribution. 
}
\label{fig:motivation}
\end{minipage}
\hfill
\begin{minipage}[]{0.485\linewidth}
\vspace{-0.1cm}
\subfigure[Gaussian \label{fig:}]{\begin{minipage}[t]{0.325\linewidth}
\begin{center}
\includegraphics[width=\linewidth]{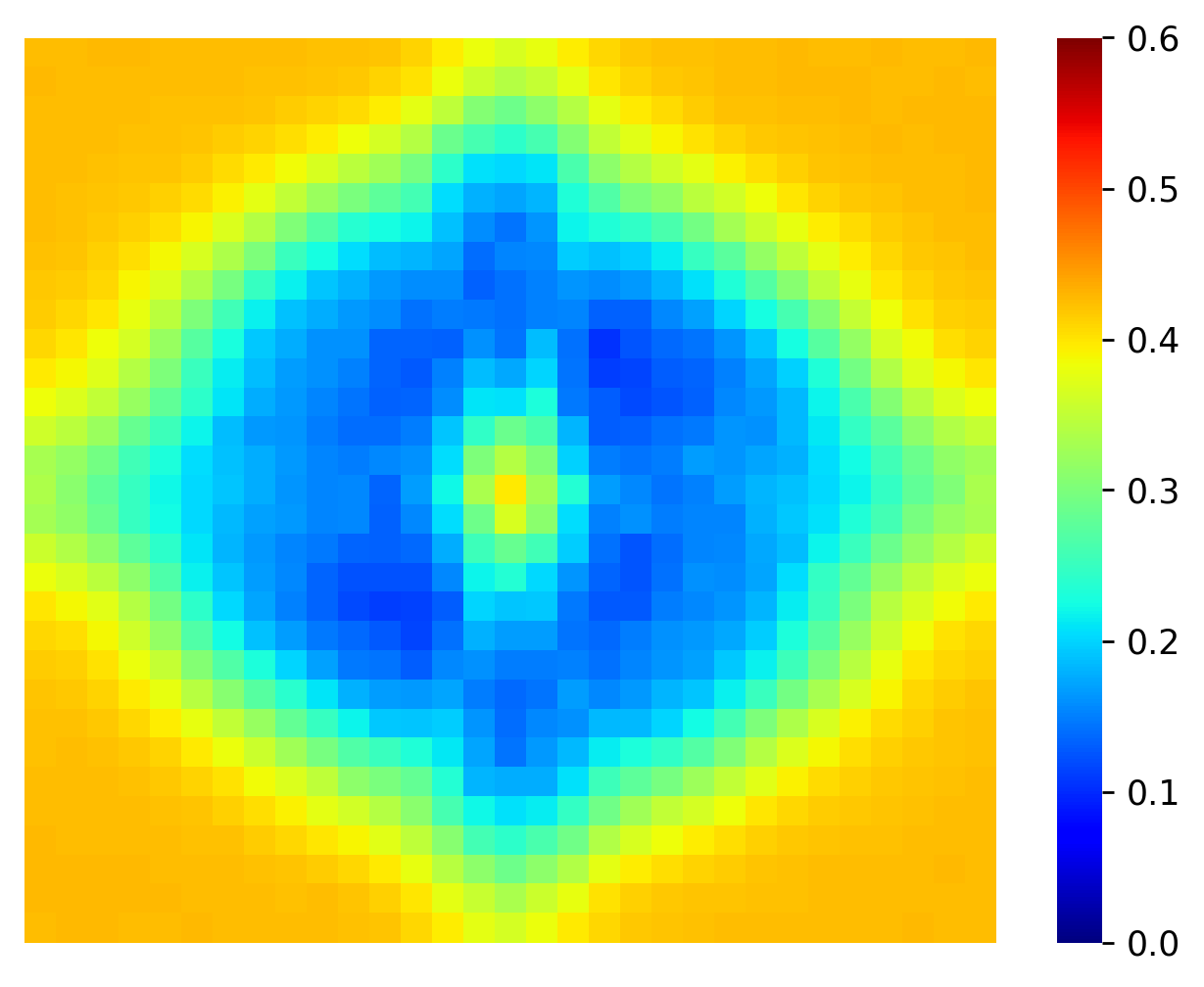} 
\end{center}
\end{minipage}}
\subfigure[SmoothAdv \label{fig:}]{\begin{minipage}[t]{0.325\linewidth}
\begin{center}
\includegraphics[width=\linewidth]{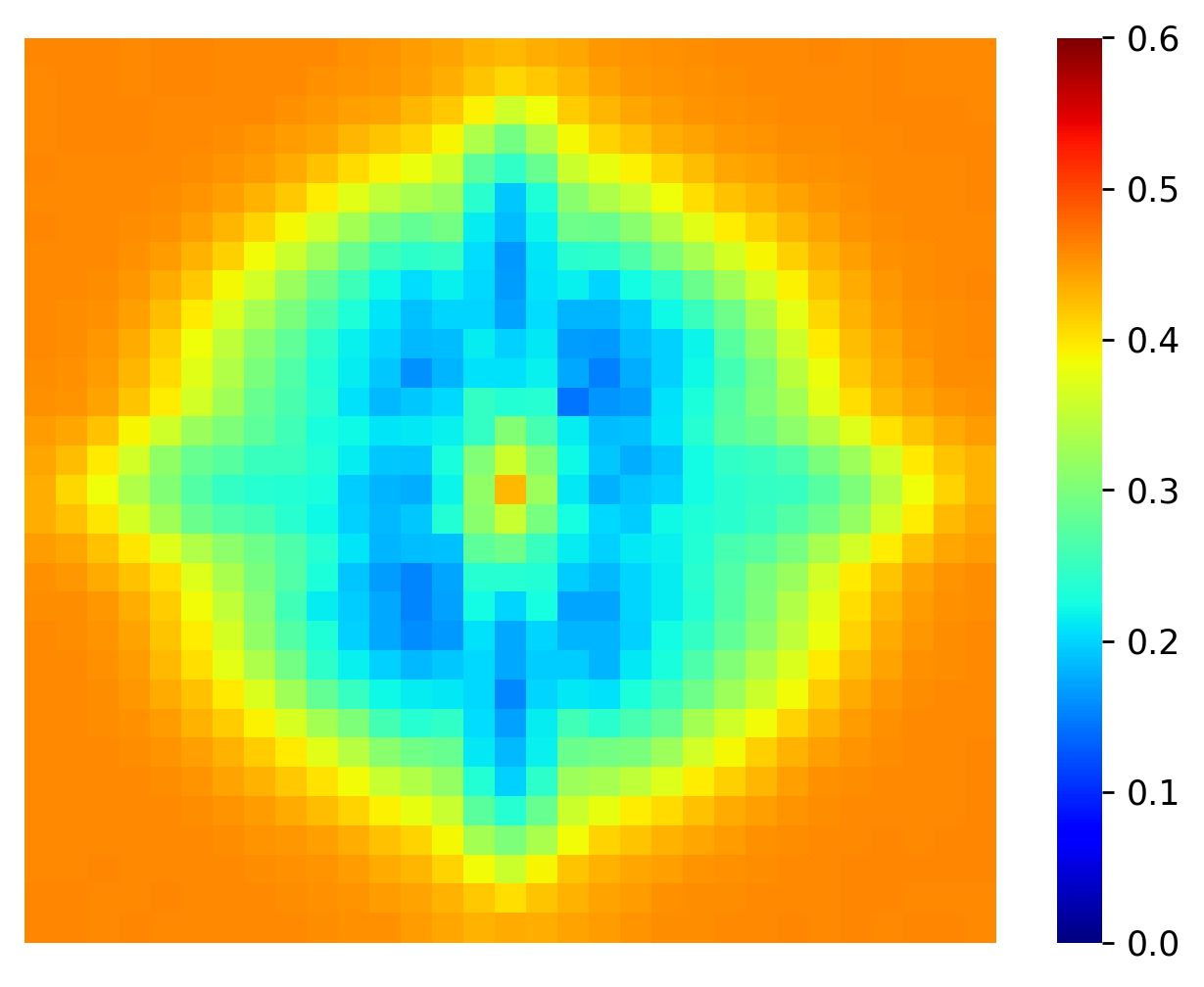} 
\end{center}
\end{minipage}}
\subfigure[MACER \label{fig:}]{\begin{minipage}[t]{0.325\linewidth}
\begin{center}
\includegraphics[width=\linewidth]{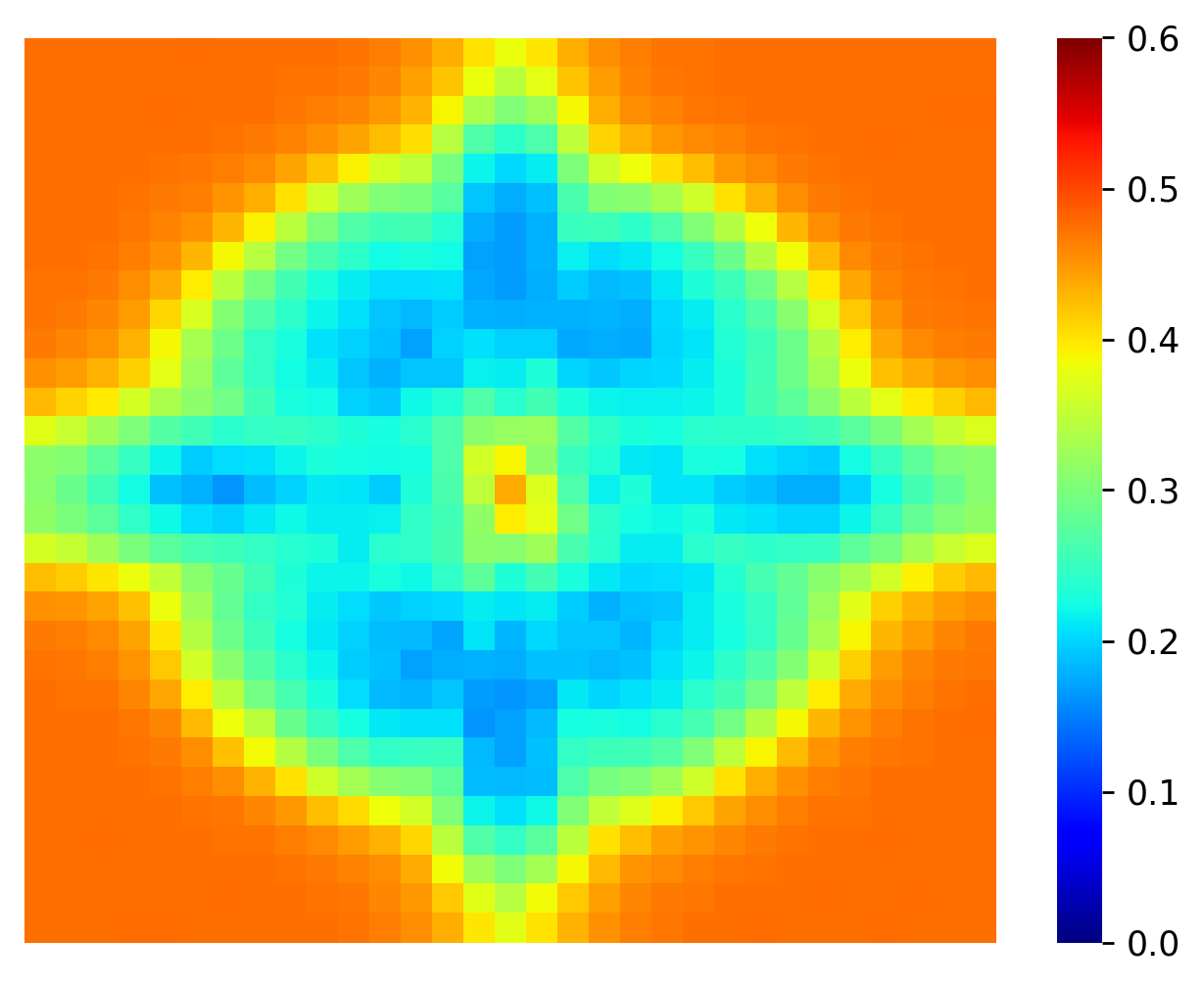} 
\end{center}
\end{minipage}}
\vspace{-0.2cm}
\caption{Fourier sensitivity analysis on CIFAR-10 shows the certified robustness guarantees (ACR) of SOTA certified defenses degrade significantly on OOD data from mid-to-low frequency region (interpreted in \S~\ref{sec:2.2.2}).}
\label{fig:fourier_basis1}
\end{minipage}

\vspace{-0.5cm}
\end{figure*}

Predictions of certifiably robust ML models are guaranteed to stay constant in a neighborhood of a test point, making them provably resilient to adversaries at the test time. This feature of certified defenses makes them an attractive candidate for real world safety-critical applications. However, progress in this area has been assessed by evaluating these models in idealistic scenarios (\ie the in-distribution setup), which is not representative of real world data distributions. To better understand the performance of certified defenses in the real world, in this section, we evaluate SOTA certified defenses against OOD shifts.


\subsection{Background on SOTA Certified Defenses}
\label{sec:metrics}
We focus our attention on the SOTA certification technique based on randomized smoothing (RS) which is efficient and scalable.
Let us consider a base classifier $\mathcal{M}$ trained on samples $\vx \in \mathcal{X} \subset \sR^{d\times d\times 3}$ and their corresponding labels $y \in \mathcal{Y} \subset \sR^{+}$, obtained from an underlying data distribution $\mathcal{D}$. 

\noindent\textbf{Certification.}\quad The RS-based certification uses a base classifier $\mathcal{M}$ and provides certified robustness guarantees for its smoothed version defined as $\hat{\mathcal{M}}(\vx) = \argmax_{c \in \mathcal{Y}}\sP(\mathcal{M}(\vx+\vdelta)=c)$ where $\vdelta \sim \mathcal{N}(0,\sigma^{2}\mathbf{I})$. 
Intuitively, $\hat{\mathcal{M}}$ returns the most probable class returned by $\mathcal{M}$ on Gaussian perturbations of the input $\vx$.
The certification guarantees that the prediction of the smoothed classifier $\hat{\mathcal{M}}$ are consistent in the $\ell_2$ radius \cite{cohen2019certified} of 
$\text{CR}(\hat{\mathcal{M}},\sigma,\vx;y) = \frac{\sigma}{2}(\Phi^{-1}(p_A)-\Phi^{-1}(p_B))$,
where $\Phi^{-1}$ is the inverse CDF of the standard Gaussian distribution, $p_A=\sP(\mathcal{M}(\vx+\vdelta)=c_{A})$ is probability of the top class ($c_A$) and $p_B=\max_{c \neq c_{A}} \sP(\mathcal{M}(\vx+\vdelta)=c_{A})$ is the probability of the runner-up class.
Monte Carlo-based sampling~\cite{hammersley2013monte} is utilized to approximate $\underline{p_A} \leq p_A$ and $\overline{p_B} = 1-\underline{p_A} \geq p_B$. The certified radius can still be computed using the same formula by replacing  $p_A$ and $p_B$ with $\underline{p_A}$ and $\overline{p_B}$.

\noindent\textbf{Improved Training.}\quad It has been observed empirically~\cite{cohen2019certified} that models trained using the standard training procedure do not provide reasonable certified robustness. Therefore, there is an increasing interest in developing improved training techniques to maximize the certified robustness. Several works \cite{li2020sok} have made significant advances on the training techniques and reported impressive gains in terms of certified radius on in-distribution test data. 
Specifically, new training methods such as Gaussian augmentation~\cite{cohen2019certified}, SmoothAdv~\cite{salman2019provably} and MACER~\cite{Zhai2020MACER} have been proposed. Intuitively, Cohen~\etal~\cite{cohen2019certified} propose to leverage Gaussian augmentation with variance $\sigma^2$ to train the base classifier. SmoothAdv~\cite{salman2019provably} and MACER~\cite{Zhai2020MACER} both use Gaussian augmentation and further improve Cohen~\etal's baseline method by adversarial training and introducing an auxiliary objective to maximize the certified radius, respectively. 
However, the effect of OOD data on the robustness guarantees of these models has been unexplored in the literature. 

\noindent\textbf{Evaluation Metrics.}\quad Similar to previous works~\cite{Zhai2020MACER, mehra2021robust,salman2019provably}, we use the \textit{average certified radius} (ACR) as our metric to evaluate the robustness of the models on in-distribution (or clean) test data. Specifically, $\text{ACR}:=\frac{1}{|\mathcal{D}_{test}|}\sum_{(\vx,y)\in \mathcal{D}_{test}} \text{CR}(\hat{\mathcal{M}},\sigma,\vx;y)\times\textbf{1}_{\hat{\mathcal{M}}(\vx,\sigma)=y}$, which is also equivalent to the area under the certified radius-accuracy curve. 
We assign $\text{CR}(\cdot)=0$ for incorrect prediction of $\hat{\mathcal{M}}$. For OOD performance, we measure the mean ACR (mACR) as an overall metric, $\text{mACR} := \frac{1}{c}\sum_{i=1}^{c}{\text{ACR}_{i}}$, where $c$ is the number of corruptions leveraged in a specific test set. For example, $c=15$ and $10$ in CIFAR-10/100-C and -$\overline{\text{C}}$ datasets, respectively. We also report the ACR for each corruption type. Unlike previous studies on empirical defenses, we do not use the \textit{empirical} clean and robust accuracy~\cite{cohen2019certified,salman2019provably,Zhai2020MACER} as a metric in this work since we focus on the \textit{certified} robustness.

\subsection{Analyzing Certified Defenses in OOD Settings}

Real-world test data oftentimes do not follow the training data distribution $\mathcal{D}$, although tangible improvements have been made on certifying robustness of in-distribution data.
Therefore, evaluating the performance of $\mathcal{M}$ on out-of-distribution (OOD) data $\{(\hat{\vx},y)_1,...,(\hat{\vx},y)_n\} \sim \hat{\mathcal{D}}$ becomes a major concern. 
We consider the impact of OOD data on models trained using SOTA robust training methods \cite{cohen2019certified,salman2019provably,Zhai2020MACER} and RS-based certified defenses.

\subsubsection{Degradation of Certified Robustness Guarantees on Common Corruptions} 

To measure the performance of certified defenses on OOD data, we use the popular common corruptions dataset CIFAR-10-C \cite{hendrycks2019benchmarking}. CIFAR-10-C contains 15 different corruptions from four categories (with 5 severity levels): noise, blur, weather, and digital corruptions. We re-arrange the corruption dataset into three groups and evaluate the ACR by increasing the severity level of the corruptions. Grouping is performed based on the visual similarity of the amplitude spectrum of corrupted images (see Appendix~\ref{psd_cifar-10-c}).
Group-H corruptions (roughly categorized as high-frequency corruption type) consist of \{\texttt{Gaussian noise}, \texttt{impulse noise}, \texttt{shot noise}, \texttt{pixelate}, \texttt{JPEG}\}; Group-M corruptions (roughly categorized as mid-frequency corruption type) consist of \{\texttt{defocus blur}, \texttt{frosted glass blur}, \texttt{motion blur}, \texttt{zoom blur}, \texttt{elastic}\}; and Group-L corruptions (roughly categorized as low-frequency corruption type) consist of \{\texttt{brightness}, \texttt{fog}, \texttt{frost}, \texttt{snow}, \texttt{contrast}\}. 


The performance of SOTA certified defenses on these groups of corruptions is presented in Figure~\ref{fig:motivation}. SmoothAdv and MACER both achieve tangible enhancements in ACR on in-distribution CIFAR-10 data compared to the Gaussian augmentation baseline. However, all methods show a sharp performance drop in ACR as we move from Group-H (high-frequency) to Group-L (low-frequency). We see that these methods are surprisingly brittle in low-frequency corruption regime, \eg we see up to 54\% drop in ACR when moving from severity 0 (\ie in-distribution) to severity 5. We emphasize that this performance drop points to a methodological shortcoming and is not due to the corruptions in Group-L being too difficult since the empirical robust accuracy (Figure~\ref{fig:motivation2} in Appendix~\ref{app:robustbench}) remains consistently high on all the groups and severity levels for empirically robust models~\cite{hendrycks2019augmix,kireev2021effectiveness,rebuffi2021fixing}. 
Even though the performance of any ML model is expected to suffer on test data that lies far away from the data used during training, the drastic performance degradation of RS-based certifiably robust models on low-frequency corruptions is particularly concerning.


\subsubsection{Validating the Brittleness of Smoothed Models Through a Spectral Lens}
\label{sec:2.2.2}


To highlight that the vulnerability to low-frequency corruptions is a limitation of provably robust ML models, in this section, we perform a more systematic analysis that corroborates that our finding is not limited to a specific benchmark and holds more broadly.
To achieve this, we perform a spectral domain analysis of SOTA smoothed models by utilizing the Fourier sensitivity analysis~\cite{yin2019fourier}, which we briefly summarize next.

A Fourier basis image in the pixel space is a real-valued matrix $\mU_{i,j} \in \sR^{d\times d}$ where its $||\mU_{i,j}||_2 = 1$, and $\text{FFT}(\mU_{i,j})$ only has two non-zero elements at $(i,j)$ and $(-i,-j)$ in the coordinate that views the image center as the origin. 
Given a test set and a smoothed model, we evaluate the CR($\cdot$) of $\widetilde{\vx}_{i,j} = \vx + r\epsilon \mU_{i,j}$ for each $\vx$ in the test set and compute their ACR, where $r$ is randomly sampled in $\{-1,1\}$, $\epsilon$ is the perturbation in $\ell_2$ norm, and we treat the RGB channels independently. Each of the evaluated ACR corresponds to a data point in the heat map located at $(i,j)$.
Figure~\ref{fig:fourier_basis1} shows the heatmaps of models trained with Gaussian augmentation\cite{cohen2019certified}, SmoothAdv\cite{salman2019provably}, and MACER\cite{Zhai2020MACER} using $\epsilon=4$~\cite{yin2019fourier}. 
The center and edges of the heatmap contain evaluation on the lowest and highest frequency perturbations, respectively. 
The results in Figure~\ref{fig:fourier_basis1} show that the certifiably robust classifiers achieve small ACR on OOD data belonging to the low-frequency region (around the center of the image) whereas they achieve a high ACR in the high-frequency region (near the edges).
In particular, the ACRs are always less than 0.3 for all three methods in the mid-to-low frequency range, while they perform well in high-frequency regime.
We emphasize that the Fourier sensitivity analysis in Figure~\ref{fig:fourier_basis1} is general and is not specific to corruptions appearing in CIFAR10-C. 
Based on our analysis, we find that certifiably robust models are biased towards high-frequency noises and perform surprisingly poor on low-frequency OOD data.

\begin{figure*}[t]
\begin{center}
\vspace{-0.1cm}
\includegraphics[width=\linewidth]{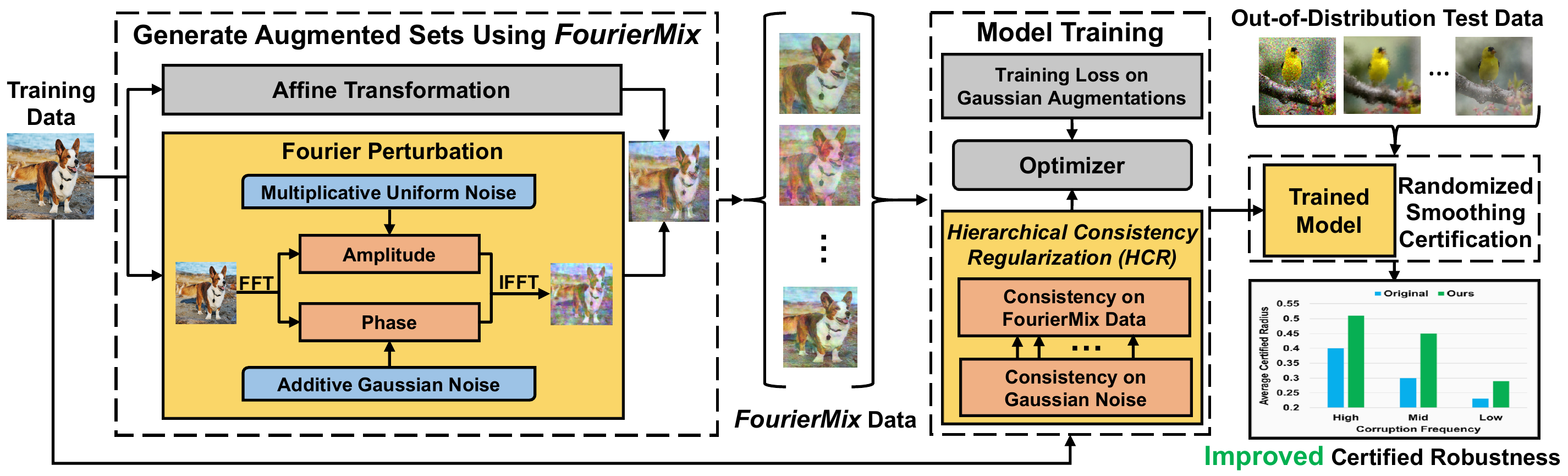} 
\vspace{-0.4cm}
\caption{Overview of our \emph{FourierMix} pipeline for generating spectrally diverse data augmentations and training of certifiably robust models with the proposed hierarchical consistency regularization (HCR). The bottom-right figure shows the significant improvement achieved by our method (in terms of ACR) in comparision to Cohen~\etal~\cite{cohen2019certified} on severity 5 corruptions from different frequencies in CIFAR-10-C.
}
\label{fig:pipeline}
\end{center}
\vspace{-0.8cm}
\end{figure*}

Following this insight, we develop a data augmentation method capable of producing spectrally diverse augmentations to make certifiably robust models perform well on OOD data across the entire frequency spectrum in \S~\ref{sec:fouriermix}.

\SetKwComment{Comment}{$\triangleright$}{}
\begin{algorithm}[t]
\footnotesize
\caption{\emph{FourierMix} Pseudocode}\label{alg:fmix}
\KwData{Model $\mathcal{M}$, Image $\vx_\text{orig}$, Affine Transformation $\mathbf{T}$, Fourier Amplitude $\mathbf{A}$ and Phase $\mathbf{P}$ Operations
}
\KwResult{$\vx_\text{aug}$ $=$ \emph{FourierMix}($\vx_\text{orig}$, ~$k=3$, ~$\alpha=1$)}
\DontPrintSemicolon
$\vx_\text{aug} = 0$\;
Sample mixing weights ($w_1, ..., w_k$) $\sim$ Dirichlet($\alpha,...,\alpha$)\;
\For{$i = 1,2,...,k$}{
  Sample random affine transformation $\mathbf{T}_i$\;
  \Comment*[l]{Fast Fourier Transformation}
  $\vx_\text{fourier} = \text{FFT}(\vx_\text{orig})$\;
  Sample severity level of operations
  $s_{\mathbf{A}},s_{\mathbf{P}}$\;
  \Comment*[l]{Fourier Perturbation}
  $\vx_\text{fourier} = (\mathbf{A}_{s_{\mathbf{A}}} \circ \mathbf{P}_{s_{\mathbf{P}}}) (\vx_\text{fourier})$\;
  \Comment*[l]{Inverse Fast Fourier Transformation}
  $\vx_{f} = \text{IFFT}(\vx_\text{fourier})$\; 
  Sample weight $t \sim \text{Beta} (\alpha,\alpha)$ \;
  $\vx_\text{aug} \;+= w_i \cdot (t\vx_{f} + (1-t)\mathbf{T}_i^{\top}\cdot\vx_\text{orig})$\; 
  }
Sample weight $m \sim \text{Beta} (\alpha,\alpha)$ \;
$\vx_\text{aug} = m \vx_\text{orig} + (1-m)\vx_\text{aug}$\;
\end{algorithm}
\setlength{\textfloatsep}{0pt}
\vspace{-0.3cm}
\section{\emph{FourierMix}: Data Augmentation with Broad Spectral Coverage}
\label{sec:fouriermix}

To improve the certified robustness of RS-based methods on OOD data, it is intuitively desirable to make the base classifier $\mathcal{M}$ robust against different types of corruptions and their Gaussian perturbations. Motivated by our Fourier sensitivity analysis (\S~\ref{sec:rs_on_c}), we propose a novel data augmentation method, \emph{FourierMix}, to boost the certified robustness performance on OOD data\footnote{As Gaussian augmentation is fundamental to RS-based certified defenses, we focus on improving the performance of Gaussian augmentation based defenses under OOD corruptions. Further ACR gains can be achieved by leveraging the techniques proposed in this paper along with more advanced SOTA certified defenses such as SmoothAdv and MACER.}.
To improve the spectral coverage, we introduce Fourier-based operations that manipulate the image in the frequency domain. We also leverage randomly sampled affine transformations to enrich the augmentations in \emph{FourierMix}. We adopt the high-level framework of AugMix~\cite{hendrycks2019augmix} for chaining and mixing different augmented images. Figure~\ref{fig:pipeline} shows the overall pipeline and Algorithm~\ref{alg:fmix} presents the pseudocode of \emph{FourierMix}.

\noindent \textbf{Fourier Operations.}\quad 
Two dimensional images can be converted into the frequency domain by applying the Fourier transform and vice versa. Fourier transform has the \textit{duality} property, which provides a unique but equivalent perspective for image analysis. 
We use fast Fourier transform (FFT) and inverse FFT (IFFT) for the transformation between the pixel and frequency domains. 
$\text{FFT}(\vx)$ is complex in general, \ie $\text{FFT}(\vx)= \text{FFT}_{real}(\vx)+i\text{FFT}_{imag}(\vx)$, with $\mA = |\text{FFT}(\vx)|$ as its amplitude and $\mP = \arctan(\text{FFT}_{imag}(\vx)/\text{FFT}_{real}(\vx))$ as its phase.
The amplitude spectrum of natural images generally follows a power-law distribution, \ie $\frac{1}{f^{\alpha}}$, where $f$ is the azimuthal frequency and $\alpha \approx 2$~\cite{burton1987color,tolhurst1992amplitude}, resulting in extremely small power in the high-frequency areas. However, the amplitude spectrum of the IID Gaussian noise is a uniform distribution, so Gaussian augmentation biases the models towards the high-frequency regime relative to original images. In order to have broad and unbiased spectral coverage, the core of \emph{FourierMix} is to allocate similar proportions of augmentations across all frequencies. We use two spectral perturbation methods in \emph{FourierMix} to achieve this goal:
\begin{equation}
\small
\mathbf{A}(u,v) = \mA_{u,v}^\text{orig} \cdot \text{U}(1-s_{\mathbf{A}},1+s_{\mathbf{A}}) 
\label{eq:fouriermix1}
\end{equation}
\begin{equation}
\small
\mathbf{P}(u,v) = \mP_{u,v}^\text{orig}+ \mathcal{N}_\text{truncated}^{s_{\mathbf{P}}}(0,\sigma^2\mathbf{I})
\label{eq:fouriermix2}
\end{equation}
where $(u,v)$ is the coordinate of the 2D frequency in the spectrum, and $s_{\mathbf{A}}$ and $s_{\mathbf{P}}$ control the severity levels of two perturbations. Formally, the PDF of  $\mathcal{N}_\text{truncated}^{s_{\mathbf{P}}}=\frac{\phi(x/\sigma)}{\sigma \cdot (2\Phi(s_{\mathbf{P}}/\sigma)-1)}$, where $\phi(\cdot)$ and $\Phi(\cdot)$ denote the PDF and CDF functions of a standard normal distribution, respectively. On one hand, we apply multiplicative factors sampled from a uniform distribution $\text{U}(\cdot)$ to all frequencies in the amplitude spectrum. Therefore, $\mathbf{A}(u,v)$ ensures that the proportions of augmentation are similar across all frequencies relative to the original spectrum. On the other hand, since the magnitude of a phase spectrum is not correlated with the 2D frequency~\cite{lim1990two}, additive noises are able to assign similar proportions of augmentations across 2D frequencies. As it is widely acknowledged that the phase component retains most of the high-level semantics~\cite{xu2021fourier,yang2020phase,kermisch1970image}, we leverage additive truncated Gaussian to constrain $\mathbf{P}(u,v)$ so that it will not destroy the semantics of the training images. Some sample images generated using \emph{FourierMix} are provided in Appendix~\ref{app:fmix-images}.


\noindent \textbf{Hierarchical Consistency Regularization (HCR).} Motivated from ~\cite{jeong2020consistency} that enforces consistency on in-distribution data, 
we propose \textit{hierarchical consistency regularization} (HCR) to further boost the performance of \emph{FourierMix} in terms of the ACR on OOD test sets:
\begin{equation}
\small
\mathcal{L}_{G} = \frac{1}{s}\sum_{i=0}^s \text{KL}(\mathcal{M}(\vx_j+\vdelta_i)\| \overline{\mathcal{M}}(\vx_j,\vdelta))
\end{equation}
\begin{equation}
\small
\mathcal{L}_{HCR} = \frac{1}{k+1}\sum_{j=0}^k \Big[\lambda \cdot \text{KL}(\overline{\mathcal{M}}(\vx_j,\vdelta)\| \overline{\mathcal{M}}(\vx,\vdelta)) + \eta \cdot \mathcal{L}_{G} \Big]
\label{eq:hcr}
\end{equation}
where $\overline{\mathcal{M}}(\vx,\vdelta)=\mathbb{E}_{j\in\{0,1,...,k\}}[\overline{\mathcal{M}}(\vx_j,\vdelta)]$, $\overline{\mathcal{M}}(\vx_j, \vdelta) = \mathbb{E}_{i\in\{1,2,...,s\}}[\mathcal{M}(\vx_j+\vdelta_i)]$, $\vx_0$ is the original training image, and $\text{KL}(\cdot\|\cdot)$ denotes the Kullback–Leibler divergence (KLD)~\cite{Joyce2011}. We use $k=2$ and $s=2$ for the \emph{FourierMix} and Gaussian augmentation with $\delta_i = \mathcal{N}(0, \sigma^2 \mathbf{I})$, respectively. Since Jensen–Shannon divergence (JSD)~\cite{fuglede2004jensen} uses the KLD to calculate a normalized score that is symmetrical, HCR essentially stacks two levels of JSD while training the base classifier to enforce the consistent representations over both augmentations. The first level of consistency $\mathcal{L}_G$ is applied to the Gaussian augmentation, rendering the Gaussian perturbed neighbors of $\vx_{0,1,2}$ have similar outputs, and the second level of consistency is on the whole $(k+1)s$ set to enforce \emph{FourierMix} augmented images with consistent outputs. We utilize $\lambda$ and $\eta$ as hyper-parameters to tune the weights of two levels of consistency. The overall training loss is: $\mathcal{L} = \frac{1}{s}\sum_{i=1}^s\mathcal{L}(\vx_0+\vdelta_{i},y) + \mathcal{L}_{HCR}$.





\noindent \textbf{Comparison with AugMix.}\quad The key difference between \emph{FourierMix} and AugMix is the base augmentation operations used in the pipeline. AugMix leverages the operations from AutoAugment~\cite{cubuk2019autoaugment} that do not overlap with ImageNet-C. 
In contrast, the augmentations in \emph{FourierMix} utilize a simpler set of generic augmentations. 
We compare the performance (\ie ACR) of \emph{FourierMix} with AugMix on multiple OOD datasets in our evaluation (\S~\ref{sec:experiment} and \S~\ref{sec:cifarf}).

\begin{table*}[t]
\scriptsize
\renewcommand\arraystretch{1.}
\setlength\tabcolsep{2pt}
  \caption{\small Models trained with \emph{FourierMix} and HCR achieve significant improvements in the certified robustness (ACR and mACR) guarantees on all popular OOD datasets.  
  \textbf{Bold} and \underline{underline} denote the best and runner-up results, respectively. 
  }
  \label{tb:total}
  \vspace{-0.2cm}
  \centering
  \resizebox{1.03\textwidth}{!}{
  \begin{tabular}{|l||c|cccc|c||c|cccc|c||c|cccc|c|}
    \specialrule{1pt}{1.1pt}{1pt}
     &CIFAR-10 &\multicolumn{4}{c|}{CIFAR-10-C} &CIFAR-10-$\bar{\text{C}}$ &CIFAR-100 &\multicolumn{4}{c|}{CIFAR-100-C} &CIFAR-100-$\bar{\text{C}}$ &ImageNet & \multicolumn{4}{c|}{ImageNet-C} &ImageNet-$\bar{\text{C}}$ \\
    \hline
    Augmentation    &ACR & mACR &-Low &-Mid &-High  & mACR &ACR & mACR &-Low &-Mid &-High  & mACR &ACR & mACR &-Low &-Mid &-High  & mACR\\
\noalign{\global\arrayrulewidth1pt}\hline\noalign{\global\arrayrulewidth0.4pt}     
    Gaussian &0.461 & 0.363 &0.301 &0.353 &0.435 &0.314 &0.238 & 0.169 &0.131 &0.182 &0.208 &0.130 &0.600 & 0.256 &0.155 &0.228 &0.385 &0.266\\
     ~+JSD &\textbf{0.535} & 0.439 &0.346 &0.451 &\underline{0.520} &0.393 & 0.291 & 0.232 &0.167 &0.248 &0.280 &0.196 & 0.736 & 0.395 &0.220 &0.382 &\textbf{0.581} &0.395\\
    \hline
     ~+AutoAugment &0.411  & 0.372 &0.312 &0.364 &0.431 &0.304 & - & - & - & - & - & - & - & - & - & - & - &-\\
      ~~+JSD  &0.432 & 0.400 &0.343 &0.395 &0.464 &0.346 &0.265 & 0.225 &0.175 &0.234 &0.265 &0.176 & - & - & - & - & - & -\\
    \hline
     ~+AugMix &0.452 & 0.385 &0.324 &0.383 &0.449 &0.341 &- & - & - & - & - & -  & - & - & - & - & - & -\\
      ~~+JSD &0.518 & 0.430 &0.357 &0.436 &0.496 &0.382 &0.286 & 0.231 &0.184 &0.240 &0.269 &0.193 &0.717 & 0.391 &0.238 &\underline{0.387} &0.550 &0.379\\
      ~~+\textbf{HCR} &0.520 &0.437 &0.369 &0.444 &0.497 &0.393 &\underline{0.296}  &\underline{0.249} &\underline{0.191} &\underline{0.263} &\underline{0.292} &\underline{0.211} &0.727 &0.390 &0.234 &0.383 &0.552 &0.378\\
    \hline
     ~+\textbf{\emph{FourierMix}} &0.455 & 0.388 &0.326 &0.386 &0.453 &0.348  & - & - & - & - & - & - & -  & - & - & - & - & - \\
      ~~+JSD &\underline{0.522} & \underline{0.444} &\underline{0.375} &\underline{0.454} &0.504 &\underline{0.397} &0.295 & 0.247 &0.190 &0.258 &\underline{0.292} &0.207 &\textbf{0.751} &\textbf{0.399} &\textbf{0.242} &\textbf{0.389} &0.564 &\textbf{0.413}\\
      
      ~~+\textbf{HCR} &\textbf{0.535} & \textbf{0.460} &\textbf{0.384} &\textbf{0.473} &\textbf{0.521} &\textbf{0.419} &\textbf{0.309} &\textbf{0.261} &\textbf{0.199} &\textbf{0.278} &\textbf{0.307} &\textbf{0.227} & \underline{0.750} &\underline{0.397} &\underline{0.239} &\underline{0.387} &\underline{0.567} &\underline{0.411}\\
\noalign{\global\arrayrulewidth1pt}\hline\noalign{\global\arrayrulewidth0.4pt}
  \end{tabular}
  }
  \vspace{-0.3cm}
\end{table*}

\vspace{-0.2cm}
\section{Experiments on Popular OOD Benchmarks}
\label{sec:experiment}
\noindent \textbf{Experimental Setup.}\quad As introduced in \S~\ref{sec:metrics}, we use ACR and mACR as the main evaluation metrics. We utilize the official implementation 
from~\cite{cohen2019certified} to compute the certified radius $\text{CR}(\cdot)$. 
We use the same base architectures leveraged in prior arts~\cite{cohen2019certified,Zhai2020MACER,salman2019provably,jeong2020consistency}, \ie ResNet-110 and ResNet-50 as the backbones, for experiments on CIFAR-10/100 and ImageNet~\cite{he2016deep}, respectively.
We use Gaussian augmentation with $\sigma=0.25$ and $0.5$ for both training and certifying the CIFAR-10/100 and ImageNet models, respectively.
Further details on training are provided in Appendix~\ref{app:training-details}. 

\noindent \textbf{Baselines.}\quad We evaluate the certified robustness of models trained with following augmentations schemes on OOD data: Gaussian~\cite{cohen2019certified}, AutoAugment~\cite{cubuk2019autoaugment}, and AugMix~\cite{hendrycks2019augmix}. 
We also compare HCR with the the baseline JSD regularization~\cite{jeong2020consistency}. 
We follow Cohen~\etal~\cite{cohen2019certified} and Jeong~\etal~\cite{jeong2020consistency} to train the Gaussian and Gaussian+JSD baseline models, respectively. 
For other augmentation methods, we apply Gaussian noise $\mathcal{N}(0,\sigma^2\mathbf{I})$ to half of the training samples in the mini-batch to ensure good certification performance using RS, and 
we follow Hendrycks~\etal to apply JSD to these augmentation methods~\cite{hendrycks2019augmix}.

\noindent\textbf{Datasets.}\quad For the in-distribution evaluation, we use CIFAR-10/100~\cite{krizhevsky2009learning} and ImageNet~\cite{deng2009imagenet} datasets. 
CIFAR-10/100 consist of small $32 \times 32$ images belonging to 10/100 classes and ImageNet consists of 1.2 millions images with 1,000 classes. 
We crop images in ImageNet into the same size of $224\times224\times3$ pixels.
For OOD data, we use the common corruptions datasets~\cite{hendrycks2019benchmarking} (CIFAR-10/100-C, ImageNet-C) and a recently proposed dataset \cite{mintun2021interaction} (CIFAR-10/100-$\overline{\text{C}}$ ImageNet-$\overline{\text{C}}$) which contains human interpretable and perceptually different corruptions as compared to those contained in CIFAR-C/ImageNet-C.

\subsection{Results on CIFAR-Based OOD Benchmarks}
\label{sec:cifar-10/100}
\begin{figure*}[t]
\vspace{-0.4cm}
\subfigure[Gaussian \label{fig:}]{\begin{minipage}[t]{0.138\linewidth}
\begin{center}
\includegraphics[width=\linewidth]{images/incorrect_radius_hm_ga.png} 
\end{center}
\end{minipage}}
\subfigure[Gaussian + JSD \label{fig:}]{\begin{minipage}[t]{0.138\linewidth}
\begin{center}
\includegraphics[width=\linewidth]{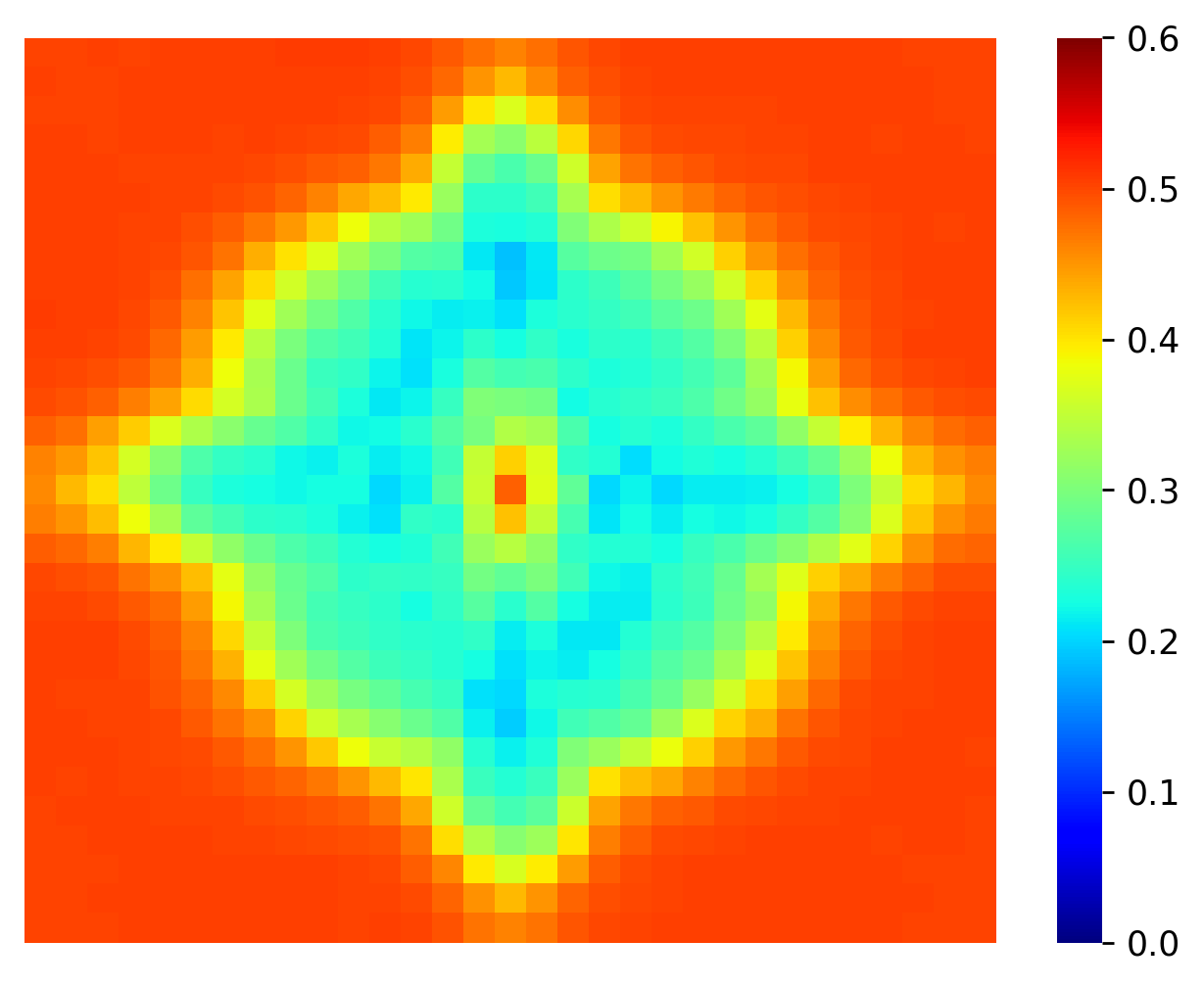} 
\end{center}
\end{minipage}}
\subfigure[AA + JSD \label{fig:}]{\begin{minipage}[t]{0.138\linewidth}
\begin{center}
\includegraphics[width=\linewidth]{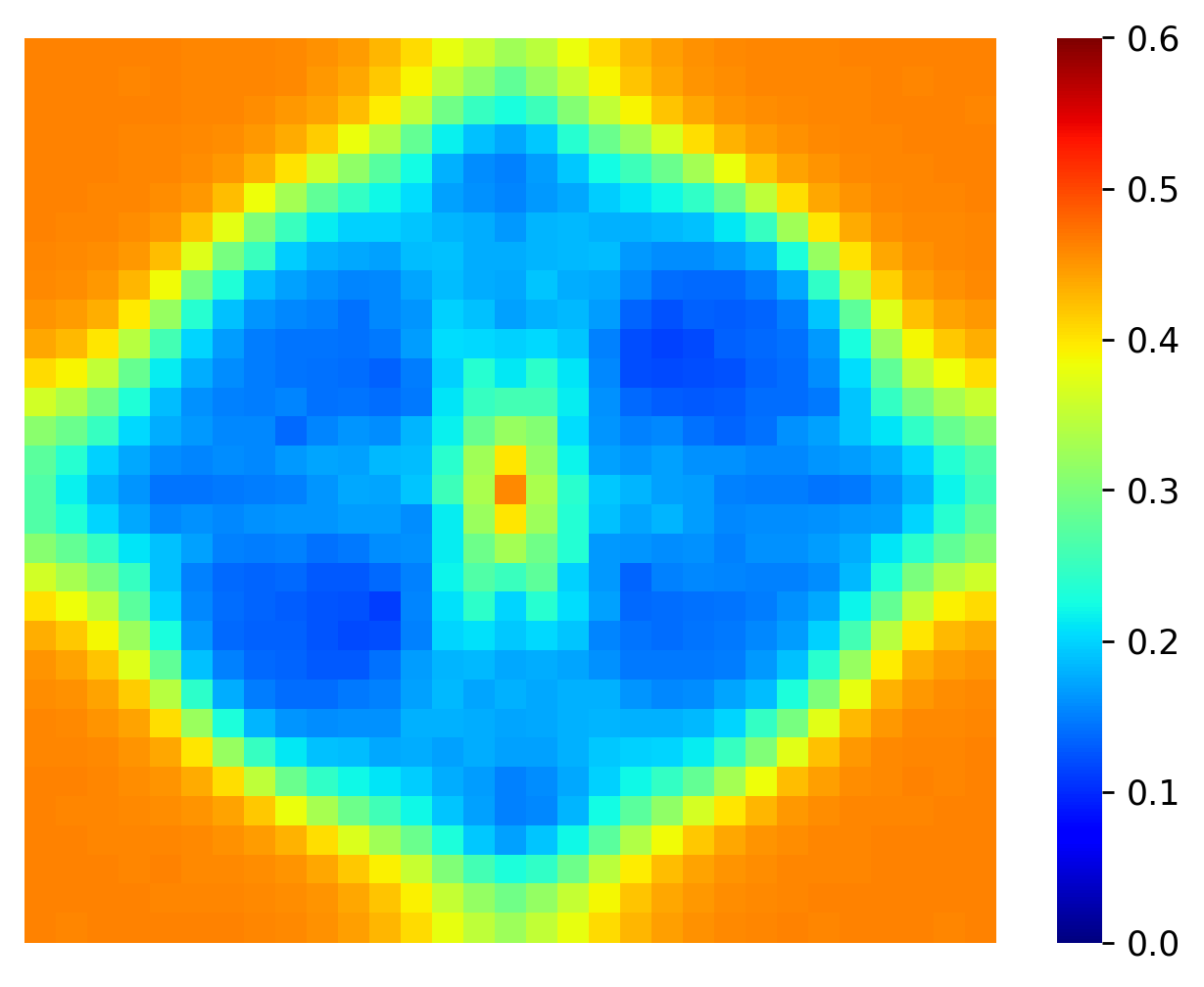} 
\end{center}
\end{minipage}}
\subfigure[AugMix + JSD \label{fig:}]{\begin{minipage}[t]{0.138\linewidth}
\begin{center}
\includegraphics[width=\linewidth]{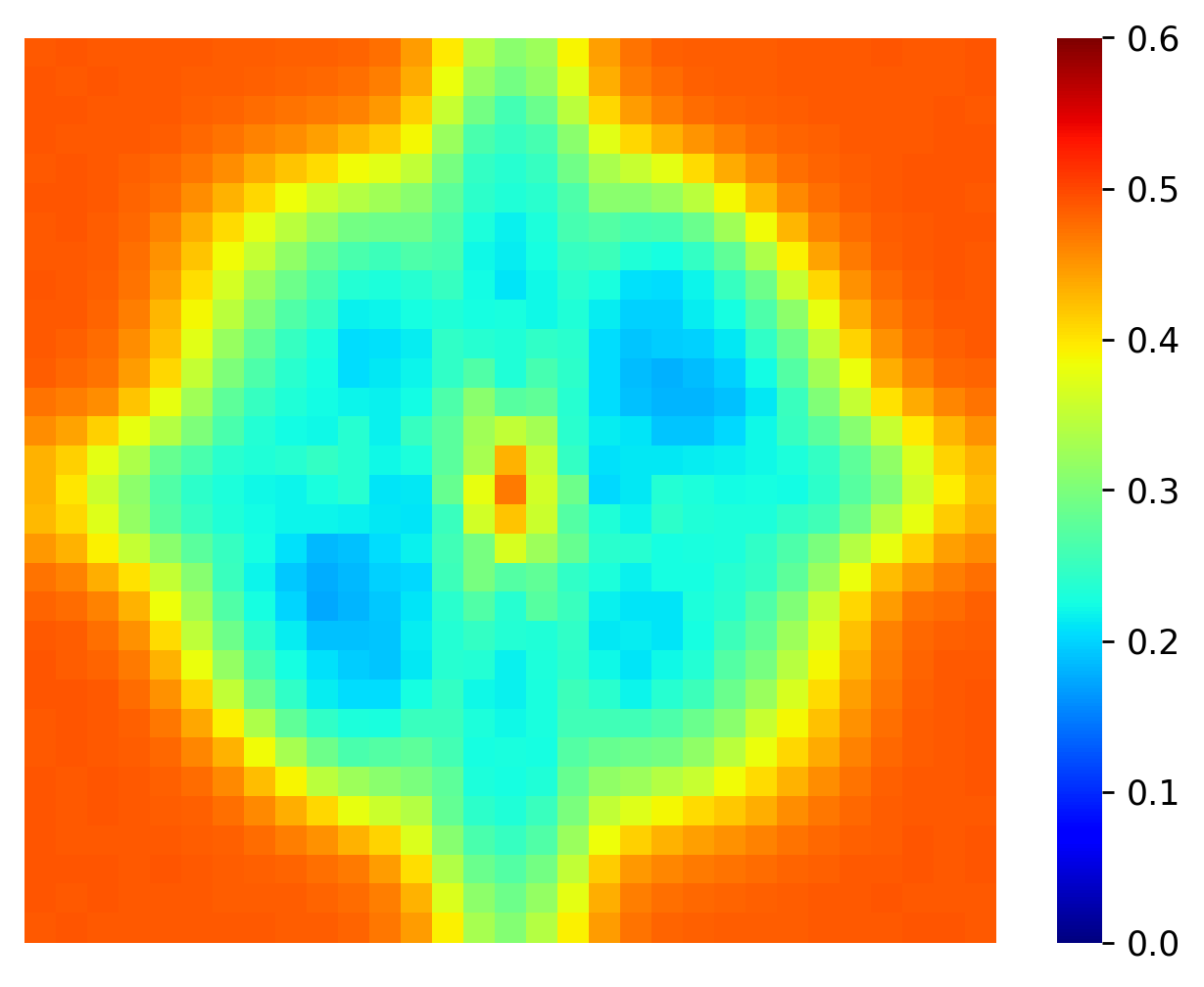} 
\end{center}
\end{minipage}}
\subfigure[AugMix + \textbf{HCR} \label{fig:}]{\begin{minipage}[t]{0.138\linewidth}
\begin{center}
\includegraphics[width=\linewidth]{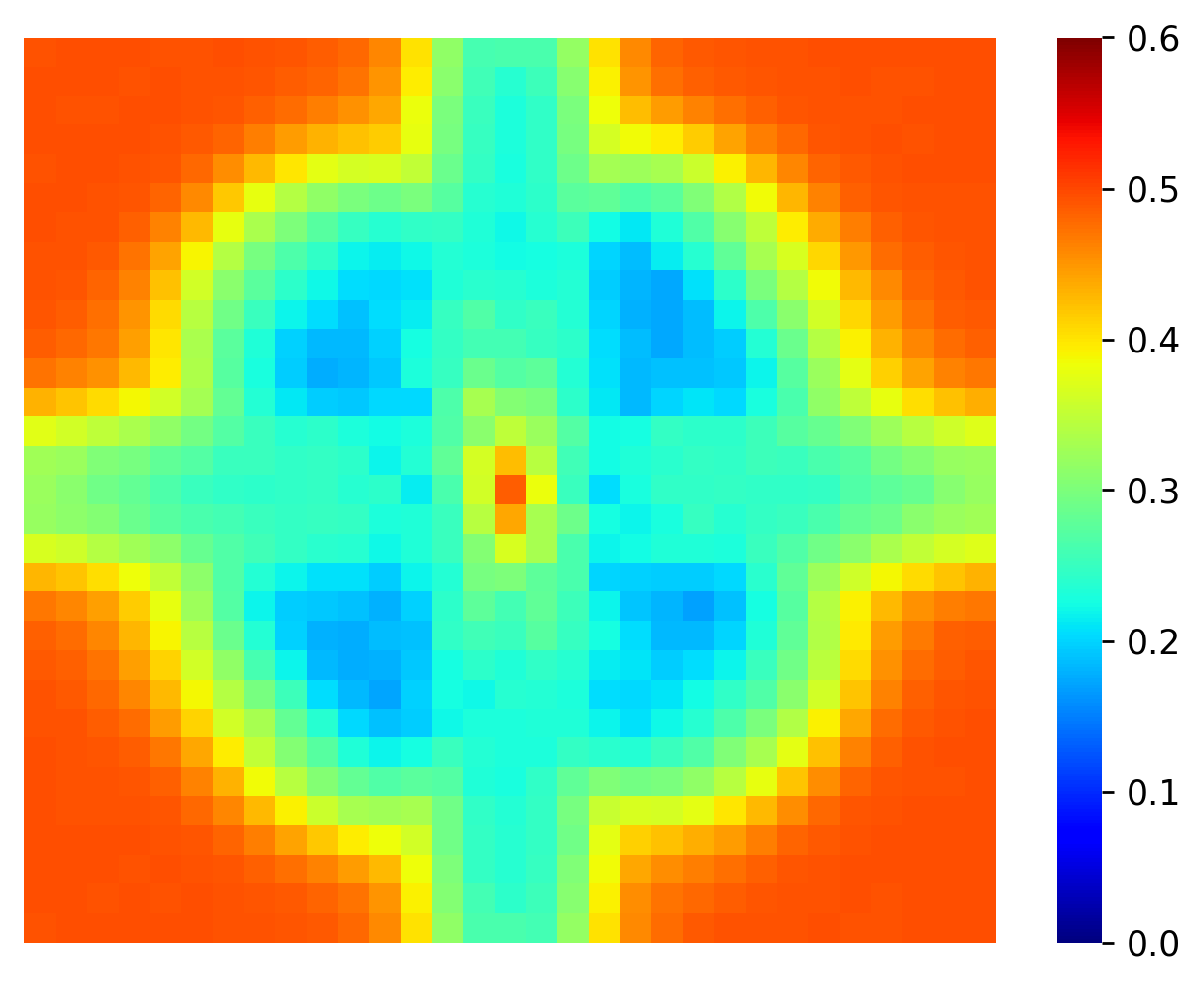}
\end{center}
\end{minipage}}
\subfigure[\textbf{F-Mix} + JSD \label{fig:}]{\begin{minipage}[t]{0.138\linewidth}
\begin{center}
\includegraphics[width=\linewidth]{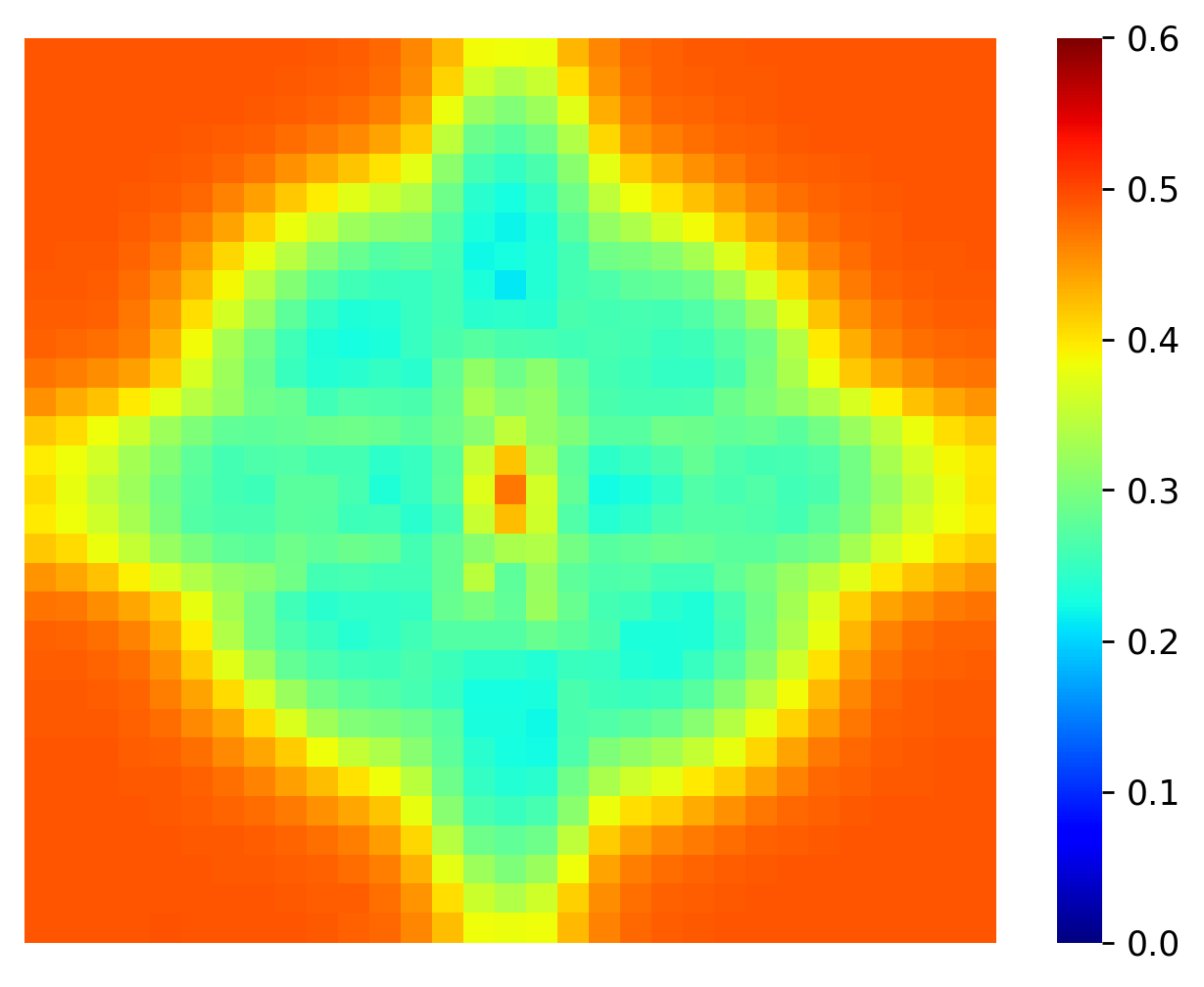}
\end{center}
\end{minipage}}
\subfigure[\textbf{F-Mix + HCR} \label{fig:}]{\begin{minipage}[t]{0.138\linewidth}
\begin{center}
\includegraphics[width=\linewidth]{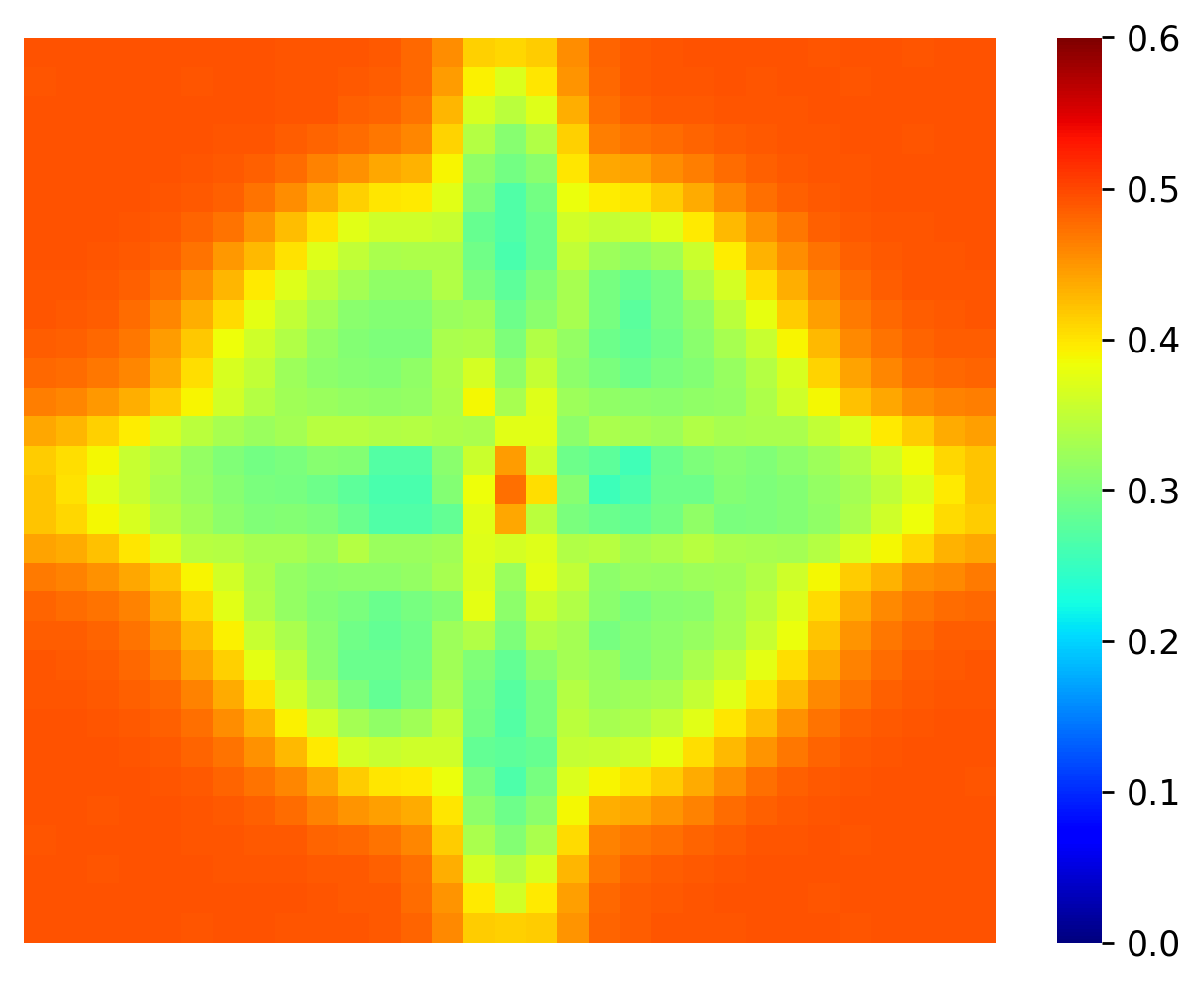}
\end{center}
\end{minipage}}
\subfigure[Gaussian \label{fig:}]{\begin{minipage}[t]{0.138\linewidth}
\begin{center}
\includegraphics[width=\linewidth]{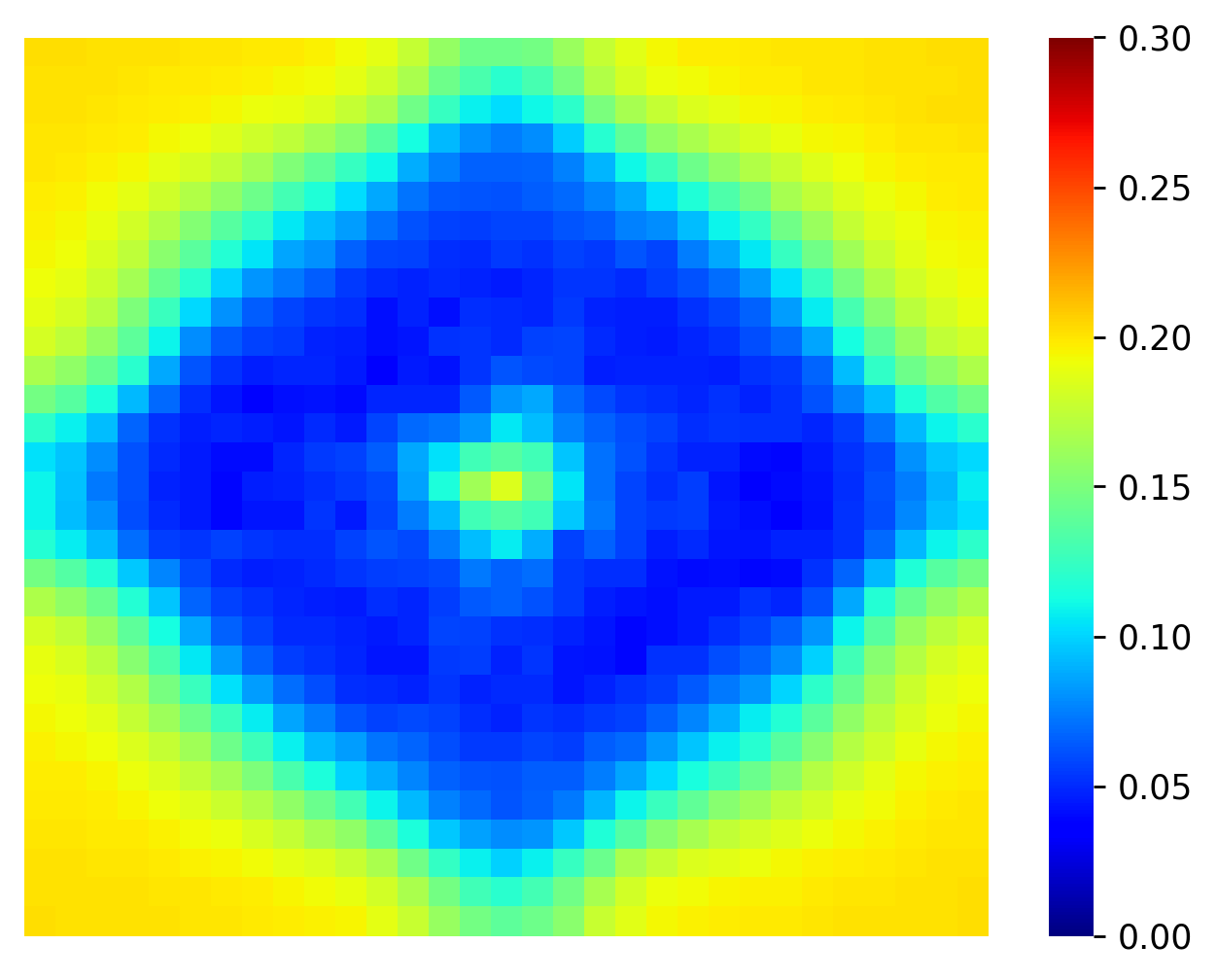} 
\end{center}
\end{minipage}}
\subfigure[Gaussian + JSD \label{fig:}]{\begin{minipage}[t]{0.138\linewidth}
\begin{center}
\includegraphics[width=\linewidth]{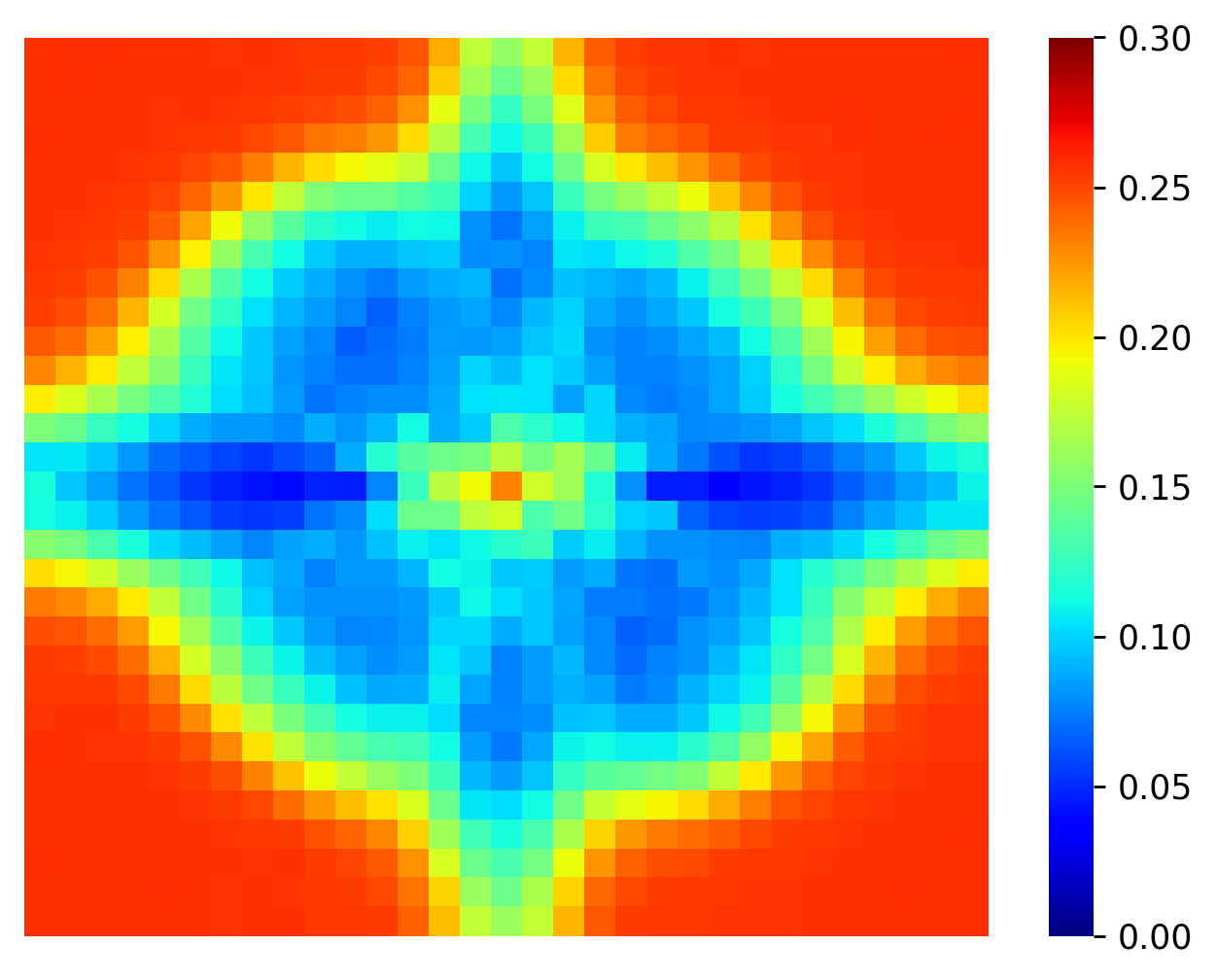} 
\end{center}
\end{minipage}}
\subfigure[AA + JSD \label{fig:}]{\begin{minipage}[t]{0.138\linewidth}
\begin{center}
\includegraphics[width=\linewidth]{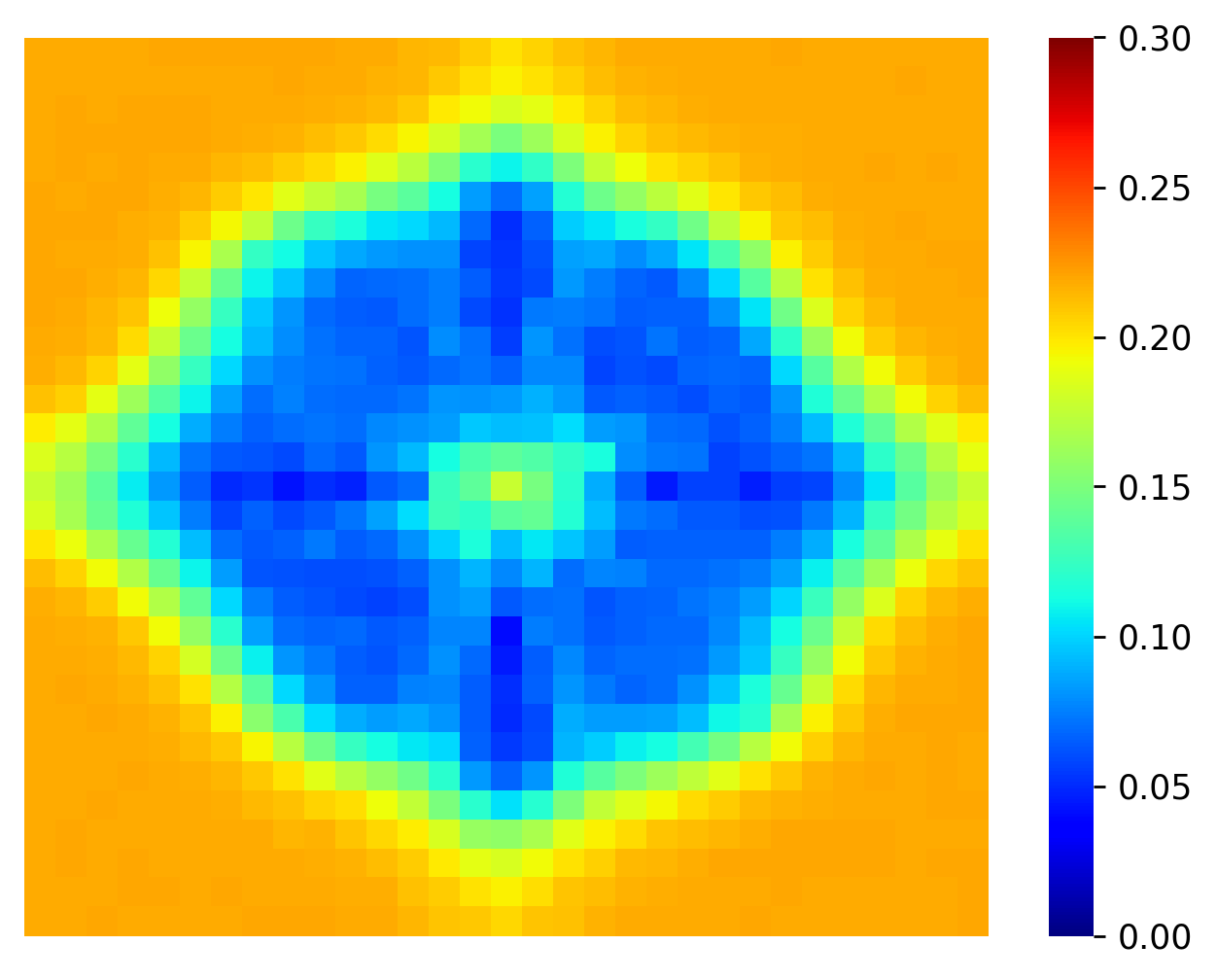} 
\end{center}
\end{minipage}}
\subfigure[AugMix + JSD \label{fig:}]{\begin{minipage}[t]{0.138\linewidth}
\begin{center}
\includegraphics[width=\linewidth]{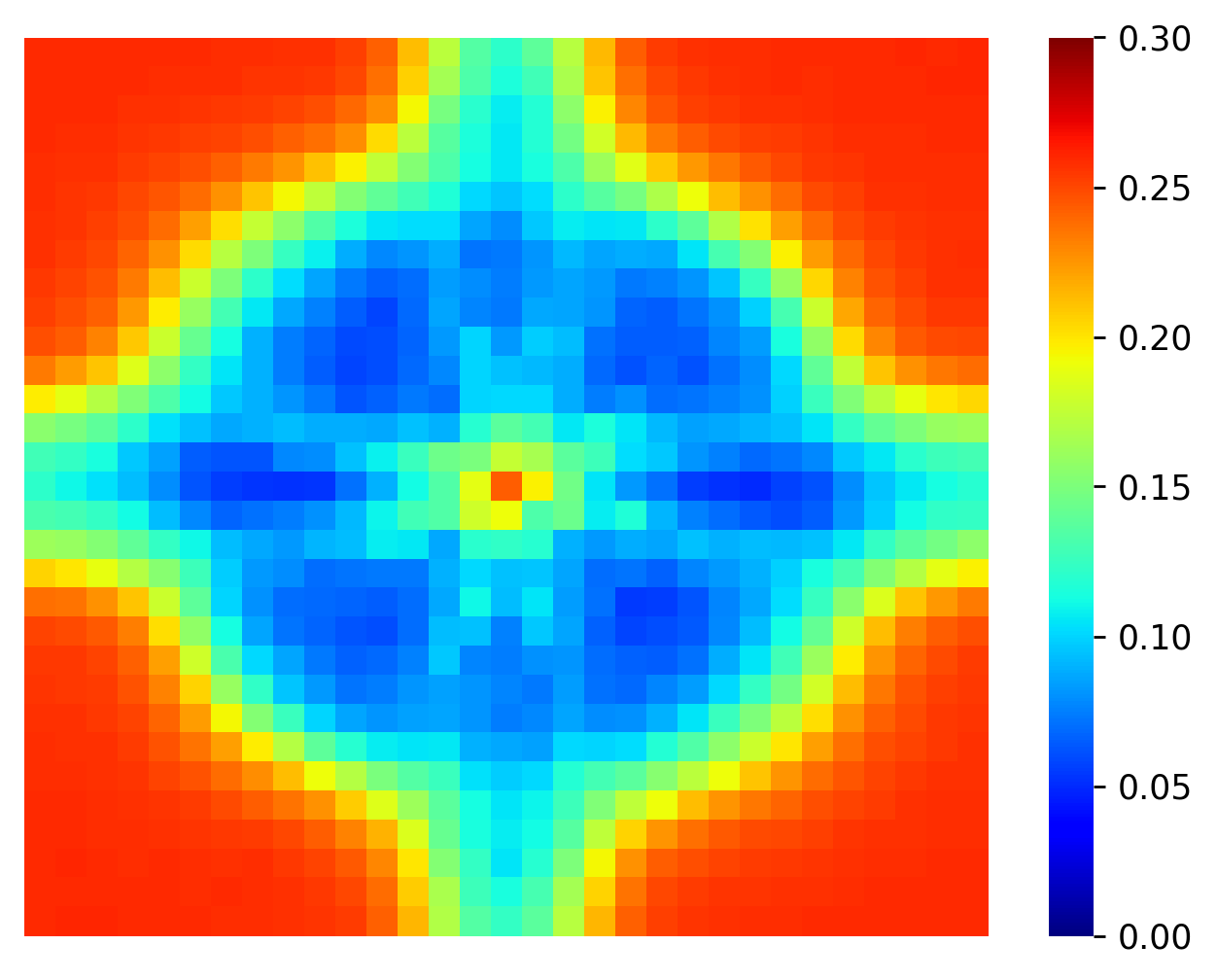} 
\end{center}
\end{minipage}}
\subfigure[AugMix + \textbf{HCR} \label{fig:}]{\begin{minipage}[t]{0.138\linewidth}
\begin{center}
\includegraphics[width=\linewidth]{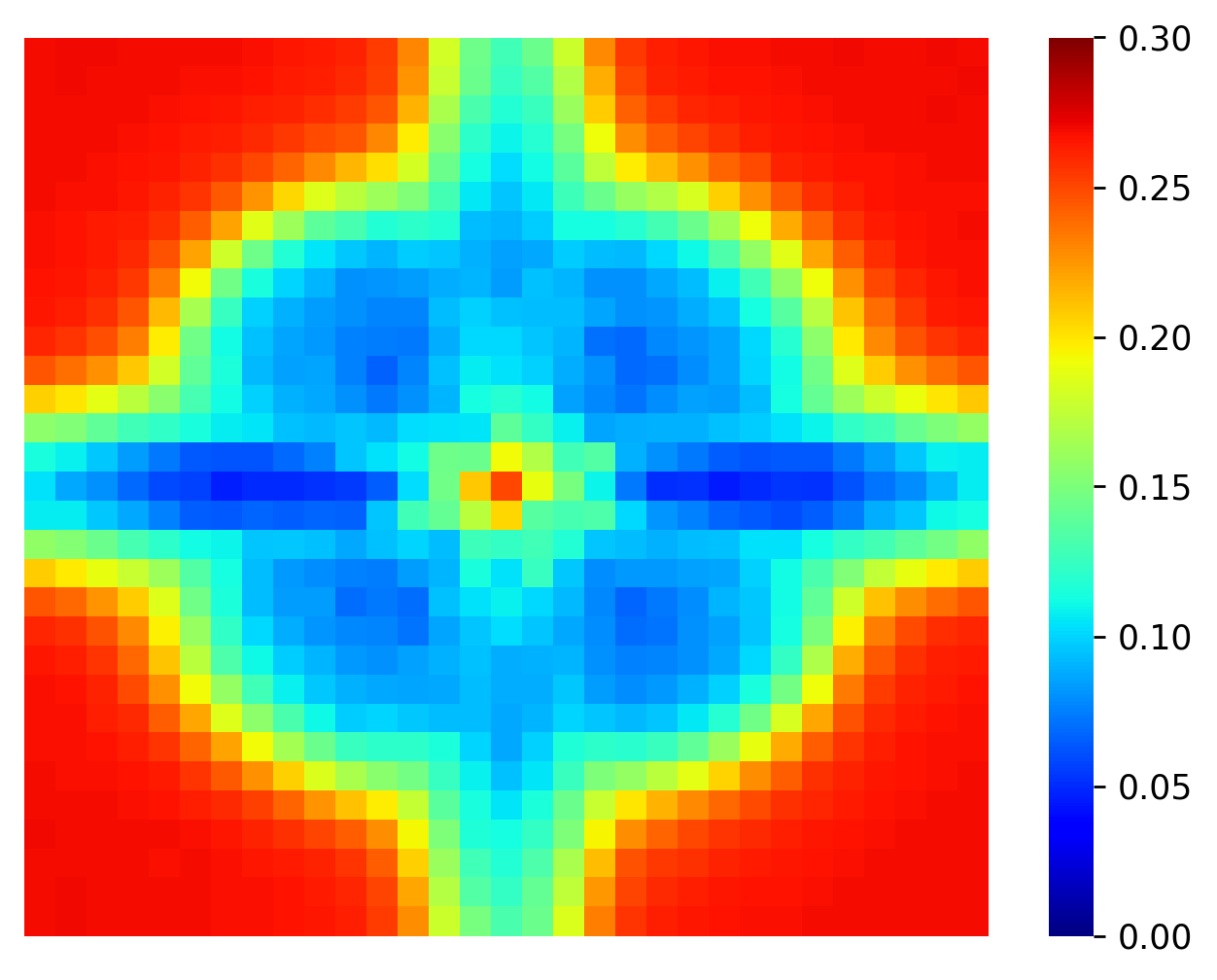}
\end{center}
\end{minipage}}
\subfigure[\textbf{F-Mix} + JSD \label{fig:}]{\begin{minipage}[t]{0.138\linewidth}
\begin{center}
\includegraphics[width=\linewidth]{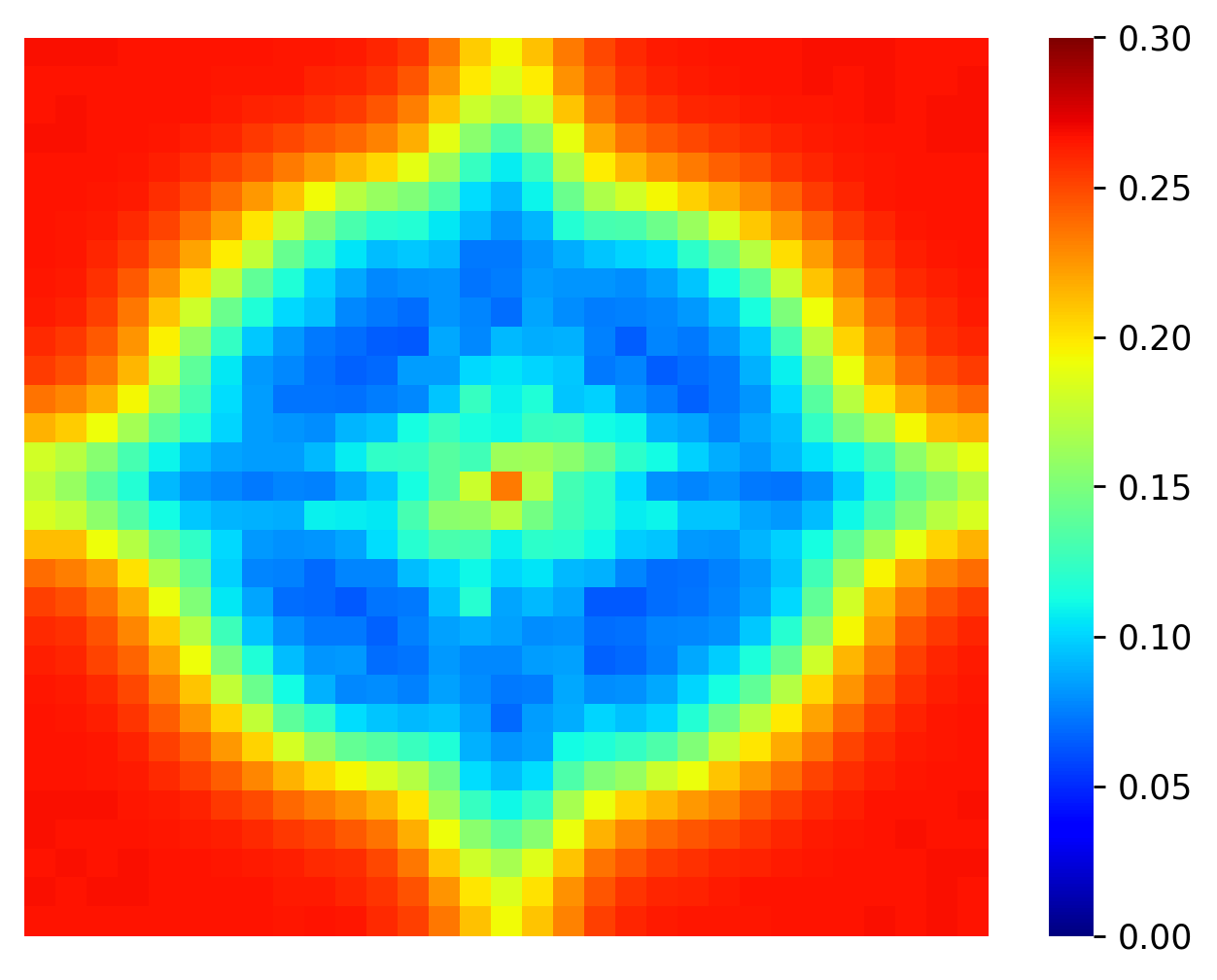}
\end{center}
\end{minipage}}
\subfigure[\textbf{F-Mix + HCR} \label{fig:}]{\begin{minipage}[t]{0.138\linewidth}
\begin{center}
\includegraphics[width=\linewidth]{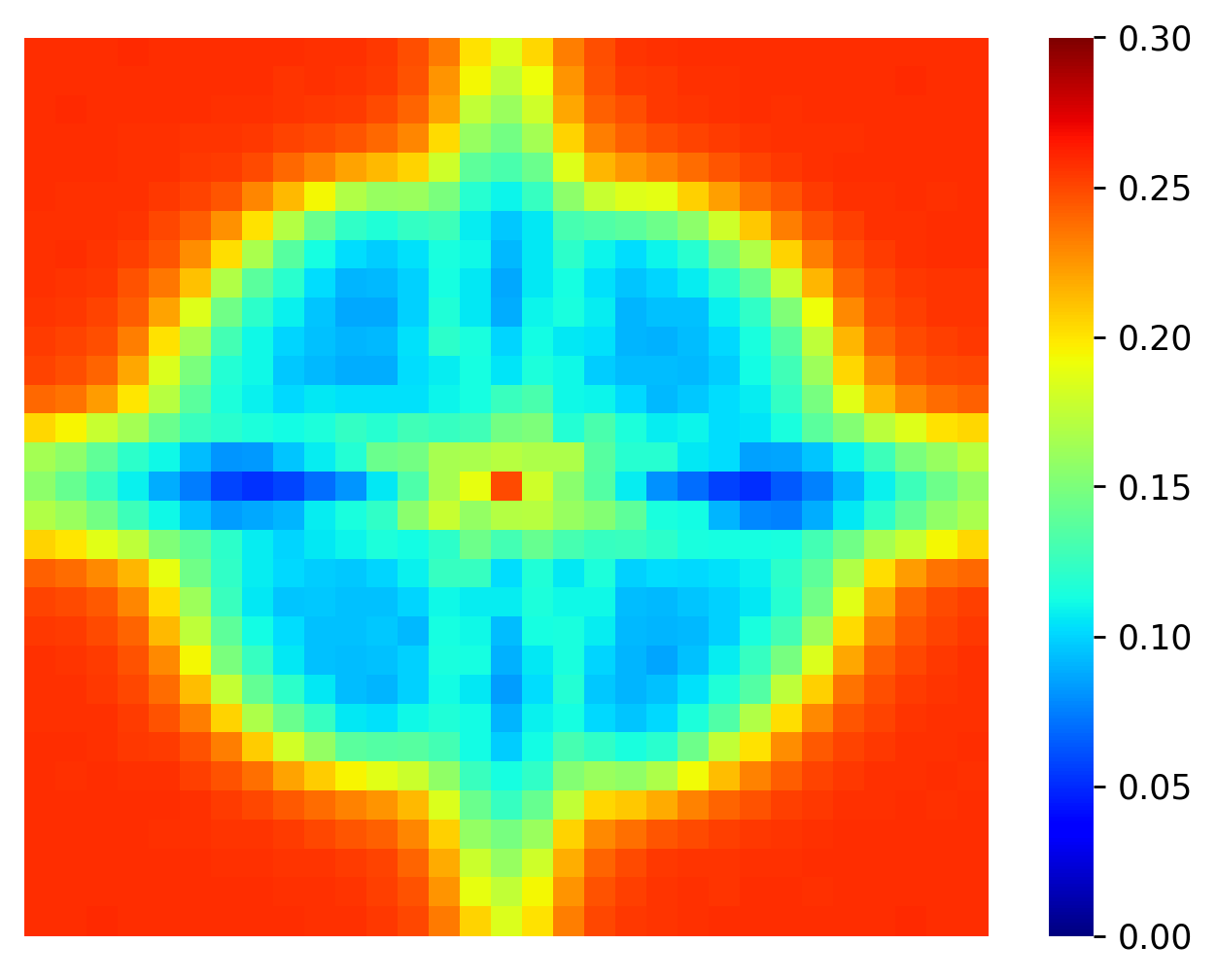}
\end{center}
\end{minipage}}
\vspace{-0.35cm}
\caption{Fourier sensitivity analysis of models trained using different augmentations and regularizers on CIFAR-10 (the first row) and CIFAR-100 (the second row) demonstrate their vulnerability to OOD data from mid-to-low frequency region (around the center of the plots). \emph{FourierMix} augmentations coupled with HCR significantly improves the certified robustness under OOD data from the entire spectrum. (AA: AutoAugment, F-Mix: \emph{FourierMix}).}
\label{fig:fourier_basis_cifar10}
\vspace{-0.35cm}
\end{figure*}


The results in Tables~\ref{tb:total} show the overall mACR of the models trained on CIFAR-10 using different augmentation and regularization methods when evalauated  on CIFAR-10-C and CIFAR-10-$\bar{\text{C}}$, respectively. 
The results show that \emph{FourierMix} consistently achieves the highest mACR across different corruption types. 
\emph{FourierMix}+HCR significantly improves upon the baseline of Gaussian augmented training by 26.7\% and 33.4\% in terms of the overall mACR on CIFAR-10-C and CIFAR-10-$\bar{\text{C}}$, respectively and also improves upon the stronger baseline, AugMix+HCR, by 5.3\% and 6.6\% on the two datasets.
We find consistency regularization to be helpful for certified robustness on OOD data. 
Specially, on mid- and high frequency corruptions adding JSD to Gaussian augmentations significantly improves the robustness on OOD data. 
We see that combining HCR with \emph{FourierMix} achieves SOTA ACRs on all corruptions types providing significant gains even on low-frequency corruptions. This success is attributed to the spectrally diverse corruptions produced by \emph{FourierMix}.
Interestingly, we find AutoAugment overfits to corruptions in CIFAR-10-C since it suffers a major performance degradation on corruptions in CIFAR-10-$\bar{\text{C}}$. 
We believe the large overlap between the leveraged augmentations and corruptions in CIFAR-10-C and limited spectral diversity are the primary reasons for this performance degradation of AutoAugment. {Detailed results for each corruption type in CIFAR-10-C/$\bar{\text{C}}$ are shown in Tables~\ref{tb:cifar10c} and~\ref{tb:cifar10c-bar} in Appendix~\ref{app:detailed_cifar10}.}


Next, we present the mACR (Table~\ref{tb:total}) of the models trained with CIFAR-100 when evaluated on OOD data (CIFAR-100-C and CIFAR-100-$\bar{\text{C}}$).
Similar to the performance of models trained with CIFAR-10, \emph{FourierMix} achieves the highest overall mACR among all augmentation methods on both OOD datasets. 
Specifically, \emph{FourierMix}+HCR outperforms the Gaussian baseline by 54.4\% and 74.6\% on two datasets, respectively. 
Compared to AugMix+HCR, \emph{FourierMix}+HCR improves the performance by 4.8\% and 7.6\% on the two datasets, respectively. {Detailed results for each corruption type in CIFAR-100-C/$\bar{\text{C}}$ are shown in Tables~\ref{tb:cifar100c} and~\ref{tb:cifar100c-bar} in Appendix~\ref{app:detailed_cifar100}.}


{
To further corroborate our findings on OOD benchmarks, we carry out the Fourier sensitivity analysis of models trained on CIFAR-10/100 in Figure~\ref{fig:fourier_basis_cifar10}.
Adding a consistency loss (Gaussian+JSD) improves the ACR of the model in the high-frequency region but is still worse than the ACR achieved by the addition of consistency loss (JSD and HCR) with \emph{FourierMix} augmentations in low-to-mid frequency regions. 
Similar to our quantitative results, AutoAugment does not improve much over the baseline of Gaussian augmentation which suggests that models trained with AutoAugment may be biased towards high frequency regions. 

}

\subsection{Results on ImageNet-Based OOD Benchmarks}
\label{sec:exp-imagenet}



Table~\ref{tb:total} presents the mACR of the models trained on ImageNet when evaluated on ImageNet-C and ImageNet-$\bar{\text{C}}$. Due to poor performance of some of the methods on CIFAR-10/100, we chose not to pursue them for ImageNet scale experiments (denoted by ``-'' in Table~\ref{tb:total}).
We observe that OOD shifts lead to a drastic decline in the certified robustness on ImageNet. 
The drop between the ACR of clean data and the mACR of OOD data is $\sim$57\%, whereas it was $\sim$30\% on CIFAR-10/100. 
Encouragingly, \emph{FourierMix} continues to achieve the highest mACR compared to other baselines. 
\emph{FourierMix} outperforms the baseline of Gaussian augmented training and AugMix+JSD by 55.9\% and 2.1\% in terms of the overall mACR, respectively. \emph{FourierMix} also realizes consistent good performance across the spectrum, whereas Gaussian+JSD and AugMix+JSD are biased to high-frequency and low-frequency corruptions, respectively. 
Despite HCR not making a significant difference over JSD regularization, it is worth noting that substantial improvements can still be gained by \emph{FourierMix} (over other baseline augmentations) on ImageNet due to its broad spectral coverage. {Detailed results for ImageNet-C/$\bar{\text{C}}$ can be found in Tables~\ref{tb:imagenet-c} and~\ref{tb:imagenet-c-bar} in Appendix~\ref{app:detailed_imagenet}.}

\noindent\textbf{Overall Insights}.\quad Our results in this section not only highlight the vulnerability of SOTA certified defenses to OOD data but also uncovers spectral biases in the benchmark datasets that are used to measure OOD robustness. In particular, methods that perform well on one corrupted dataset may not work well on other dataset due to differences in the spectral signatures of the corruptions. For example, AutoAugment performs well on CIFAR10-C corruptions but is significantly worse on CIFAR10-$\bar{\text{C}}$ and CIFAR100-C/$\bar{\text{C}}$. Moreover, we find that such ranking discrepancies not only exist when measuring ACR but are also evident when measuring empirical robust accuracy. Figure~\ref{fig:robustbench} in Appendix~\ref{app:robustbench} shows that there is no single SOTA model from RobustBench leaderboard~\cite{croce2020robustbench} which performs the best on all benchmark datasets or corruption types. 
This makes it incredibly important to obtain a comprehensive view of the model robustness to avoid issues such as leaderboard bias~\cite{mishra2021robust} and model overfitting to a specific benchmark~\cite{mintun2021interaction}. 
To enable researchers achieve this objective, next we propose a new benchmark which has a collection of spectrally diverse OOD datasets.
\section{A Spectral OOD Benchmarking Suite}
\label{sec:cifarf}



Here we discuss the creation and evaluation of models on the proposed OOD benchmarking suite. The goal of this new OOD suite is to complement (and not replace) the existing benchmark datasets and enable researchers to uncover and resolve spectral biases of their models. 
\begin{figure*}[t]
\vspace{-0.25cm}
\subfigure[CIFAR-10-F with $\alpha=0.5$.]{\begin{minipage}[t]{0.245\linewidth}
\begin{center}
\includegraphics[width=\linewidth, height=3cm]{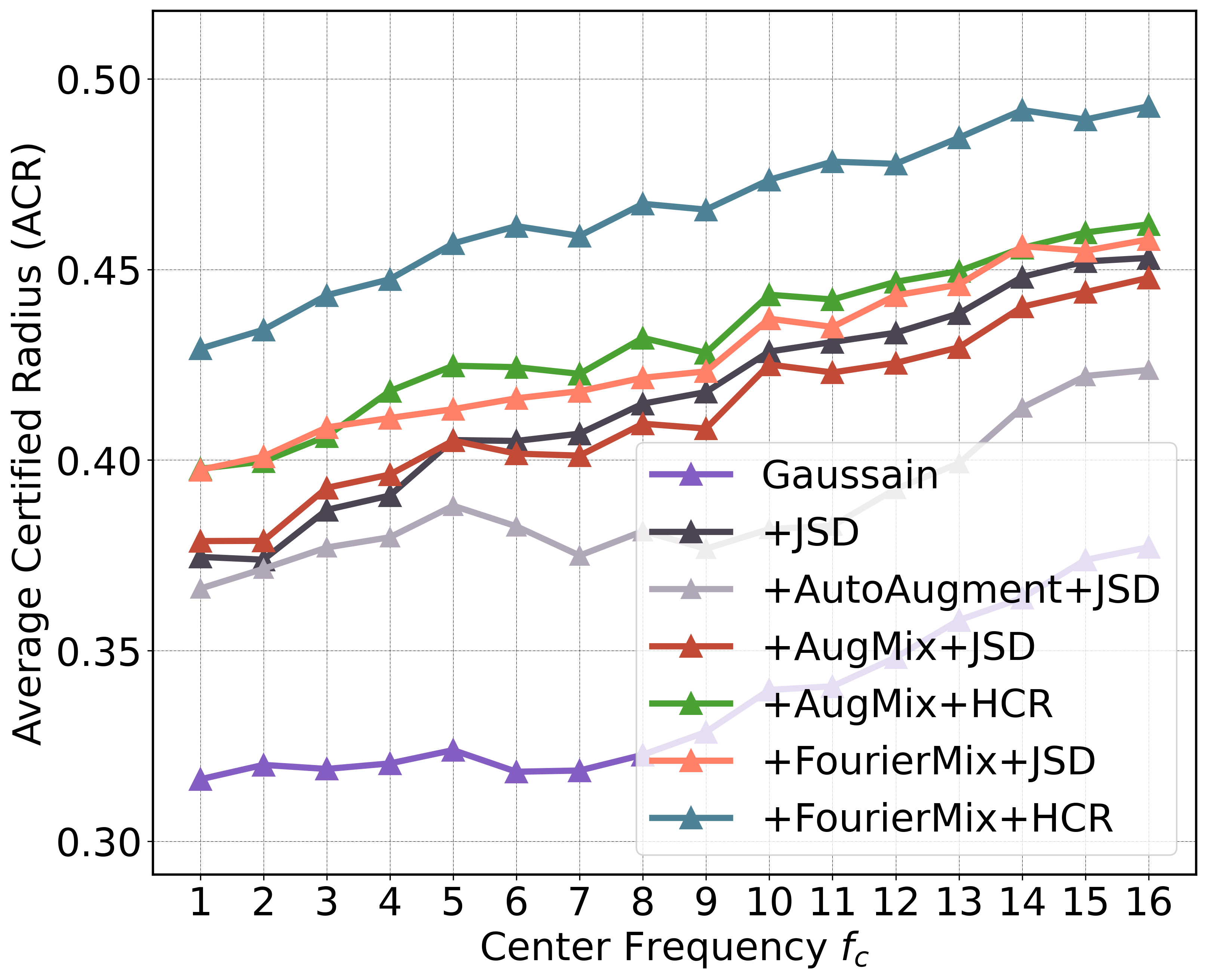} 
\end{center}
\end{minipage}}
\subfigure[CIFAR-10-F with $\alpha=1$.]{\begin{minipage}[t]{0.245\linewidth}
\begin{center}
\includegraphics[width=\linewidth, height=3cm]{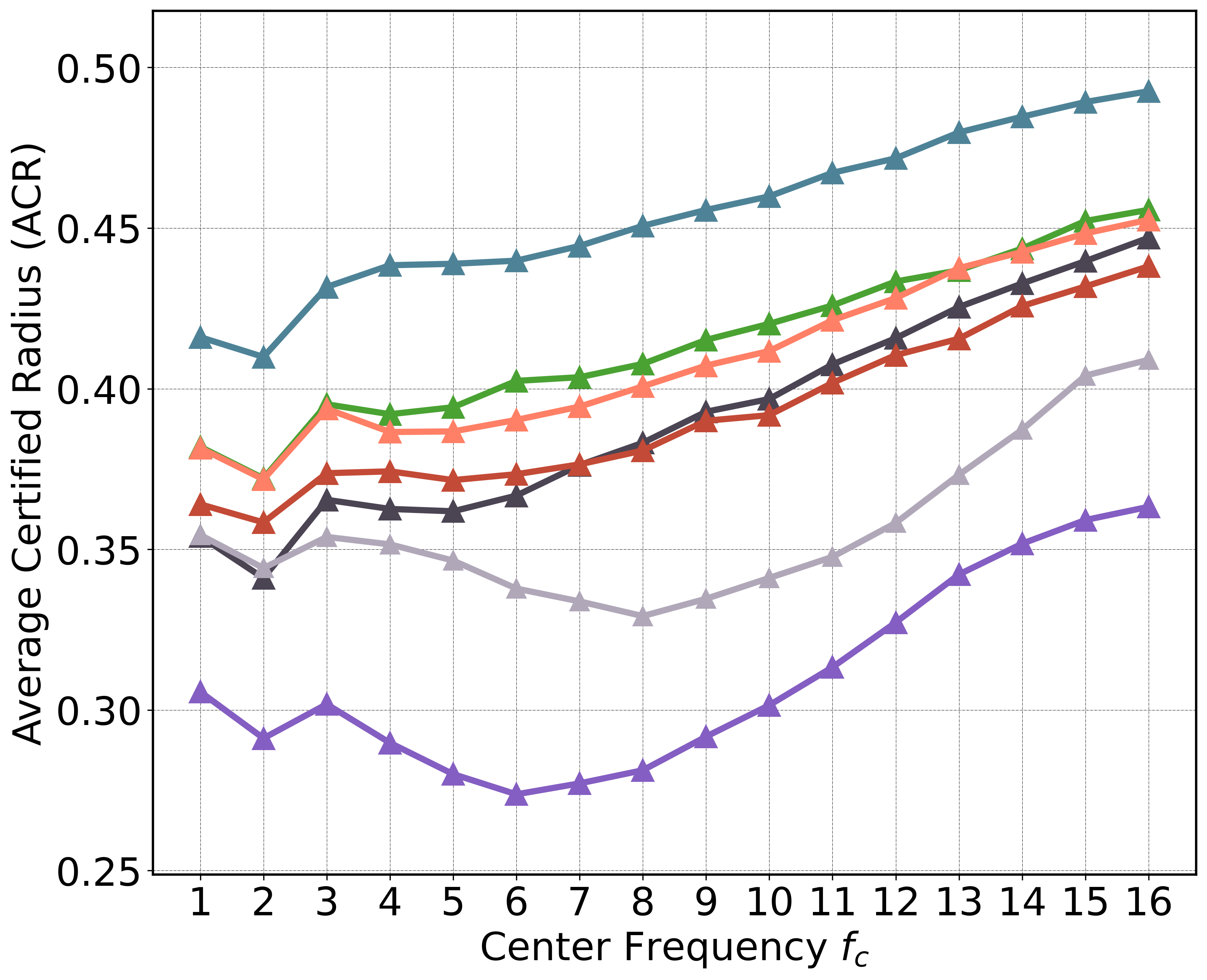} 
\end{center}
\end{minipage}}
\subfigure[CIFAR-10-F with $\alpha=2$.\label{fig:cifar-10-f-2}]{\begin{minipage}[t]{0.245\linewidth}
\begin{center}
\includegraphics[width=\linewidth, height=3cm]{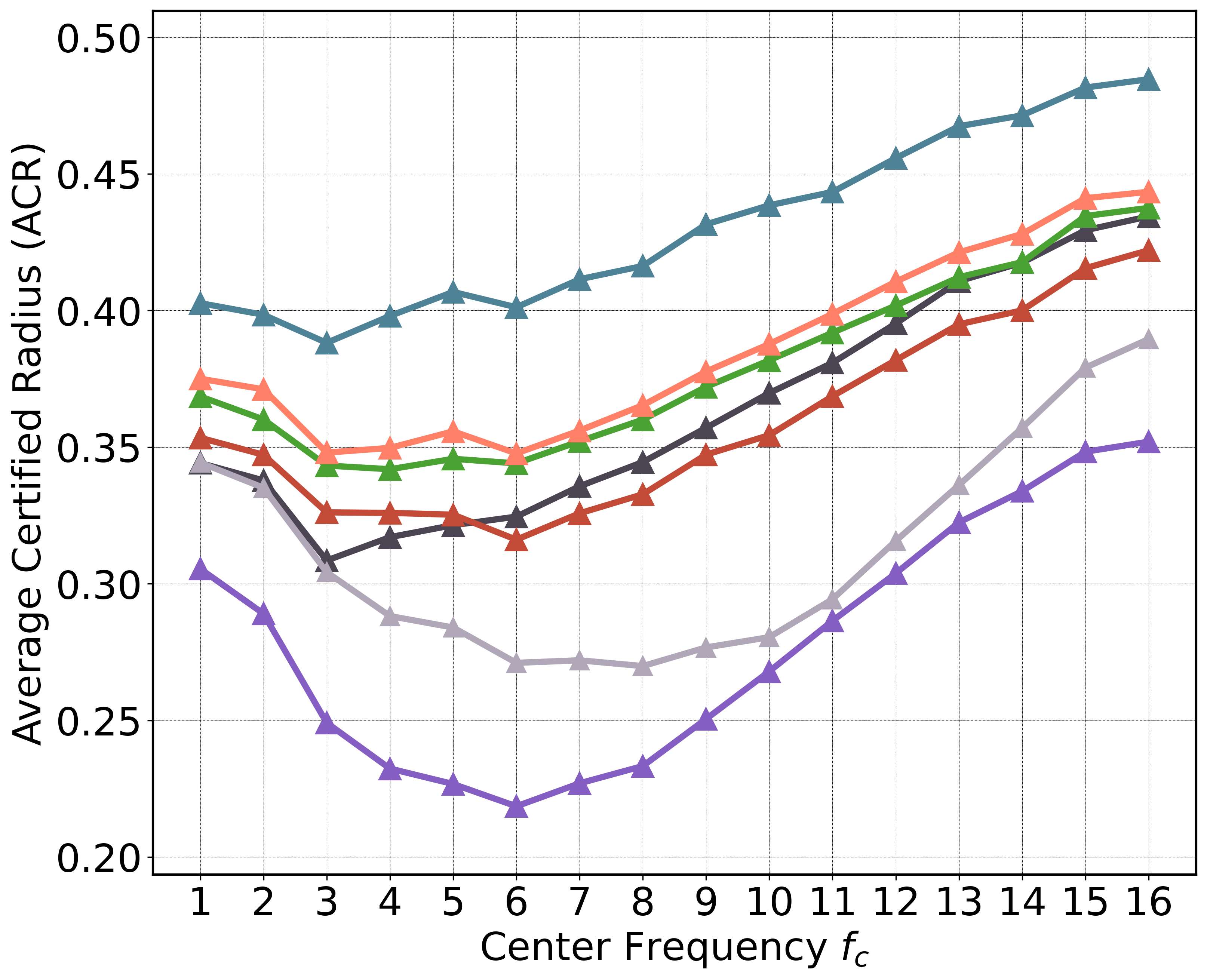} 
\end{center}
\end{minipage}}
\subfigure[CIFAR-10-F with $\alpha=3$.\label{fig:cifar-10-f-3}]{\begin{minipage}[t]{0.245\linewidth}
\begin{center}
\includegraphics[width=\linewidth, height=3cm]{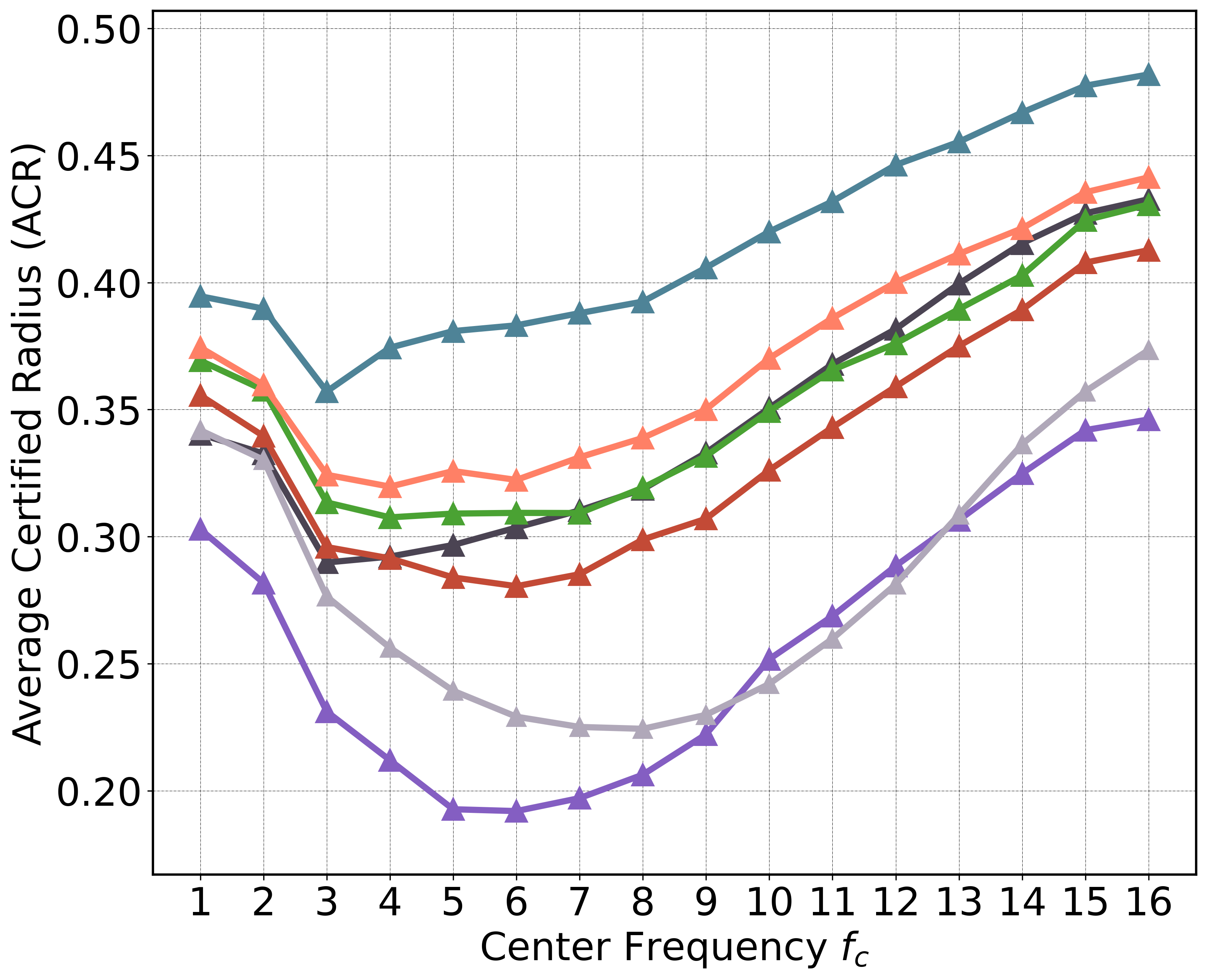} 
\end{center}
\end{minipage}}
\subfigure[CIFAR-100-F with $\alpha=0.5$.]{\begin{minipage}[t]{0.245\linewidth}
\begin{center}
\includegraphics[width=\linewidth, height=3cm]{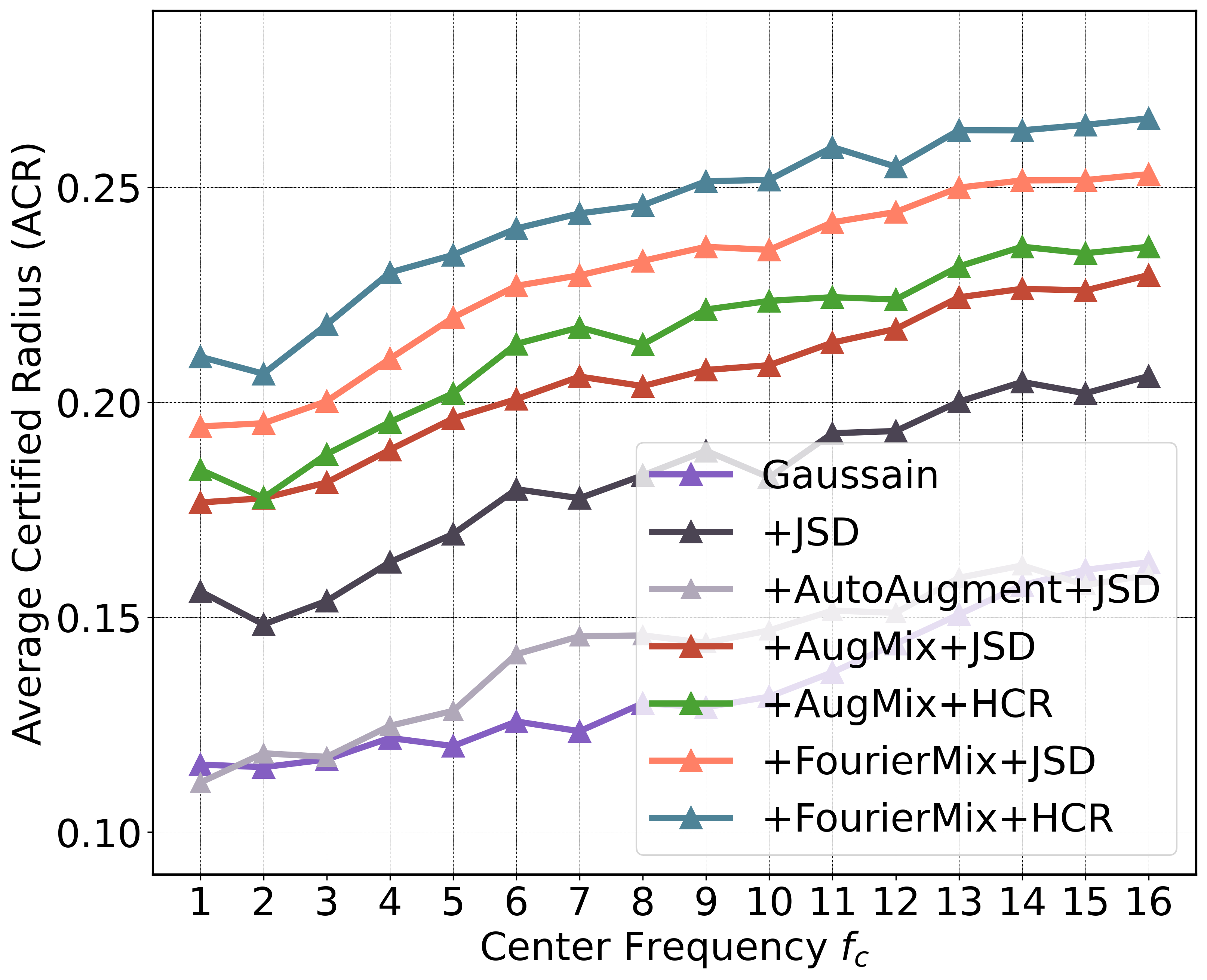} 
\end{center}
\end{minipage}}
\subfigure[CIFAR-100-F with $\alpha=1$.]{\begin{minipage}[t]{0.245\linewidth}
\begin{center}
\includegraphics[width=\linewidth, height=3cm]{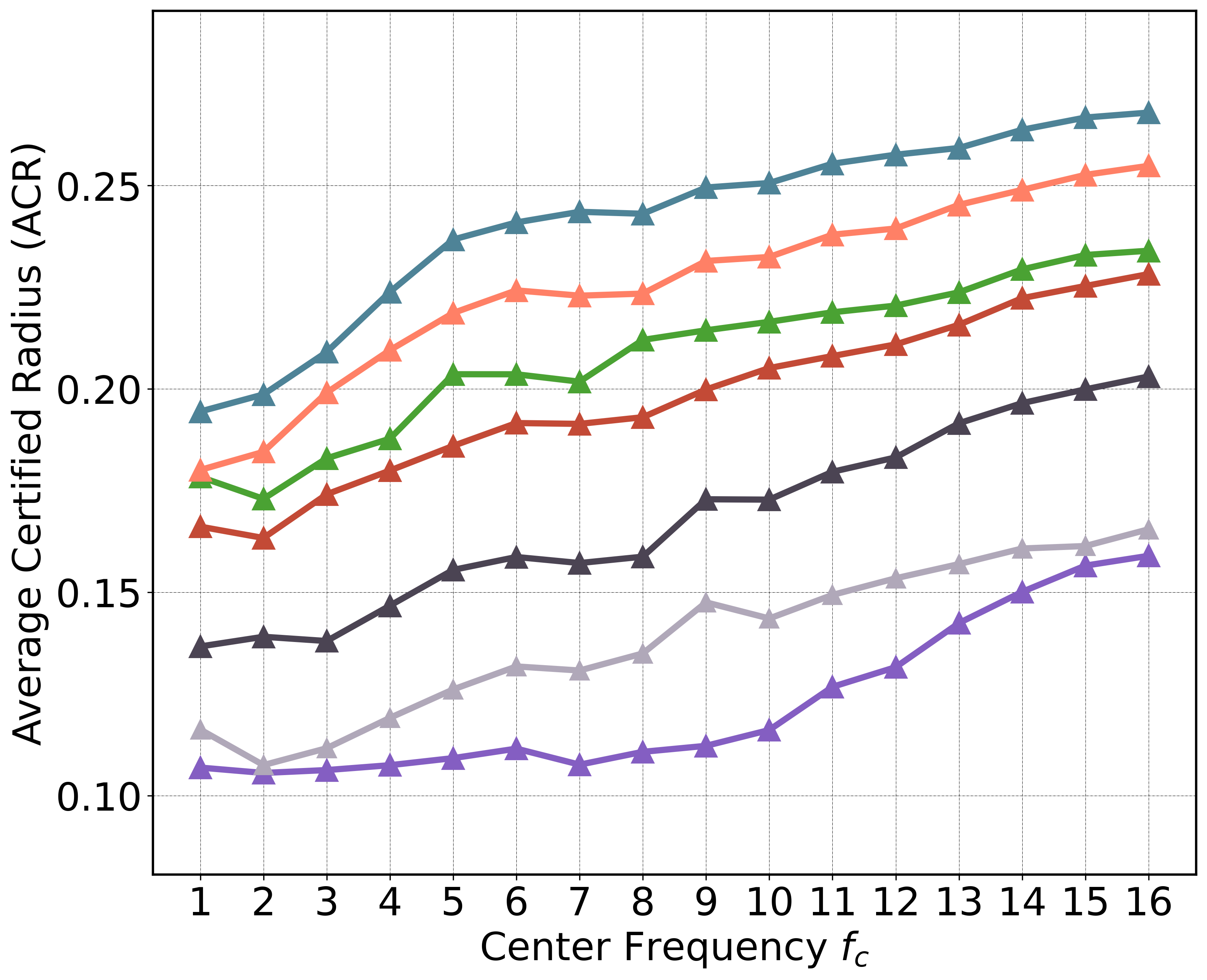} 
\end{center}
\end{minipage}}
\subfigure[CIFAR-100-F with $\alpha=2$.]{\begin{minipage}[t]{0.245\linewidth}
\begin{center}
\includegraphics[width=\linewidth, height=3cm]{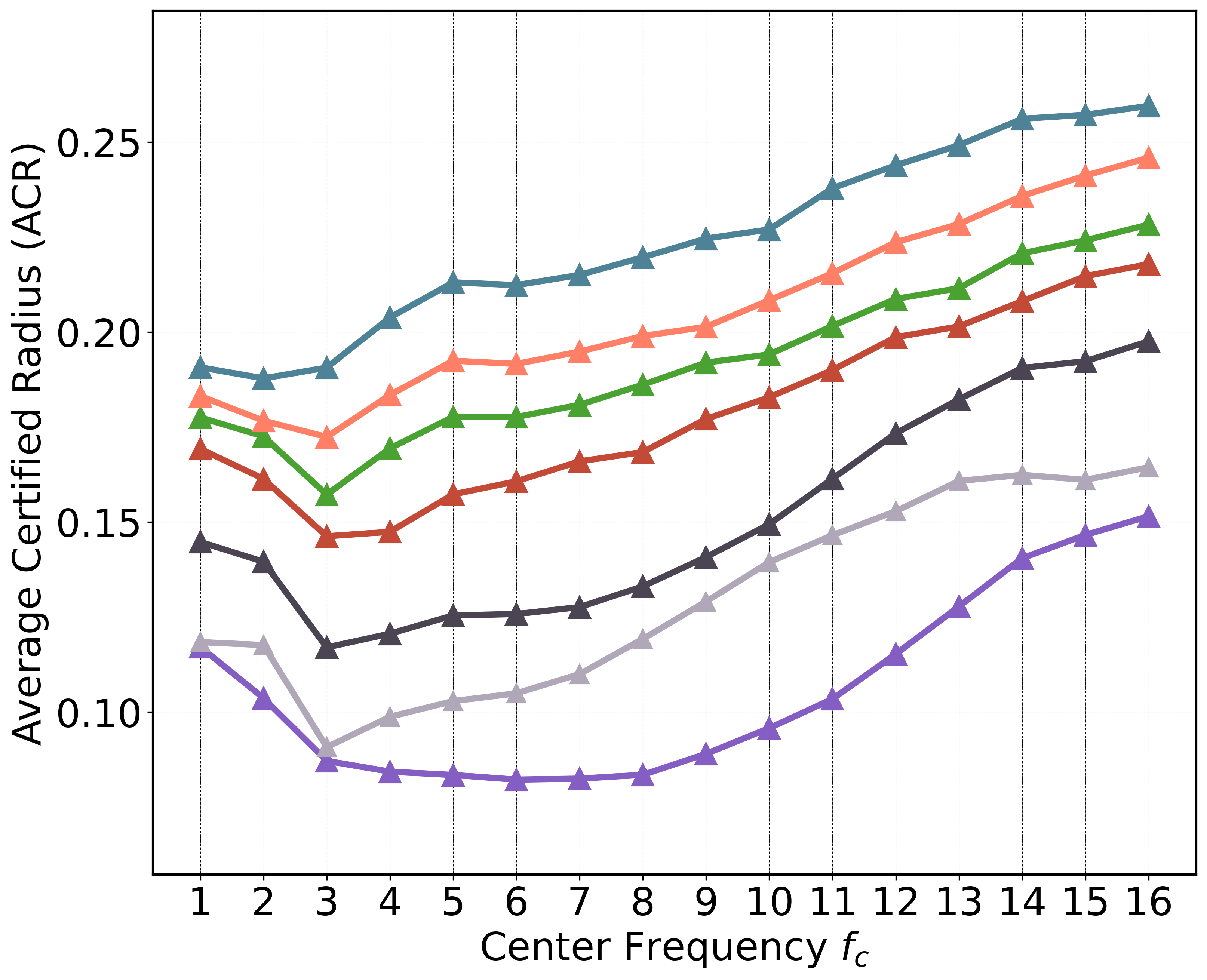} 
\end{center}
\end{minipage}}
\subfigure[CIFAR-100-F with $\alpha=3$.]{\begin{minipage}[t]{0.245\linewidth}
\begin{center}
\includegraphics[width=\linewidth, height=3cm]{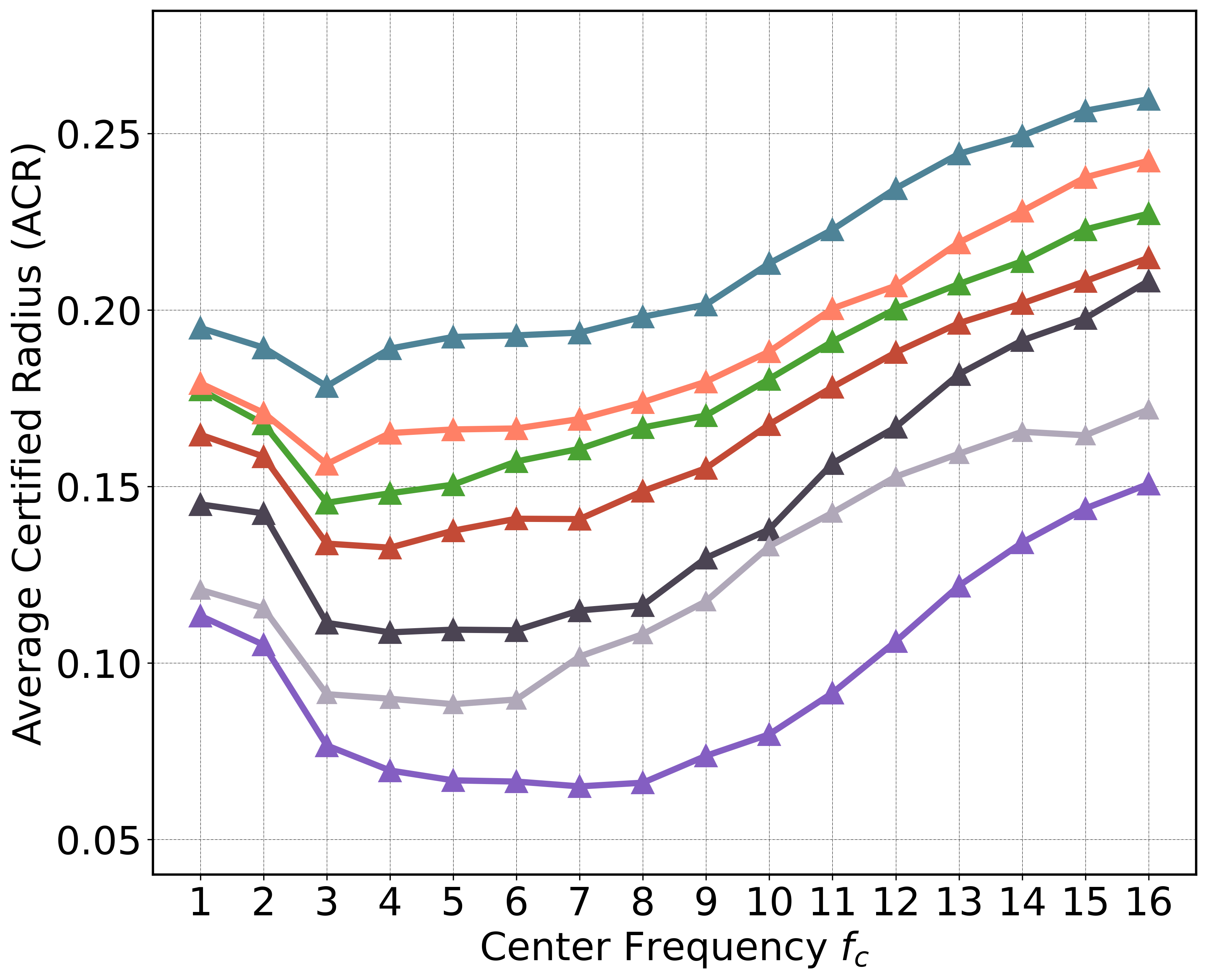} 
\end{center}
\end{minipage}}
\vspace{-0.4cm}
\caption{ACRs on the proposed CIFAR-10/100-F dataset averaged over 3 severity levels show that \emph{FourierMix} based models perform consistently better than other baselines across entire spectrum. Increasing $\alpha$ (from left to right), decreases the spread of the frequencies. Sharp dips in ACRs in mid-frequency regions (\eg, (c) and (d)) demonstrate the vulnerability of models to low-to-mid frequency corruptions.}
\label{fig:cifar-10-f}
\vspace{-0.5cm}
\end{figure*}

\subsection{Protocol for Dataset Generation}
The proposed benchmark is a collection of datasets (CIFAR-10/100-F) each focusing on a specific frequency range while collectively covering the entire frequency spectrum.
Different from the Fourier sensitivity analysis that only perturb a single frequency using the Fourier basis, CIFAR-10/100-F leverages power law-based noise~\cite{powerlaw} to generate complex perturbations in the spectral domain~\cite{johnson1925schottky}. Note that power spectrum of several natural data distributions (e.g., natural images) follow power-law distribution~\cite{powerlaw}. 
Inspired by this, we model the amplitude perturbation as $\vdelta_{\text{Fourier}}(f)_{\mathbf{A}} = \frac{P(f)}{(|f-f_c| + 1)^{\alpha}} \cdot \text{U}(1-b,1+b)$, where $P(f)$ approximates the tolerance of corruptions at azimuthal frequency $f=\sqrt{u^2+v^2}$, $f_c$ is the central frequency that the perturbation focuses on, and $\alpha$ denotes the power of the power law distribution. We also use a uniform distribution $\text{U}(1-b,1+b)$ with $b$ as a hyper-parameter ($b=0.2$ in our study) to diversify the perturbations.
We define $P(f) = \text{clip} (\mA_{\vx}^{\text{clean}}(f), a_{\text{lower}}, a_{\text{upper}})$ which adds the amount of perturbation based on the power associated with the different frequencies in the clean image~\cite{joubert2009rapid}, \ie frequencies with higher power have larger perturbations.
We leverage the $\text{clip}(\cdot)$ function to bound the amount of corruption in each spatial frequency. 
The maximum and minimum values are chosen to ensure that perturbations do not affect the semantic content of the images.
The phase perturbation is formulated as $\vdelta_{\text{Fourier}}(f)_{\mathbf{P}}=\text{U}(0,2\pi)$ to simulate real world noises. 
Given each pair $(\vx^i,y^i)$ in the original validation set, we synthesize CIFAR-10/100-F images as
\begin{equation}
\small
    \vx_F^i = \vx^i+ \gamma
\cdot    \text{IFFT}(\vdelta_{\text{Fourier}}),
\end{equation}
where $\gamma = \frac{\epsilon}{||\text{IFFT}(\vdelta_{\text{Fourier}})||_2}$ normalizes the spreading effect of the power-law distribution and, thus, controls the severity level of CIFAR-10/100-F. 
We create both CIFAR-10/100-F with 3 severity levels with $\epsilon \in \{8,10,12\}$. As the images in CIFAR-10/100 are of size $32\times32$, their FFT spectrums have discrete azimuthal frequencies from 0 to 16. Since zero-frequency noise is a constant in the pixel space, we set the center frequency $f_c \in \{1,2,...,16\}$. 
We leverage $\alpha \in \{0.5,1,2,3\}$ because power law noises with $0<\alpha\leq 3$ arise in both natural signals and in man-made processes~\cite{powerlaw}. In total, our CIFAR-10/100-F datasets consists of  $3\times4\times16=192$ test sets from different regions of the frequency spectrum thereby increasing the spectral coverage of the dataset. 
\begin{figure}[t]
\begin{center}
\includegraphics[width=0.485\textwidth]{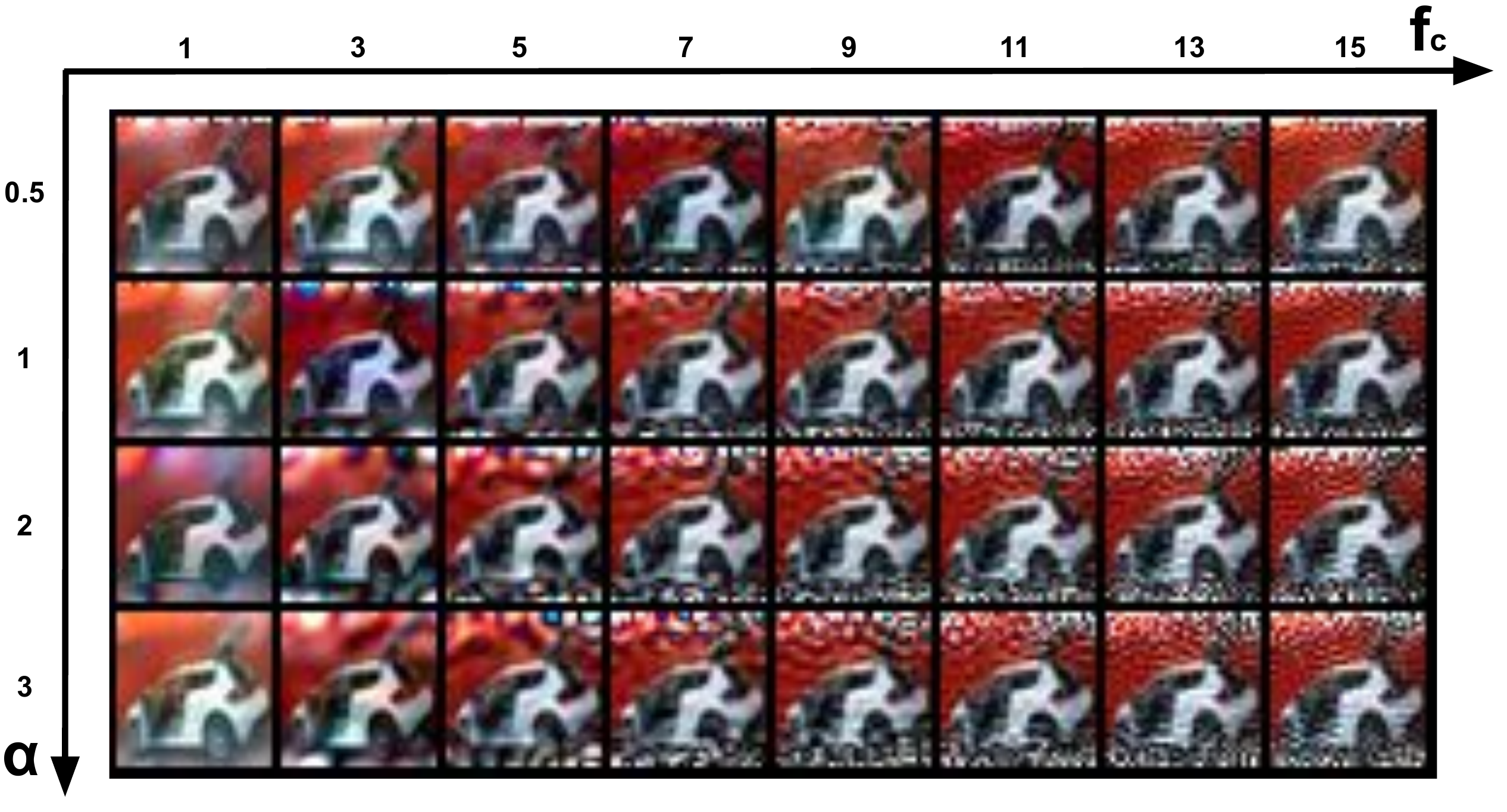}
\vspace{-0.5cm}
\caption{Sample images from CIFAR-10/100-F when $\epsilon=12$.}
\label{fig:cifarf-illu}
\vspace{-0.2cm}
\end{center}
\end{figure}

\noindent \textbf{Visual Effect of Varying $\alpha$ and $f_c$}.\quad To better understand our proposed benchmark, we explain the effect of $\alpha$ and $f_c$ with some sample images.  
As shown in Figure~\ref{fig:cifarf-illu}, $\alpha$ controls the frequency dispersion of the corruption at $f_c$. 
With a smaller $\alpha$, \eg $\alpha=0.5$, the spreading effect of the power law distribution is more significant. 
The corrupted images thus contain noises across all azimuthal frequencies. 
In contrast, for larger $\alpha$, the corruptions will be focused more on a single frequency \eg $\alpha=3$. 
As evident from Figure~\ref{fig:cifarf-illu} higher $f_c$ leads to a higher corruption frequency.
More images from CIFAR-10/100-F from different classes and severity levels are shown in Appendix~\ref{app:sample_cifarf}.  


\subsection{Results on CIFAR-10/100-F}
Figure~\ref{fig:cifar-10-f} reports the performance of models used in \S~\ref{sec:cifar-10/100} on CIFAR-10-F benchmark. 
Our results show that both AutoAugment~\cite{cubuk2019autoaugment} and AugMix~\cite{hendrycks2019augmix} based smoothed models are relatively biased towards low-frequency corruptions. 
The effect of high frequency corruptions is more pronounced on models trained with AutoAugment which behave similar to the simple baseline of Gaussian augmentation (Figures~\ref{fig:cifar-10-f-2} and~\ref{fig:cifar-10-f-3}). 
The intersection of the curves of AugMix+JSD and Gaussian+JSD in the mid frequency region in CIFAR-10-F (Figure~\ref{fig:cifar-10-f}), illustrates the different spectral biases introduced by different augmentation methods. 
Unlike CIFAR10-F, we find that Gaussian and Gaussian+JSD perform relatively worse on CIFAR-100-F compared to other augmentation methods.
In comparison to other methods, we find that models trained with \emph{FourierMix} and HCR do not show significant spectral biases and serve as a strong baseline.
Specifically, models trained with \emph{FourierMix}+HCR, on average,  outperforms AugMix+HCR, by 11.8\% and 16.0\% on CIFAR-10/100-F, respectively.
We emphasize that models trained with \emph{FourierMix} do not overfit to CIFAR-10/100-F datasets since they have different formulations and even visual patterns (see Appendix~\ref{app:fmix-images} and~\ref{app:sample_cifarf}).
Moreover, \emph{FourierMix} models provide consistently better performance on other OOD benchmarks as well, demonstrating its generality.

\section{Discussion and Conclusion}
\label{sec:discussion}


Our work showed that certified defenses are surprisingly brittle to OOD shifts such as low-frequency corruptions. To alleviate this issue, we proposed \emph{FourierMix} augmentation to increase the spectral coverage of the training data.  We also presented a benchmarking suite to gain a comprehensive understanding of the model's OOD robustness.
Some of our findings are consistent with past results in that we also show that the model evaluation in OOD settings is a challenging problem, and one should not rely on a single benchmark~\cite{hendrycks2021many, mintun2021interaction}. However, as opposed to the existing works that focus on empirical robustness, we show that these issues persist and may even be more prominent on the problem of certified adversarial defense. Even though evaluation against all possible types of OOD data is infeasible, our results highlighted that eliminating spectral biases of the models improves the certified robustness on OOD data.

Although we have taken some first steps to address this challenging problem, there are still many questions that remain to be answered. 
First, bridging the gap between robustness guarantees in high-frequency and low-frequency corruption regimes is still an open problem. 
A deeper theoretical understanding of this phenomenon will likely motivate systematic approaches to overcome this issue. 
Next, we encourage future research to pursue test-time adaptation ideas~\cite{mueller2020certify, diffenderfer2021winning} in the context of robustness certification. We expect that designing data-efficient and unsupervised adaptation methods that improve robustness guarantees under OOD shifts can be a worthwhile direction.  
Finally, the analysis done in this work can be explored in the context of certifying other $\ell_p$ norms~\cite{yang2020randomized} and semantic transformations~\cite{li2021tss}. It is also worth noting that we have only focused on incomplete verification techniques in this work, and we expect the OOD brittleness issue to be a general phenomenon affecting complete verification techniques as well~\cite{li2020sok}.

We hope that our work will motivate researchers to study these crucial issues at the intersection of certified adversarial defense and OOD robustness and help the community design ML methods that work reliably in the wild.

\section*{Acknowledgements} 
This work was performed under the auspices of the U.S. Department of Energy by the Lawrence Livermore National Laboratory under Contract No. DE-AC52-07NA27344 and LLNL LDRD Program Project No. 20-ER-014.

{\small
\bibliographystyle{ieee_fullname}
\bibliography{cvpr}
}

\clearpage
\appendix

\begin{table*}[h]
\scriptsize
\renewcommand\arraystretch{0.9}
\setlength\tabcolsep{2.5pt}
  \caption{\small  Average Certified Radius (ACR) of Models Trained with Different Methods on CIFAR-10-C. Models trained with \emph{FourierMix} and HCR achieve significant improvements in the certified robustness (ACR) guarantees on all corruption types from the CIFAR-10-C dataset. 
  }
  \label{tb:cifar10c}
  \vspace{-0.2cm}
  \centering
  \begin{tabular}{|l|c|cccc|ccccccccccccccc|}
    \noalign{\global\arrayrulewidth1pt}\hline\noalign{\global\arrayrulewidth0.4pt}

    Augmentation    &CIFAR-10 & mACR &-Low &-Mid &-High  &Gauss. &Shot  &Impulse  &Defocus   &Glass  &Motion &Zoom &Snow  &Frost  &Fog  &Bright  &Contrast &Elastic &Pixel  &JPEG \\
\noalign{\global\arrayrulewidth1pt}\hline\noalign{\global\arrayrulewidth0.4pt}     
    Gaussian &0.461 & 0.363 &0.301 &0.353 &0.435 &0.448 &0.448 &0.421 &0.380 &0.346 &0.338 &0.357 &0.394 &0.347 &0.187 &0.439 &0.137 &0.342 &0.420 &0.440\\
     ~+JSD &\textbf{0.535} & 0.439 &0.346 &0.451 &\underline{0.520} &\textbf{0.529} &\underline{0.514} &\textbf{0.528} &0.471 &0.445 &0.443 &0.453 &0.449 &0.378 &0.235 &0.485 &0.185 &0.444 &\underline{0.506} &\underline{0.521}\\
    \hline
     ~+AutoAugment &0.411  & 0.372 &0.312 &0.364 &0.431 &0.451 &0.452 &0.419 &0.411 &0.356 &0.342 &0.360 &0.403 &0.354 &0.201 &0.446 &0.158 &0.352 &0.429 &0.445\\
      ~~+JSD  &0.432 & 0.400 &0.343 &0.395 &0.464 &0.473 &0.476 &0.443 &0.423 &0.385 &0.394 &0.390 &0.427 &0.403 &0.212 &0.483 &0.189 &0.382 &0.453 &0.473\\
    \hline
     ~+AugMix &0.452 & 0.385 &0.324 &0.383 &0.449 &0.459 &0.460 &0.436 &0.412 &0.369 &0.372 &0.391 &0.413 &0.374 &0.216 &0.457 &0.159 &0.371 &0.439 &0.453\\
      ~~+JSD &0.518 & 0.430 &0.357 &0.436 &0.496 &0.504 &0.507 &0.481 &0.461 &0.426 &0.429 &0.441 &0.452 &0.408 &0.240 &0.501 &0.185 &0.425 &0.485 &0.502\\
      ~~+\textbf{HCR} &0.520 &0.437 &0.369 &0.444 &0.497 &0.505 &0.506 &0.484 &0.464 &0.438 &0.435 &0.447 &0.460 &0.426 &0.252 &0.505 &0.200 &0.437 &0.487 &0.501\\
    \hline
     ~+\textbf{\emph{FourierMix}} &0.455 & 0.388 &0.326 &0.386 &0.453 &0.461 &0.462 &0.446 &0.417 &0.369 &0.378 &0.393 &0.415 &0.376 &0.220 &0.457 &0.160 &0.373 &0.439 &0.456\\
      ~~+JSD &\underline{0.522} & \underline{0.444} &\underline{0.375} &\underline{0.454} &0.504 &0.512 &0.513 &0.491 &\underline{0.474} &\underline{0.448} &\underline{0.446} & \underline{0.456} &\underline{0.464} &\underline{0.432} &\underline{0.257} &\textbf{0.519} &\underline{0.201} &\underline{0.445} &0.495 &0.508\\
      
      ~~+\textbf{HCR} &\textbf{0.535} & \textbf{0.460} &\textbf{0.384} &\textbf{0.473} &\textbf{0.521} &\underline{0.528} &\textbf{0.530} &\underline{0.513} &\textbf{0.492} &\textbf{0.470} &\textbf{0.464} &\textbf{0.477} &\textbf{0.477} &\textbf{0.432} &\textbf{0.275} &\underline{0.517} &\textbf{0.220} &\textbf{0.462} &\textbf{0.511} &\textbf{0.524}\\
\noalign{\global\arrayrulewidth1pt}\hline\noalign{\global\arrayrulewidth0.4pt}
  \end{tabular}

\end{table*}

\begin{figure*}[t]
    \centering
    \vspace{-0.25cm}
    \includegraphics[width=0.85\linewidth]{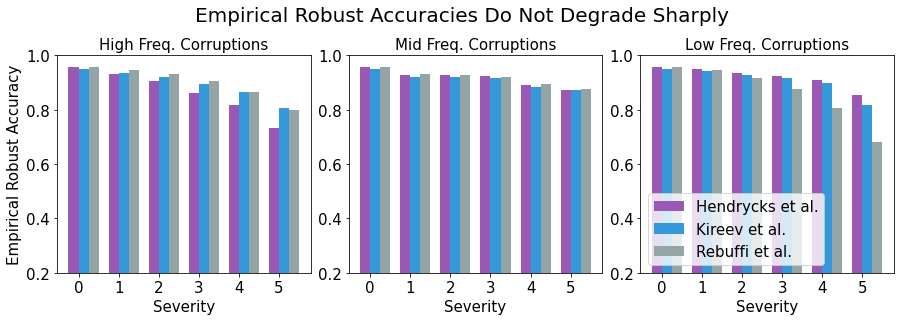}
    \vspace{-0.25cm}
    \caption{The performance gaps (\ie the robust accuracy) are remained small/reasonable in state-of-the-art empirically robust models~\cite{hendrycks2019augmix,kireev2021effectiveness,rebuffi2021fixing}. Severity 0 denotes the in-distribution data.}
    \vspace{-0.3cm}
    \label{fig:motivation2}
\end{figure*}

\begin{figure}[h]
    \centering
    \includegraphics[width=0.95\linewidth]{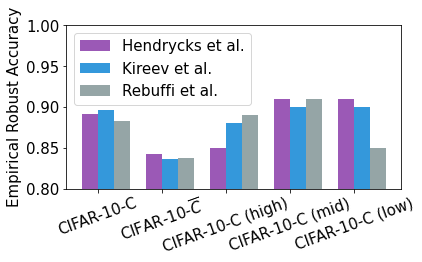}
    \vspace{-0.3cm}
    \caption{The ranking of SOTA models~\cite{hendrycks2019augmix,kireev2021effectiveness,rebuffi2021fixing} (based on empirical robust accuracy) changes across datasets and corruption types, suggesting there is no single model which performs the best on different OOD benchmarks.}
    \label{fig:robustbench}
\end{figure}

\begin{table*}[h]
\footnotesize
\renewcommand\arraystretch{0.9}
\setlength\tabcolsep{2.5pt}
  \caption{\small Average Certified Radius (ACR) of Models Trained with Different Methods on CIFAR-10-$\bar{\text{C}}$. Models trained with \emph{FourierMix} and HCR achieve significant improvements in the certified robustness (ACR) guarantees on corruptions from the CIFAR-10-$\bar{\text{C}}$ dataset. 
  }
  \label{tb:cifar10c-bar}
  \vspace{-0.2cm}
  \centering
  \begin{tabular}{|l|c|cccccccccc|}
    \noalign{\global\arrayrulewidth1pt}\hline\noalign{\global\arrayrulewidth0.4pt}
    Augmentation     & mACR  &Blue &Brown  &Checkerboard  &Circular   &Inv. Sparkle  &Lines &Pinch &Ripple  &Sparkles  &Trans. Chromatic  \\
\noalign{\global\arrayrulewidth1pt}\hline\noalign{\global\arrayrulewidth0.4pt}     
    Gaussian & 0.314  &0.351 &0.255 &0.310 &0.386 &0.222 &0.336 &0.398 &0.365 &0.251 &0.269\\
     ~+JSD & 0.393 &\underline{0.458} &0.303 &\underline{0.395} &0.452 &0.252 &\underline{0.430} &\underline{0.492} &0.463 &0.306 &0.376\\
    \hline
     ~+AutoAugment & 0.304  & 0.351 &0.263 &0.312 &0.395 &0.223 &0.348 &0.406 &0.248 &0.235 &0.256\\
      ~~+JSD & 0.346 &0.354 &0.297 &0.335 &0.445 &0.238 &0.374 &0.436 &0.402 &0.269 &0.308\\
    \hline
     ~+AugMix & 0.341  &0.389 &0.269 &0.334 &0.439 &0.233 &0.358 &0.416 &0.397 &0.272 &0.307\\
      ~~+JSD & 0.382  &0.429 &0.303 &0.372 &0.483 &0.255 &0.404 &0.467 &0.450 &0.306 &0.350\\
      ~~+\textbf{HCR} &0.393  &0.442 &\underline{0.309} &0.384 &\underline{0.486} &\underline{0.268} &0.419 &0.471 &\underline{0.464} &\underline{0.320} &0.368\\
    \hline
     ~+\textbf{\emph{FourierMix}} & 0.348  &0.391 &0.269 &0.331 &0.441 &0.237 &0.368 &0.432 &0.401 &0.280 &0.325\\
      ~~+JSD & \underline{0.397} &0.445 &0.307 &\underline{0.395} &0.482 &0.265 &\underline{0.430} &0.490 &0.463 &\underline{0.320} &\underline{0.377}\\
      ~~+\textbf{HCR} & \textbf{0.419}  &\textbf{0.474} &\textbf{0.317} &\textbf{0.418} &\textbf{0.504} &\textbf{0.289} &\textbf{0.459} &\textbf{0.501} &\textbf{0.486} &\textbf{0.339} &\textbf{0.406}\\
    \noalign{\global\arrayrulewidth1pt}\hline\noalign{\global\arrayrulewidth0.4pt}
  \end{tabular}
\end{table*}

\begin{center}
{\LARGE \bf Appendix}
\end{center}

\begin{table*}[t]
\scriptsize
\renewcommand\arraystretch{0.9}
\setlength\tabcolsep{2.0pt}
  \caption{\small Average Certified Radius (ACR) of Models Trained with Different Methods on CIFAR-100-C. Models trained with \emph{FourierMix} and HCR achieve significant improvements in the certified robustness (ACR) guarantees on corruptions from the CIFAR-100-C dataset. 
  }
   \label{tb:cifar100c}
   \vspace{-0.2cm}
  \centering
  \begin{tabular}{|l|c|cccc|ccccccccccccccc|}
    \noalign{\global\arrayrulewidth1pt}\hline\noalign{\global\arrayrulewidth0.4pt}
    Augmentation  & CIFAR-100   & mACR &-Low &-Mid &-High &Gauss. &Shot  &Impulse  &Defocus   &Glass  &Motion &Zoom &Snow  &Frost  &Fog   &Bright  &Contrast &Elastic &Pixel  &JPEG \\
    \noalign{\global\arrayrulewidth1pt}\hline\noalign{\global\arrayrulewidth0.4pt}
    Gaussian & 0.238 & 0.169 &0.131 &0.182 &0.208 & 0.214 &0.218 &0.193 &0.181 &0.170 &0.157 &0.169 &0.177 &0.153 &0.069 &0.207 &0.051 &0.159 &0.206 &0.209\\
    ~+JSD & 0.291 & 0.232 &0.167 &0.248 &0.280 &0.283 &0.285 &0.273 &0.261 &0.252 &0.240 &0.250 &0.226 &0.188 &0.104 &0.242 &0.079 &0.235 &0.278 &0.281\\
    \hline
     ~+AutoAugment + JSD &0.265 & 0.225 &0.175 &0.234 &0.265 &0.275 &0.273 &0.252 &0.248 &0.230 &0.230 &0.238 &0.232 &0.202 &0.104 &0.257 &0.082 &0.225 &0.261 &0.266\\
    \hline
     ~+AugMix + JSD & 0.286 & 0.231 &0.184 &0.240 &0.269 &0.274 &0.278 &0.256 &0.255 &0.236 &0.233 &0.243 &0.239 &0.211 &0.111 &0.267 &0.092 &0.232 &0.267 &0.270\\
     ~+AugMix + \textbf{HCR} &\underline{0.296}  &\underline{0.249} &\underline{0.191} &\underline{0.263} &\underline{0.292} &\underline{0.296} &\underline{0.301} &0.282 &\underline{0.278} &\underline{0.264} &\underline{0.255} &\underline{0.263} &0.249 &0.215 &\underline{0.118} &0.274 &\underline{0.097} &\underline{0.253} &\underline{0.291} &\underline{0.292}\\
    \hline
     ~+\emph{FourierMix} + JSD &0.295 & 0.247 &0.190 &0.258 &\underline{0.292} &0.295 &0.300 &\underline{0.283} &0.273 &0.257 &0.249 &0.260 &\underline{0.251} &\underline{0.217} &0.115 &\underline{0.275} &0.092 &0.250 &0.288 &\underline{0.292}\\
     ~+\textbf{\emph{FourierMix} + HCR} &\textbf{0.309} & \textbf{0.261} &\textbf{0.199} &\textbf{0.278} &\textbf{0.307} &\textbf{0.310} &\textbf{0.313} &\textbf{0.302} &\textbf{0.291} &\textbf{0.283} &\textbf{0.270} &\textbf{0.277} &\textbf{0.260} &\textbf{0.221} &\textbf{0.128} &\textbf{0.284} &\textbf{0.102} &\textbf{0.267} &\textbf{0.303} &\textbf{0.307}\\
\noalign{\global\arrayrulewidth1pt}\hline\noalign{\global\arrayrulewidth0.4pt}
  \end{tabular}
\end{table*}

\begin{table*}[t]
\footnotesize
\renewcommand\arraystretch{0.9}
\setlength\tabcolsep{3pt}
  \caption{\small Average Certified Radius (ACR) of Models Trained with Different Methods on CIFAR-100-$\bar{\text{C}}$. Models trained with \emph{FourierMix} and HCR achieve significant improvements in the certified robustness (ACR) guarantees on corruptions from the CIFAR-100-$\bar{\text{C}}$ dataset. }
  \label{tb:cifar100c-bar}
  \vspace{-0.2cm}
  \centering
  \begin{tabular}{|l|c|cccccccccc|}
\noalign{\global\arrayrulewidth1pt}\hline\noalign{\global\arrayrulewidth0.4pt}
    Augmentation     & mACR  &Blue &Brown  &Checkerboard  &Circular   &Inv. Sparkle  &Lines &Pinch &Ripple  &Sparkles  &Trans. Chromatic   \\
\noalign{\global\arrayrulewidth1pt}\hline\noalign{\global\arrayrulewidth0.4pt}
    Gaussian  & 0.130 &0.151 &0.070 &0.114 &0.159 &0.097 &0.137 &0.199 &0.160 &0.097 &0.116\\
    ~+JSD & 0.196 &0.228 &0.106 &0.186 &0.233 &0.124 &0.221 &0.274 &0.242 &0.151 &0.193\\
    \hline
     ~+AutoAugment + JSD & 0.176 & 0.211 &0.087 &0.152 &0.229 &0.119 &0.184 &0.236 &0.217 &0.140 &0.185\\
    \hline
     ~+AugMix + JSD & 0.193 &0.227 &0.107 &0.176 &0.259 &0.131 &0.206 &0.260 &0.244 &0.153 &0.191\\
     ~+AugMix + \textbf{HCR} &\underline{0.211}  &\underline{0.253} &\underline{0.120} &\underline{0.199} &\underline{0.276} &\underline{0.136} &0.224 &\underline{0.283} &\underline{0.263} &\underline{0.156} &0.203\\
    \hline
    ~+\textbf{\emph{FourierMix}} + JSD & 0.207 &0.243 &0.106 &0.194 &0.262 &\underline{0.136} &\underline{0.226} &0.281 &0.258 &0.154 &\underline{0.205}\\
     ~+\textbf{\emph{FourierMix} + HCR} & \textbf{0.227} &\textbf{0.260} &\textbf{0.129} &\textbf{0.219} &\textbf{0.281} &\textbf{0.151} &\textbf{0.247} &\textbf{0.300} &\textbf{0.278} &\textbf{0.172} &\textbf{0.228}\\
\noalign{\global\arrayrulewidth1pt}\hline\noalign{\global\arrayrulewidth0.4pt}
  \end{tabular}
  \vspace{-0.2cm}
\end{table*}

\section{Amplitude Spectrum of CIFAR-10-C/$\overline{\text{C}}$}
\label{psd_cifar-10-c}

\begin{figure*}[t]
\begin{center}
\vspace{-0.2cm}
\includegraphics[width=\linewidth]{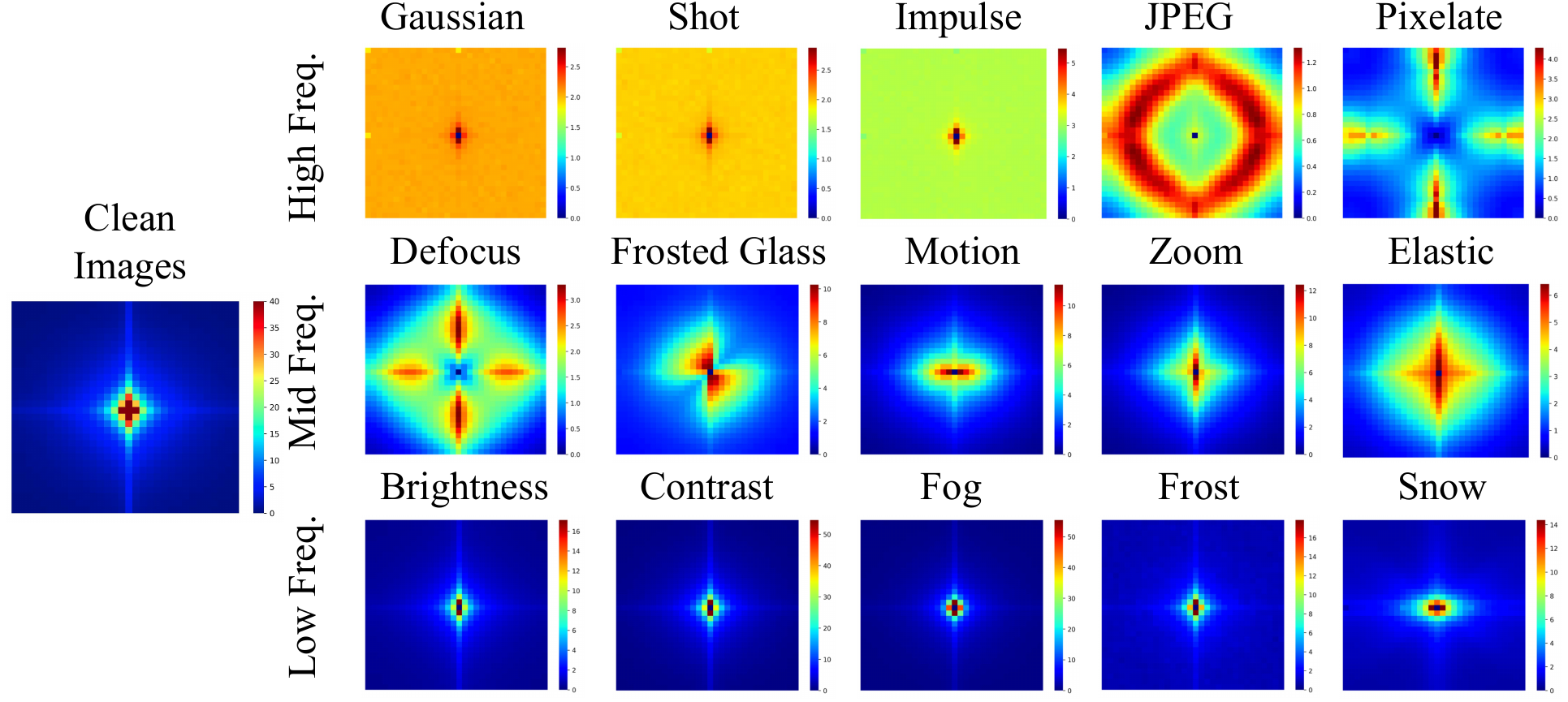} 
\vspace{-0.3cm}
\caption{Amplitude Spectrum $\mA$ of Different Corruptions in CIFAR-10/100-C with severity 3.}
\label{fig:psd_cifar10c}
\end{center}
\vspace{-0.2cm}
\end{figure*}

As introduced in \S~\ref{sec:rs_on_c}, we arrange the amplitude specturm of corruptions from CIFAR-10-C into three groups, roughly categorized as high/mid/low-frequency corruptions. Specifically, we compute the the $\mathbb{E}[\text{FFT}(\vx)]$ and $\mathbb{E}[\text{FFT}(C(\vx)=\vx)]$ by averaging over all the validation images~\cite{yin2019fourier} for CIFAR-10 and each corruption in CIFAR-10-C, respectively, where $C(\cdot)$ denotes the corruption function. As Figure~\ref{fig:psd_cifar10c} shows, CIFAR-10 (clean) images follow a distribution of $\frac{1}{f^\alpha}$, where $f=\sqrt{u^2+v^2}$ is the azimuthal frequency and $\alpha \approx 2$. Therefore, clean images have extremely low power in the high-frequency regions (the edges and corner). Due to this, all the noise perturbations corresponding to \texttt{JPEG} and \texttt{pixelate} can be considered as high-frequency corruptions, relative to the clean images' distribution. On the other hand, weather-related and \texttt{contrast} corruptions are all centered in the low-frequency region. We categorize remaining perturbations as mid-frequency corruptions. 

\begin{figure*}[t]
\begin{center}
\vspace{-0.2cm}
\includegraphics[width=0.9\linewidth]{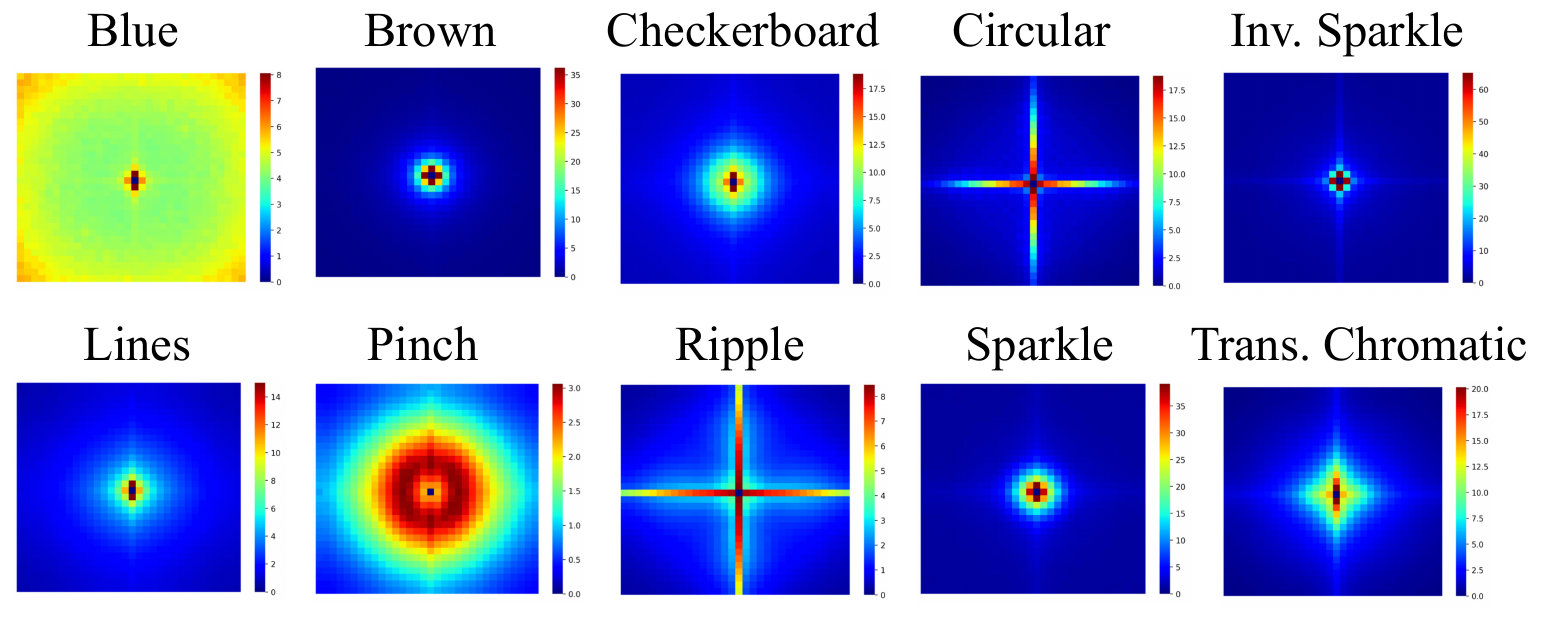} 
\vspace{-0.3cm}
\caption{Amplitude Spectrum $\mA$ of Different Corruptions in CIFAR-10/100-$\bar{\text{C}}$ with severity 3.}
\label{fig:psd_cifar10cbar}
\end{center}
\vspace{-0.2cm}
\end{figure*}

We also visualize the amplitude spectrum of corruptions from CIFAR-10-$\bar{\text{C}}$ in Figure~\ref{fig:psd_cifar10cbar}. We find that most of the corruptions from CIFAR-10-$\bar{\text{C}}$ are centered in the low/mid-frequency ranges, explaining why \emph{FourierMix} achieves lager improvements on CIFAR-10-$\bar{\text{C}}$ than CIFAR-10-C compared to spectrally-biased baselines.

\section{Empirical Robust Accuracy of SOTA Models on Corrupted Data}
\label{app:robustbench}
{
The results in Figure~\ref{fig:robustbench} show empirical robust accuracy of state-of-the-art models on existing OOD benchmarks. We use the recently proposed RobustBench~\cite{croce2020robustbench} benchmark and selected the top-performing models on CIFAR-10-C for this experiment~\cite{kireev2021effectiveness, hendrycks2019augmix,rebuffi2021fixing}. As evident from the figure, the performance of the models varies across datasets and corruption types showing that a single model is not able to achieve the best performance on all types of OOD data. Evaluating the models on a single benchmark is not enough to obtain the true picture of the OOD robustness of a model. Thus to eliminate the biases present in OOD benchmarks, one should gauge the OOD robustness of a model by evaluating it on a variety of datasets. Our proposed CIFAR-10/100-F benchmark can be used by designers to probe the spectral biases of the models.
}

\section{Detailed Evaluation Results}

\subsection{CIFAR-10-Based OOD Benchmarks}
\label{app:detailed_cifar10}
In this section, we present detailed results for our evaluation on CIFAR-10-C/$\bar{\text{C}}$.  We fix $\eta=10$ and use $\lambda=40$ for HCR (Equation~\ref{eq:hcr}) in our experiments on CIFAR-10. Tables~\ref{tb:cifar10c} and~\ref{tb:cifar10c-bar} present the ACR on individual corruption types from CIFAR-10-C/$\bar{\text{C}}$, respectively. \emph{FourierMix} consistently achieves the highest ACR on most of the corruption types in both OOD datasets. Especially, we find \emph{FourierMix} helps achieve larger improvements on weather-related corruptions, which have real-world implications (\eg safety of autonomous driving). 


\begin{table*}[t]
\scriptsize
\renewcommand\arraystretch{0.9}
\setlength\tabcolsep{2pt}
  \caption{\small Average Certified Radius (ACR) of Models Trained with Different Methods on ImageNet-C. Spectrally diverse augmentations from \emph{FourierMix} brings significant gains to certified robustness of the models trained on ImageNet against corruptions from ImageNet-C.
  }
  \label{tb:imagenet-c}
  \vspace{-0.2cm}
  \centering
  \begin{tabular}{|l|c|cccc|ccccccccccccccc|}
\noalign{\global\arrayrulewidth1pt}\hline\noalign{\global\arrayrulewidth0.4pt}
    Augmentation  &ImageNet & mACR &-Low &-Mid &-High &Gauss. &Shot  &Impulse  &Defocus   &Glass  &Motion &Zoom &Snow &Frost  &Fog   &Bright  &Contrast &Elastic &Pixel  &JPEG \\
\noalign{\global\arrayrulewidth1pt}\hline\noalign{\global\arrayrulewidth0.4pt}
    Gaussian &0.600 & 0.256 &0.155 &0.228 &0.385 & 0.342 &0.324 &0.310 &0.174 &0.227 &0.212 &0.201 &0.148 &0.170 &0.013 &0.419 &0.027 &0.325 &0.440 &0.507\\
    ~+JSD & 0.736 & 0.395 &0.220 &0.382 &\textbf{0.581} & \textbf{0.537} &\textbf{0.519} &\textbf{0.508} &0.289 &0.378 &0.351 &\textbf{0.374} &\underline{0.254} &0.245 &0.013 &\textbf{0.551} &0.039 &\textbf{0.518} &\textbf{0.640} &\textbf{0.702}\\
    \hline
     ~+AugMix + JSD & 0.717 & 0.391 &0.238 &\underline{0.387} &0.550 & 0.496 &0.489 &0.473 &\textbf{0.329} &\textbf{0.395} &0.376 &0.352 &\textbf{0.255} &0.286 &\textbf{0.041} &0.542 &0.064 &0.481 &0.622 &0.668\\
     ~+AugMix + \textbf{HCR} &0.727 &0.390 &0.234 &0.383 &0.552 &0.500 &0.494 &0.480 &\underline{0.320} &\underline{0.391} &0.374 &0.349 &0.249 &0.283 &\underline{0.040} &0.539 &0.061 &0.481 &0.624 &0.662\\
    \hline
     ~+\textbf{\emph{FourierMix}} + JSD &\textbf{0.751} &\textbf{0.399} &\textbf{0.242} &\textbf{0.389} &0.564 &0.515 &0.493 &0.483 &0.315 &0.384 &\textbf{0.380} &\underline{0.370} &\underline{0.254} &\textbf{0.300} &\textbf{0.041} &\underline{0.544} &\textbf{0.073} &\underline{0.497} &\underline{0.637} &\underline{0.694}\\
     ~+\textbf{\emph{FourierMix} + HCR} & \underline{0.750} &\underline{0.397} &\underline{0.239} &\underline{0.387} &\underline{0.567} &\underline{0.518} &\underline{0.499} &\underline{0.492} &0.312 &0.382 &\underline{0.377} &\underline{0.370} &0.249 &\underline{0.295} &0.039 &\underline{0.544} &\underline{0.069} &0.494 &\underline{0.637} &0.689\\
\noalign{\global\arrayrulewidth1pt}\hline\noalign{\global\arrayrulewidth0.4pt}
  \end{tabular}
\end{table*}

\begin{table*}[t]
\footnotesize
\renewcommand\arraystretch{0.9}
\setlength\tabcolsep{4pt}
  \caption{\small Average Certified Radius (ACR) of Models Trained with Different Methods on ImageNet-$\bar{\text{C}}$. Spectrally diverse augmentations from \emph{FourierMix} brings significant gains to certified robustness of the models trained on ImageNet against corruptions from ImageNet-$\bar{\text{C}}$.}
  \label{tb:imagenet-c-bar}
  \vspace{-0.2cm}
  \centering
  \begin{tabular}{|l|c|cccccccccc|}
\noalign{\global\arrayrulewidth1pt}\hline\noalign{\global\arrayrulewidth0.4pt}
    Augmentation  & mACR & Blue & Brown &Caustic &Checkboard &Cocentric &Inv. Sparkle &Perlin &Plasma &Single Freq. &Sparkle\\
\noalign{\global\arrayrulewidth1pt}\hline\noalign{\global\arrayrulewidth0.4pt}
    Gaussian &0.266 &0.394 &0.284 &0.325 &0.250 &0.235 &0.152 &0.274 &0.065 &0.284 &0.400\\
    ~+JSD &0.395 &\textbf{0.579} &0.395 &\textbf{0.512} &\textbf{0.370} &\textbf{0.374} &\textbf{0.224} &0.404 &0.113 &0.408 &\textbf{0.567}\\
    \hline
     ~+AugMix + JSD &0.379 &0.560 &0.381 &0.461 &\underline{0.365} &0.342 &0.212 &\textbf{0.413} &0.121 &0.397 &0.538\\
     ~+AugMix + \textbf{HCR} &0.378 &0.563 &0.377 &0.464 &0.361 &0.342 &0.210 &\underline{0.410} &0.115 &0.396 &0.539\\
    \hline
     ~+\textbf{\emph{FourierMix}} + JSD &\textbf{0.413} &0.562 &\textbf{0.544} &0.479 &\textbf{0.370} &0.366 &\underline{0.215} &\textbf{0.413} &\textbf{0.227} &\textbf{0.417} &0.547\\
     ~+\textbf{\emph{FourierMix} + HCR} &\underline{0.411} &\underline{0.565} &\underline{0.535} &\underline{0.481} &\underline{0.365} &\underline{0.367} &0.210 &0.408 &\underline{0.215} &\underline{0.415} &\underline{0.550}\\
\noalign{\global\arrayrulewidth1pt}\hline\noalign{\global\arrayrulewidth0.4pt}
  \end{tabular}
\end{table*}

\begin{table*}[t]
\footnotesize
\renewcommand\arraystretch{1.0}
\setlength\tabcolsep{2.5pt}
  \caption{\small Average Certified Radius (ACR) of In-distribution (CIFAR10) and OOD (CIFAR10-C) Data with $\sigma=0.25$ Using SOTA Certified Defense Methods.}
  \vspace{-0.2cm}
  \centering
  \begin{tabular}{|l|c|c|ccccccccccccccc|}
\noalign{\global\arrayrulewidth1pt}\hline\noalign{\global\arrayrulewidth0.4pt}
    Method & clean & mACR  &Gauss. &Shot  &Impulse  &Defocus   &Glass  &Motion &Zoom &Snow  &Frost &Fog  &Bright  &Contrast &Elastic &Pixel  &JPEG \\
    \hline
     Gaussian & 0.461 & 0.363 &0.448 &0.448 &0.421 &0.380 &0.346 &0.338 &0.357 &0.394 &0.347 &0.187 &0.439 &0.137 &0.342 &0.420 &0.440\\
    \hline
     MACER & \textbf{0.539} &\textbf{0.426} & \textbf{0.509} &\textbf{0.509} &\textbf{0.492} & \textbf{0.460} &\textbf{0.436} &0.422 &\textbf{0.433} &\textbf{0.443} &\textbf{0.381} &\textbf{0.232} &\textbf{0.477} &\textbf{0.185} &\textbf{0.428} &\textbf{0.490} &\textbf{0.503}\\
    \hline
     SmoothAdv & 0.519 &0.411 &0.483 &0.485 &0.471 &0.448 &0.426 &\textbf{0.423} &0.425 &0.418 &0.361 &0.222 &0.451 &0.175 &0.415 &0.472 &0.483\\
\noalign{\global\arrayrulewidth1pt}\hline\noalign{\global\arrayrulewidth0.4pt}
  \end{tabular}
\end{table*}

\begin{table*}[h]
\footnotesize
\renewcommand\arraystretch{1.0}
\setlength\tabcolsep{2.5pt}
  \caption{\small Failure of Existing Methods: Average Certified Radius (ACR) of CIFAR10-C with $\sigma=0.25$ Using Test-Time Adaptation.}
  \label{tb:testtime}
  \vspace{-0.2cm}
  \centering
  \begin{tabular}{|l|c|ccccccccccccccc|}
\noalign{\global\arrayrulewidth1pt}\hline\noalign{\global\arrayrulewidth0.4pt}
    Adaptation & mACR  &Gauss. &Shot  &Impulse  &Defocus   &Glass  &Motion &Zoom &Snow  &Frost &Fog  &Bright  &Contrast &Elastic &Pixel  &JPEG \\
    \hline
     Gaussian & \textbf{0.363} &\textbf{0.448} &\textbf{0.448} &\textbf{0.421} &\textbf{0.380} &\textbf{0.346} &\textbf{0.338} &\textbf{0.357} &\textbf{0.394} &\textbf{0.347} &\textbf{0.187} &\textbf{0.439} &\textbf{0.137} &\textbf{0.342} &\textbf{0.420} &\textbf{0.440}\\
    \hline
     ~+BN& 0.356 &0.441 &0.442 &0.417 &0.369 &0.338 &0.326 &0.345 &0.392 &\textbf{0.347} &0.181 &0.436 &0.133 &0.332 &0.411 &0.432\\
    \hline
     ~+TENT & 0.357 &0.442 &0.442 &0.419 &0.369 &0.337 &0.328 &0.346 &\textbf{0.394} &0.345 &0.182 &0.436 &0.132 &0.330 &0.412 &0.434\\
\noalign{\global\arrayrulewidth1pt}\hline\noalign{\global\arrayrulewidth0.4pt}
  \end{tabular}
\end{table*}

\subsection{CIFAR-100-Based OOD Benchmarks}
\label{app:detailed_cifar100}

In this section, we present detailed results for our evaluation on CIFAR-100-C/$\bar{\text{C}}$ and CIFAR-100-F. We fix $\eta=10$ and use $\lambda=20$ for HCR in our experiments on CIFAR-100. Tables~\ref{tb:cifar100c} and~\ref{tb:cifar100c-bar} present the ACR on individual corruption types from CIFAR-100-C/$\bar{\text{C}}$, respectively. CIFAR-100 is more difficult for RS-based certification compared to CIFAR-10. We find that \emph{FourierMix}+HCR helps achieve the highest ACR on \emph{all} corruption types in both datasets with significant enhancements compared to existing augmentation methods.


\subsection{ImageNet-Based OOD Benchmarks}
\label{app:detailed_imagenet}

ImageNet appears to be the most challenging dataset for certified defenses, to which only RS-based techniques can be applied ~\cite{cohen2019certified}. We select representative combinations of augmentations and regularization schemes that perform well on CIFAR-10/100 for our experiments on ImageNet. We exclude the input normalization layer, which trades off the ACR on clean data for the ACR on OOD data. We use $\eta=5$ and $\lambda=5$ for our experiments with HCR. Tables~\ref{tb:imagenet-c} and~\ref{tb:imagenet-c-bar} present the detailed results on our evaluation on ImageNet-C/$\bar{\text{C}}$. Note the the corruption types in ImageNet-$\bar{\text{C}}$ are different from the ones in CIFAR-10/100-$\bar{\text{C}}$. We find that the spectral biases of other baselines become much more noticeable on ImageNet-based OOD benchmarks. Gaussian+JSD accomplishes the highest ACR on high-frequency corruptions, while AugMix+JSD performs the best on several low-frequency corruptions in ImageNet-C. As RS-based models generally suffer performance degradation on low-frequency corruptions, Gaussain+JSD beats AugMix+JSD in terms of overall mACR. However, \emph{FourierMix} performs well across the spectrum, reaching the highest mACR on both datasets. Although tangible improvements have been realized by \emph{FourierMix} on ImageNet-based OOD benchmarks, we want to highlight that there is still large room for future research to improve over our baselines. We hope this work will motivate more studies on certified defenses for ImageNet under OOD shifts, as discussed in~\S~\ref{sec:discussion}.

\section{\emph{FourierMix} Details}
\label{app:fmix-images}
\noindent\textbf{Hyper-parameter Settings.} We detail the chosen hyper-parameters used in the experiments with \emph{FourierMix}. As illustrated in Algorithm~\ref{alg:fmix} and Equations~\ref{eq:fouriermix1} and~\ref{eq:fouriermix2}, we leverage 5 different severity levels and truncated Gaussian distribution. We use a large $\sigma=5$ for the truncated Gaussian distribution to make \emph{FourierMix} render more diverse augmentation. For CIFAR-10/100, we set $s_\mathbf{A} \in [0.2,0.3,0.4,0.5,0.6]$ and $s_\mathbf{P} \in [\frac{\pi}{12},\frac{\pi}{10},\frac{\pi}{8},\frac{\pi}{6},\frac{\pi}{4}]$ as the 5 severity levels in Equations~\ref{eq:fouriermix1} and~\ref{eq:fouriermix2}, respectively. For ImageNet, we use the same set of $s_\mathbf{A}$ and set $s_\mathbf{P} \in [\frac{\pi}{4},\frac{3\pi}{10},\frac{3\pi}{8},\frac{\pi}{2},\frac{3\pi}{4}]$ since high-resolution images can tolerate more perturbations in the phase spectrum.

\begin{figure*}
    \centering
    \includegraphics[width=0.9\linewidth]{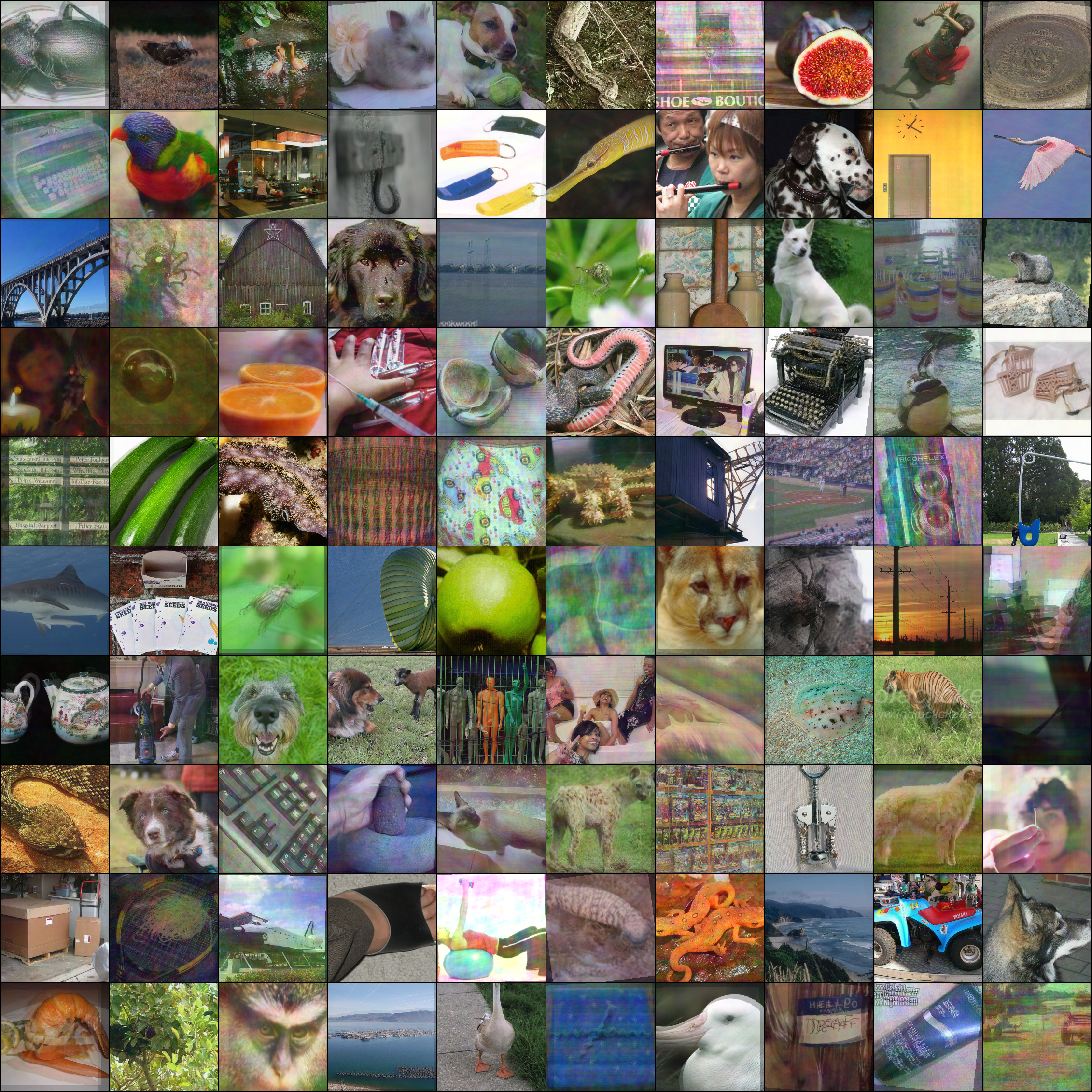}
    \caption{Sample Images from \emph{FourierMix} Data Augmentation on ImageNet. To better highlight the visual patterns of \emph{FourierMix}, we utilize the highest severity level for $\mathbf{A}(\cdot)$ and $\mathbf{P}(\cdot)$ in this figure.}
    \label{fig:fmiximagenet}
\end{figure*}

\noindent\textbf{Sample Images from \emph{FourierMix}}.\quad We visualize randomly sampled images from CIFAR-10/100 and ImageNet in Figure~\ref{fig:fmixcifar10},~\ref{fig:fmixcifar100}, and~\ref{fig:fmiximagenet}, respectively.

\section{Training and Evaluation Details}
\label{app:training-details}

\noindent\textbf{Training.}\quad We train CIFAR-10/100 and ImageNet models for 200 and 90 epochs for all methods with an SGD optimizer, respectively~\cite{ruder2016overview}.  We exclude the input normalization layer as it will degrade the certification performance on OOD data. We use different $\sigma$ to train CIFAR-10/100 and ImageNet models, as specified in~\S~\ref{sec:experiment}.

\noindent\textbf{Evaluation.}\quad Recall from the theorem derived in \S~\ref{sec:rs_on_c} of Cohen~\etal, $\text{CR}(\cdot)$ approaches $\infty$ when $\underline{p_A}$ approaches the value $1$~\cite{cohen2019certified}. However, this will also require the Gaussian perturbed samples $n \approx \infty$. Consider that the base classifier $\mathcal{M}(\vx+\vdelta)$ has observed $n$ samples that all equal to $c_A$, $p_A \geq \alpha^{(1/n)}$ has a probability $1-\alpha$~\cite{cohen2019certified}. To both constrain the computational complexity and achieve a tight bound,
we use $n=100,000$, $n_0=100$, and $\alpha=0.001$ as the hyper-parameters to get high confidence of the computed radius, following prior arts~\cite{cohen2019certified,Zhai2020MACER,salman2019provably,jeong2020consistency}. Since we need to evaluate OOD datasets with $125 \times$ larger sizes than the original test sets, we certify 500 and 350 examples from each corruption and each severity level of the CIFAR-10/100 and ImageNet OOD datasets (\ie -C/$\bar{\text{C}}$). For the Fourier sensitivity analysis of CIFAR-10/100, each data point in the heat map is the corresponding ACR of 200 examples.


\section{Sample Images from CIFAR-10/100-F}
\label{app:sample_cifarf}

\begin{figure*}[t]
\subfigure[Examples of the Ship Class]{\begin{minipage}[t]{\linewidth}
\begin{center}
\includegraphics[width=0.9\linewidth]{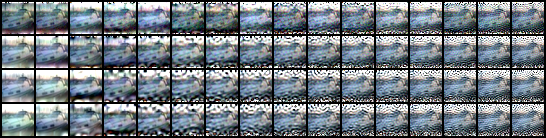} 
\end{center}
\end{minipage}}
\subfigure[Examples of the Airplane Class]{\begin{minipage}[t]{\linewidth}
\begin{center}
\includegraphics[width=0.9\linewidth]{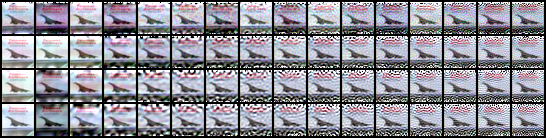} 
\end{center}
\end{minipage}}
\subfigure[Examples of the Bird Class]{\begin{minipage}[t]{\linewidth}
\begin{center}
\includegraphics[width=0.9\linewidth]{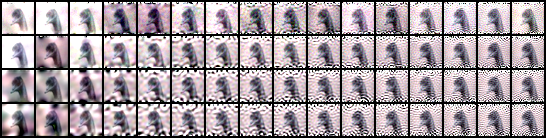} 
\end{center}
\end{minipage}}
\subfigure[Examples of the Horse Class]{\begin{minipage}[t]{\linewidth}
\begin{center}
\includegraphics[width=0.9\linewidth]{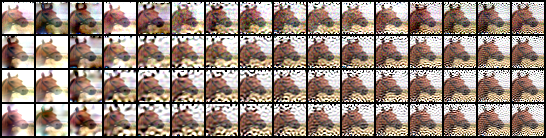} 
\end{center}
\end{minipage}}
\vspace{-0.35cm}
\caption{Sample Images from CIFAR-10/100-F with $\epsilon=12$. From top down row-wise, the images are from $\alpha \in \{0.5,1,2,3\}$ and from left to right column-wise, the images are from $f_c \in \{1,2,...,16\}$}
\label{fig:fsamples}
\end{figure*}

\begin{figure*}
    \centering
    \includegraphics[width=0.9\linewidth]{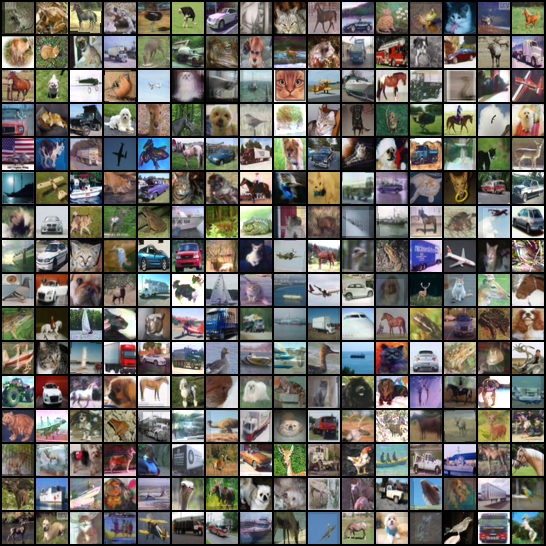}
    \caption{Sample Images from \emph{FourierMix} Data Augmentation on CIFAR-10. To better highlight the visual patterns of \emph{FourierMix}, we utilize the highest severity level for $\mathbf{A}(\cdot)$ and $\mathbf{P}(\cdot)$ in this figure. }
    \label{fig:fmixcifar10}
\end{figure*}

\begin{figure*}
    \centering
    \includegraphics[width=0.9\linewidth]{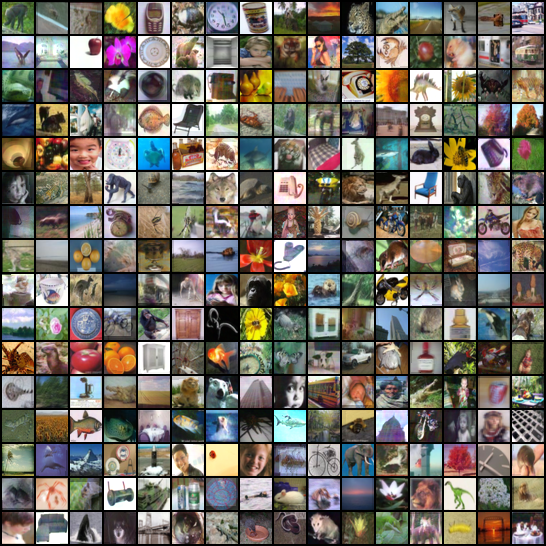}
    \caption{Sample Images from \emph{FourierMix} Data Augmentation on CIFAR-100. To better highlight the visual patterns of \emph{FourierMix}, we utilize the highest severity level for $\mathbf{A}(\cdot)$ and $\mathbf{P}(\cdot)$ in this figure.}
    \label{fig:fmixcifar100}
\end{figure*}

We visualize more sample images from our created datasets in Figure~\ref{fig:fsamples} using different classes. It is also worth noting that \emph{FourierMix} augmented images (Figure~\ref{fig:fmixcifar10} and~\ref{fig:fmixcifar100}) have different patterns with CIFAR-10/100-F.

\section{Discussion on Test-Time Adaptation}

As discussed in Appendix~\ref{sec:related}, another widely acknowledged approach to counter OOD shifts is test-time adaptation. We thus perform a preliminary study on how test-time adaptation will affect the certified robustness. Specifically, we use BN~\cite{saenko2010adapting} and TENT~\cite{wang2021tent} as representative methods. Since the theorem derived by Cohen~\etal~\cite{cohen2019certified} requires the base classifier $\mathcal{M}$ to be deterministic, we cannot apply BN and TENT in an online manner. To deal with such a problem, while evaluating the ACR of OOD data from a specific corruption type, we randomly sample 500 (out of 10,000) images from the OOD test set for the adaptation. We follow other settings specified in~\cite{saenko2010adapting,wang2021tent} for our experimentation. Table~\ref{tb:testtime} presents the detailed results on CIFAR-10-C. We find that test-time adaptations fail to improve the ACR in the OOD setting. The reason is that \emph{one-shot} adaptation relies upon a small amount of data which is not sufficient to correct the OOD shift.
In contrast, it may cause the base classifier $\mathcal{M}$ to become biased towards the small subset of test data used for adaptation. We highlight that certification of adaptive models is also a potential direction that can help with OOD certified robustness. More theoretical support is needed in this direction, and we leave it as a promising future work.

\end{document}